%% file: loop.tex
\documentclass{article} %
\usepackage{loop,times}
\renewcommand{\headrulewidth}{0pt} %

\input{math_commands.tex}

\usepackage{hyperref}
\usepackage{url}
\usepackage{graphicx}
\usepackage{tikz}
\usepackage{xcolor}
\usepackage{booktabs}
\usepackage{caption}
\usepackage{subcaption}
\usepackage{wrapfig}
\usepackage{tabularx}
\usepackage{algorithm}
\usepackage{algpseudocode}
\usepackage{xcolor}
\usepackage{comment}
\usepackage{amssymb}

\usepackage{adjustbox}
\usetikzlibrary{positioning,arrows.meta,calc,decorations.pathreplacing}

\title{
How Model Growth, Recursion, and Boundary Operators Influence Scaling Exponents
}

\author{Zixi Chen$^{1\dagger}$ \quad Akshay Vegesna$^{2}$ \quad Samip Dahal$^{2}$ \quad Andrew Gordon Wilson$^{1,2}$ \\
\vspace{2pt}
$^{1}$New York University \quad $^{2}$Q Labs
}

\usepackage{xspace}

\newcommand{\Kone}{Operator-1\xspace}
\newcommand{\Ktwo}{Loop-2\xspace}
\newcommand{\Dep}{Untied-2\xspace}
\newcommand{\KtwoGrow}{Loop-Grow\xspace}
\newcommand{\DepGrow}{Untied-Grow\xspace}

\newcommand{\eff}{\mathrm{eff}}

\definecolor{ugBoundary}{HTML}{006FD8}
\definecolor{ugGrowth}{HTML}{C9302D}
\newcommand{\ugboundary}[1]{\begingroup\setlength{\fboxsep}{2pt}\colorbox{ugBoundary!15}{\strut #1}\endgroup}
\newcommand{\uggrowth}[1]{\begingroup\setlength{\fboxsep}{2pt}\colorbox{ugGrowth!15}{\strut #1}\endgroup}

\newcommand\blfootnote[1]{%
  \begingroup
  \renewcommand\thefootnote{}%
  \footnotetext{#1}%
  \endgroup
}

\NewDocumentCommand{\paperfigure}{m m m O{htbp}}{%
\begin{figure}[#4]
\centering
\includegraphics[width=\linewidth]{#3}
\caption{#2}
\label{#1}
\end{figure}%
}

\iclrfinalcopy %
\begin{document}
\maketitle
\blfootnote{$\dagger$: Work done as an intern at Q Labs.}
\blfootnote{Code available at \url{https://github.com/qlabs-eng/scaling-exponents}}

\begin{abstract}
Scaling laws predict how loss decreases with increases in computation. We show, contrary to conventional wisdom, that architectural interventions can modify scaling exponents in pre-training, leading to power-law improvements in performance as computation increases. As an anchoring point, we consider the architectural formulation of looped transformers. Although not typically used in this way, looping, also known as recursive depth, provides a mechanism for model growth, by increasing the number of loops during training. Model growth, with and without shared weights, provides the biggest changes to the scaling exponents. In particular, a 7.4B model growth architecture matches GPT-3 13B on CORE with roughly $20\times$ less compute, and has compute efficiency gains that increase with scale. Moreover, simply using a boundary operator in a vanilla transformer, which normalizes and injects an earlier block, also provides an exponent increase, although to a lesser extent. In the data-constrained, multi-epoch setting, standard looping has a useful regularizing effect, where we find it is compute-optimal to increase the number of loops with scale. These results can be understood through the lens of computational depth: for a given computational budget, we wish to increase the usable depth of the transformer, which can lead to efficiency gains that increase with scale.
\end{abstract}

\section{Introduction}

\emph{Scaling laws} predict loss as a function of numbers of parameters and datapoints. They provide a recipe for configuring a balance of training data and model size to be on the \emph{compute-optimal frontier}, providing the lowest loss for any computational budget \citep{kaplan2020scaling,hoffmann2022training}. Scaling laws follow a power law that depends on scaling constants and exponents. 
The scaling constants govern the vertical translation of loss curves as a function of compute, while the exponents affect the shapes of the curves themselves \citep{kaplan2020scaling, hoffmann2022training}. Modifying even the constants can have a significant effect on common practice. For example, \citet{qiu2026hyperparameter} and \citet{liu2025muon_scalable} recently showed that with the correct hyperparameter scaling, the Muon optimizer can provide a 40\% compute efficiency gain over the optimizer AdamW, across scales. Muon is thus a promising candidate as the new default optimizer, de-throning Adam after nearly a decade of dominance.

In this paper, we ask what architectural interventions could possibly influence scaling \emph{exponents}. It is the conventional wisdom that changes to the architecture generally only affect the scaling constants \citep{bansal2022data,hestness2017deep,attention2026residual}. But a change to the scaling exponent could be transformative, leading to power-law improvements in performance with increases in computation. And perhaps a change to the exponent is not as elusive as it might seem --- even seemingly small hyperparameter details can influence whether an intervention affects the scaling law, as has been seen with Muon \citep{qiu2026hyperparameter}.

\begin{figure}[t]
\centering
\includegraphics[width=\linewidth]{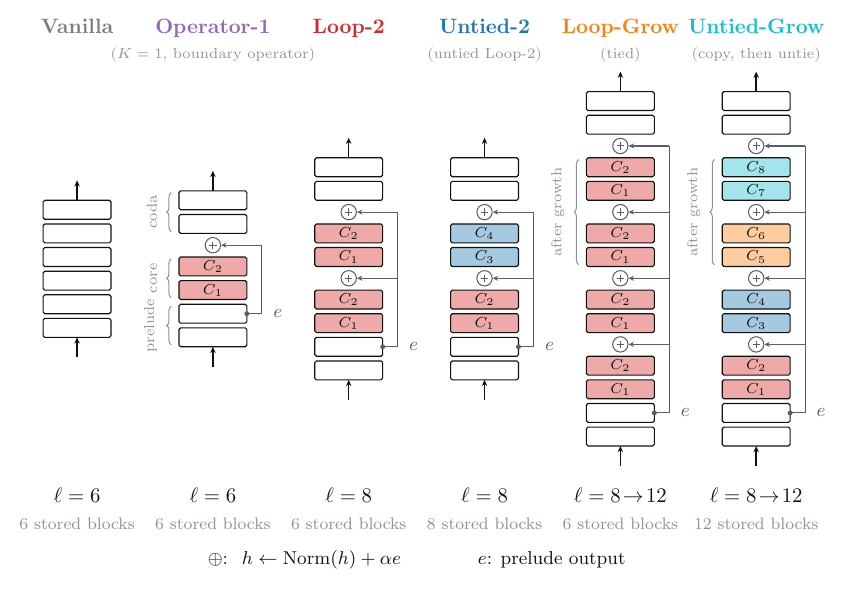}
\caption{\textbf{Model growth, looping, and untied looping are members of one prelude--core--coda family.} In this family, a prelude of transformer blocks embeds the input into a representation $e$, a core of blocks is applied $K$ times to a state $h$, and a coda of blocks produces the output. All variants share the same structure and differ on three axes, each isolated by one comparison. First, Vanilla and \Kone match in parameters and FLOPs and isolate the boundary operator $\oplus$, which normalizes the residual stream and adds back $e$. Second, \Ktwo and \Dep have the same FLOPs and isolate weight sharing, since \Ktwo stores one core and \Dep stores two. Third, \KtwoGrow and \DepGrow isolate growth, with braces marking the passes that turn on at the transition and $\ell$ giving the depth before and after.}
\label{fig:arch}
\end{figure}

Our starting intuition is the idea of \emph{computational depth}: we may wish to achieve the greatest depth for any computational budget, in order to capture hierarchical structure in data, and compose many steps of computation. To this end, we consider \emph{model growth}, whereby we grow the depth of the model during training. This approach is motivated by evidence that neural networks tend to learn simpler patterns early in training, with more complex functions or finer-scale components emerging as training progresses \citep{nakkiran2019sgd,rahaman2019spectral}. We hypothesize that, under a fixed computational budget, allocating greater depth to these later stages may therefore be beneficial. Although not typically used for this purpose, \emph{looping} \citep{dehghani2019universal,yang2024looped}, also known as \emph{recursive depth} \citep{geiping2026scaling}, provides a mechanism for model growth. 
A looped transformer applies the same core block of layers several times, corresponding to the number of loops, in a single forward pass, and every pass shares one set of weights.
Typically the number of loops is fixed or random, and thus does not provide model growth. However, if we increase the number of loops during training, we effectively increase depth, without increasing the number of parameters. Alternatively, we can \emph{untie} the weights, giving each pass its own copy of the core, so that growing the number of passes adds new blocks with distinct parameters. To explore these questions, we use the architectural formulation of looped transformers in \citet{geiping2026scaling}, which compartmentalizes a transformer into prelude, core, and coda blocks, with a core block that is looped. We also consider standard looped transformers, which do not provide any model growth during training, and the architectural specification of the looped transformer without looping, which is simply a standard transformer but with a boundary operator between blocks and prelude injection. We illustrate each of these architectures in Figure~\ref{fig:arch}. 

With increased computation, we scale each of these architectures in a compute-optimal fashion, which means increasing the size of prelude, core, and coda blocks equally. We distinguish standard compute-optimal scaling, which scales the size of the model and data with increased computation, and model growth, which grows the size of the model during training itself. We consider performance in data unconstrained settings, and data constrained settings where we train for multiple epochs. 
We note that looping is mostly used for \emph{inference-time} compute scaling in reasoning tasks, or for parameter efficiency, rather than as a way to train more efficiently for a fixed compute budget \citep{geiping2026scaling,saunshi2025reasoning,yang2024looped}. 

We highlight some of our key results in Figure~\ref{fig:main-results}:
\begin{itemize}
\item Contrary to conventional wisdom, it is possible to change the scaling exponent in pre-training through architectural interventions. Earlier growth and looping studies do not focus on compute-optimal scaling with architecture-specific scaling hyperparameters, which may explain why the effect has gone unnoticed.
\item The exponent differences are most obvious in looking at \emph{compute multipliers}: the multiple of compute a standard transformer would require to reach the same value of the loss. For a scaling exponent improvement, the compute multipliers \emph{increase} with scale, as they do for every variant we consider, both in FLOPs and in wall-clock time (Figure~\ref{fig:ladder-runtime}).
\item Growing the model during training, by increasing the number of core passes with \emph{untied} weights, improves the scaling exponent. The grown model reaches the loss of a standard transformer with $1.55\times$ less compute at $10^{20}$ FLOPs, and this gap widens with scale.
\item Looping provides a mechanism for parameter-efficient model growth. Growing the number of loops with tied weights obtains the exponent improvement from growth with the same number of parameters as a vanilla transformer, and trails untied growth by only a small constant factor, for a $1.36\times$ compute multiplier over a standard transformer at $10^{20}$ FLOPs.
\item Notably, simply using the boundary operator in a vanilla transformer, which normalizes and injects the prelude block, also improves the scaling exponent, though less than model growth, with a $1.25\times$ compute multiplier over a standard transformer at $10^{20}$ FLOPs.
\item In the multi-epoch setting, looping has a helpful regularizing effect. Training for 10 epochs on 100M tokens, the optimal number of loops increases with compute, and a larger number of loops decreases overfitting. Scaling the number of loops reaches the best loss of a weight-decay-tuned standard transformer with $2.2\times$ less compute.
\end{itemize}

Moreover, Figure~\ref{fig:dep-grow-extrap} shows that our scaling laws hold under extrapolation. A $7.4$B model growth architecture trained at $8\times$ the largest fitted compute lands on the predicted loss curve, and matches GPT-3 13B on CORE \citep{li2024datacomp,miniseries} with roughly $20\times$ less compute \citep{GPT3}. Because the compute multiplier grows with scale, the $1.8\times$ advantage over a standard transformer at this budget ($1.2\times10^{21}$ FLOPs), marked in the figure, is projected to reach $2.7\times$ at $10^{25}$ FLOPs.

We can gain insights into these results through the lens of \emph{computational depth}. In particular, we define the computational depth as the number of layers meaningfully influencing the predictive distribution for a given computational budget. In Section~\ref{sec:discussion}, we interpret our results through the frame of computational depth.

The rest of the paper is organized as follows. Section~\ref{sec:background} provides background on scaling laws, model growth, looped transformers, and the curse of depth. Section~\ref{sec:arch} introduces the prelude--core--coda family that unifies model growth, looping, and standard transformers. Section~\ref{sec:compute-optimal} considers single-epoch training, where we fit a compute-optimal recipe for each architecture (Section~\ref{sec:recipe}), show that model growth and the boundary operator improve the scaling exponent (Section~\ref{sec:ladder-results}), validate the fitted laws through extrapolation (Section~\ref{sec:extrap}), and give prescriptions for practitioners (Section~\ref{sec:prescription}). Section~\ref{sec:data-constr} then turns to multi-epoch training in the data-constrained regime, where we find that scaling the number of loops is preferable to scaling parameters. Section~\ref{sec:discussion} interprets these results through the lens of computational depth. Finally, in Section~\ref{sec:conclusion} we discuss directions for future work.

\paperfigure{fig:main-results}{
    \textbf{In single-epoch compute-optimal training, model growth and the boundary operator improve the scaling exponent, and in multi-epoch training the optimal loop count grows with compute.}
    \textbf{Left: single epoch.} In the top panel, we fit scaling laws to validation loss on FineWeb, showing that the boundary operator and model growth can improve scaling exponents (Equation~\ref{eq:compute-optimal}). In the bottom panel, we see these interventions lead to compute efficiency multipliers over vanilla transformers that increase with scale. The gain is largest for \DepGrow, which reaches $1.55\times$ at $10^{20}$ FLOPs.  
    \textbf{Right: multi epoch.} We fix a pool of 100M unique FineWeb tokens, and train for ten epochs. In this regime, the optimal number of loops increases with compute, and scaling the number of loops beats scaling parameters. At the marked point, \Kone needs $2.2\times$ the compute to match the looped loss. The top panel shows validation loss against compute at fixed numbers of loops from 1 to 12 and fixed weight decay. The bottom panel compares two ways of spending more compute: scaling the number of loops at fixed weight decay, or scaling the size of \Kone with tuned weight decay. Scaling the number of loops does not require much hyperparameter tuning, since the dashed line, which tunes weight decay at every loop count, stays close to the fixed-weight-decay curve.
}{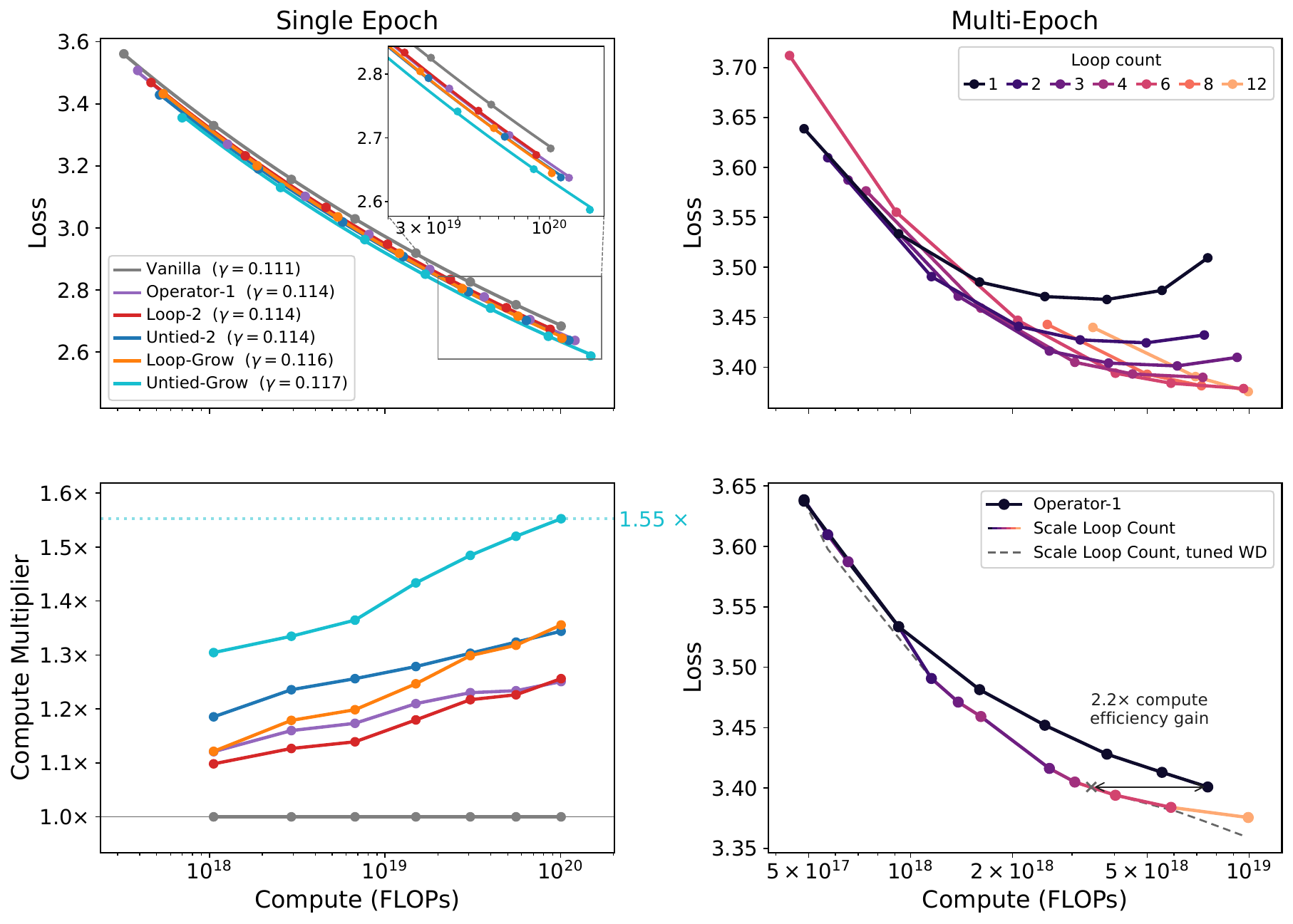}[t]
\section{Background}
\label{sec:background}

\paragraph{Scaling laws and compute multipliers.}
A scaling law predicts how loss falls as training compute grows. \citet{hoffmann2022training} write loss as three terms: a power law in the parameter count $N$, a power law in the number of training tokens $T$, and an irreducible floor $E$, giving $L(N,T)=E+(N/N_0)^{-\alpha}+(T/T_0)^{-\beta}$ with $N_0$, $T_0$, $\alpha$, $\beta$ fitted. Each training token costs about $6N$ floating-point operations, so training compute is $C\approx 6NT$ \citep{kaplan2020scaling}, and a fixed compute budget leads to a trade-off between parameters and tokens. Choosing the best split at every budget gives the compute-optimal loss, which is again a power law,
\begin{equation}
L(C)=E+A\left(\frac{C}{C_0}\right)^{-\gamma},
\label{eq:compute-optimal}
\end{equation}
where $A$ and $\gamma$ are fitted and $C_0$ is a chosen base compute, which we set to the compute at which Vanilla is tuned. An architecture is therefore a family of models indexed by compute. The compute-optimal recipe depends on how $N$ and $T$ grow with compute \citep{hoffmann2022training} and on how hyperparameters are scaled \citep{yang2022tensorprogramsvtuning,qiu2026hyperparameter}.

To compare two architectures, we invert Equation~\ref{eq:compute-optimal} and ask how much compute each needs to reach the same loss. Let $\hat{C}_A(\ell)$ denote the compute at which architecture $A$'s fitted law reaches loss $\ell$. The compute multiplier of architecture $B$ over $A$ at loss $\ell$ is then $\pi(\ell)=\hat{C}_A(\ell)/\hat{C}_B(\ell)$. For example, $\pi(\ell)=1.25$ means $A$ needs $1.25\times$ the compute of $B$ to reach loss $\ell$. Throughout, $A$ is the standard transformer and $B$ is the variant. The multiplier can also be indexed by the standard transformer's compute, $\pi(C)=\pi(L_A(C))$, the multiplier at the loss the standard transformer reaches with budget $C$.

Techniques that help at small scale can fail to improve, or the gap can close, at larger scales \citep{rae2021scaling}, so improvements must be shown across scales \citep{kaplan2020scaling,liu2025muon_scalable,krajewski2024scaling,andres2026einsum}. Some interventions improve the exponent: transformers scale with a better parameter scaling exponent than LSTMs by exploiting long contexts \citep{kaplan2020scaling}, and Mixture-of-Experts scaling laws predict a growing compute-efficiency advantage over dense transformers as training budgets increase, with further gains from optimizing expert granularity \citep{krajewski2024scaling}. Others improve the constant, including structured matrices for MoE \citep{andres2026einsum} and the Muon optimizer over Adam \citep{liu2025muon_scalable}. Constant improvements are more common than exponent improvements. On the theoretical side, \citet{bordelon2025feature} show that feature learning improves the exponent on hard tasks.

Compute-optimal scaling laws depend on the training recipe. In particular, incorrect hyperparameter scaling yields a worse compute-optimal law \citep{yang2022tensorprogramsvtuning,qiu2026hyperparameter}. To rule out hyperparameter scaling issues from the comparison, we fit a separate compute-optimal recipe for every architecture (Section~\ref{sec:compute-optimal}).

\paragraph{Model growth.}
When training a ladder of models, each larger model is trained from scratch, so the compute spent on the smaller runs is wasted. Model growth addresses this waste by reusing the trained weights of a smaller model to initialize a larger training run \citep{chen2015net2net, du2024stacking}. \citet{du2024stacking} compare several ways to expand the parameter count and find that stacking transformer blocks, i.e., copying the trained blocks to increase depth partway through training, is the most efficient, reaching the same loss with a 50\% speedup. However, \citet{liew2025reusing} show that the more extensively a base model is pretrained, the less benefit further pretraining provides. This finding suggests that the benefits of checkpoint reuse may depend on the allocation of tokens between the two pretraining stages, motivating joint optimization of their token budgets. Related growth work, from staged training with function-preserving operators \citep{shen2022staged} to recycling converged mixture-of-experts checkpoints \citep{wang2026beyond}, reports gains at one or a few target sizes rather than a change in the compute-optimal scaling law. In this work, we set aside the checkpoint-reuse motivation altogether and instead treat model growth as a way to increase the computational depth of the final model.

\paragraph{Looped transformers.}
A looped transformer applies the same block of layers several times in one forward pass \citep{dehghani2019universal,yang2024looped}. Looping therefore raises computational depth by reusing a block rather than adding a new one. Looping has two main motivations: an inductive bias toward iterative computation \citep{dehghani2019universal}, and the ability to trade extra passes for accuracy at test time \citep{geiping2026scaling}. At matched computational depth, \citet{saunshi2025reasoning} find that looping performs similarly to a dense transformer on reasoning tasks but much worse on memorization tasks. Nonetheless, recursive reasoning models, a close relative of looped transformers, reach results comparable to state-of-the-art models of the time on reasoning tasks with far fewer parameters and far less training compute \citep{wang2025hierarchicalreasoningmodel, jolicoeur2025less}.

In the pretraining setting, HRM-Text \citep{wang2026hrm}, a 1B-parameter recursive reasoning model trained from scratch on 40B synthetic and real tokens for about \$1,500, was recently shown to perform competitively with 2--7B-parameter open models on reasoning benchmarks. However, its setup differs substantially from standard pretraining, since it trains on instruction--response pairs with a task-completion objective rather than raw text, which makes the contribution of looping hard to isolate. Controlled comparisons in standard pretraining are more mixed. \citet{prairie2026parcae} show that looped transformers achieve lower loss than dense transformers at matched parameter and data budgets, and study compute-optimal allocation between looping and data at fixed model size. They do not, however, establish an advantage over dense transformers when model size and training data are jointly optimized for compute. Similarly, \citet{schwethelm2026much} show that, at matched computational depth, more recurrence leads to strictly worse performance when model size and data are jointly optimized. In this work, we show that with a proper block allocation and an optimal loop count, looping can improve performance under either control. Concurrently, \citet{wang2026smelt} show that a sparse looped MoE has compute-efficiency gains that grow with scale, on proprietary data and architecture. In our setting, by contrast, we find that weight sharing alone is not enough to improve the exponent.

\paragraph{Depth scaling and the curse of depth.}
The effectiveness of depth scaling has been contested. On one side, \citet{tay2021scale} argue that deep-narrow T5 models are Pareto-better on downstream tasks despite similar pretraining losses, and \citet{liu2024mobilellm} find that scaling depth beats scaling width at sub-billion scale. On the other, \citet{kaplan2020scaling} find that depth does not have a strong effect on the scaling law, and \citet{levine2020limits} derive that the optimal depth should grow only logarithmically with width. One explanation for depth's limited returns is known as the \emph{curse of depth}. In pre-norm transformers, where normalization is applied to the input of each block rather than to the residual stream itself, the residual stream grows with depth, so each block's update is a shrinking fraction of the stream and deeper blocks drift toward doing nothing \citep{liu2020understanding, sun2025curse}. The curse can be measured with the logit lens, which decodes the residual stream after each block into a prediction and measures how far it is from the model's final output distribution. The depth beyond which blocks stop changing the prediction is the effective depth \citep{logit2020lens, csordas2026language}. We refer to this quantity as the KL effective depth for clarity in this work.

Prior work mitigates the curse by scaling the pre-normalization output inside the residual branch~\citep{sun2025curse}, normalizing the residual stream~\citep{wang2026hrm, loshchilov2025ngpt}, or injecting earlier representations into later ones~\citep{wang2026hrm, nanochat}. Similar ingredients appear in the boundary operators of looped transformers. These depth-related interventions improve loss at a fixed model size, but none has been shown to improve the scaling exponent, leaving open whether mitigating the curse of depth can change the scaling law.

\paragraph{Data-constrained scaling and multi-epoch training.}
Compute is growing faster than the stock of high-quality text \citep{villalobos2024rundatalimitsllm}, so pretraining will increasingly repeat data over multiple epochs. Repeated tokens are worth less than fresh ones. \citet{muennighoff2023scaling} fit a scaling law in which repeated data counts for less than new data, and later work studies how loss behaves when the amount of unique data is fixed and compute keeps growing \citep{kim2026pre, lovelace2026prescriptive}. The cost of repetition is that the models start overfitting, which can be exacerbated with model size. Weight decay is the standard remedy. \citet{kim2026pre} show that larger models need more of it, so it must be retuned at every scale, and \citet{lovelace2026prescriptive} show that at a fixed weight decay the overfitting penalty follows a power law in model size. Looping adds depth without adding parameters, and Section~\ref{sec:data-constr} tests whether this lets a model add capacity in this regime without adding overfitting.

\section{Architectures}
\label{sec:arch}

Every model we train is one three-stage network, following the prelude--core--coda formulation of \citet{geiping2026scaling}: a prelude of transformer blocks embeds the input, a core of blocks is applied $K$ times, and a coda of blocks produces the output (Equation~\ref{eq:looped-general}). Model growth, looping, and deep transformers are all variants of this one network, so we can compare them on equal footing.

\begin{equation}
    e = \mathcal{P}(s),
    \qquad
    h_{k} = \mathcal{R}_{\theta_k}\big(\phi(h_{k-1}, e)\big),
    \quad k = 1, \dots, K,
    \qquad
    y = \mathcal{C}\big(\rho(h_K, e)\big)
    \label{eq:looped-general}
\end{equation}

Here $\mathcal{P}$, $\mathcal{R}$, and $\mathcal{C}$ are the prelude, core, and coda, $\phi$ and $\rho$ are boundary operators that mix the state with the prelude output $e$ before each core pass and before the coda, $\theta_k$ are the weights of the $k$th pass, and $K$ is the loop count. With $P$, $C$, and $D$ blocks in the three stages, a token passes through $\ell = P + KC + D$ blocks, which we call the executed depth. We distinguish executed depth from the number of stored blocks, which we use for model size throughout. We fix the width-depth ratio at 128.

The architecture variants we study differ from one another on three axes: what happens at the boundary between passes, whether the passes share weights, and when the passes turn on. We isolate each axis with a matched comparison that holds everything else fixed (Figure~\ref{fig:arch}), so a difference in scaling can be attributed to a single change.

\paragraph{Boundary operator (BO).}
A standard transformer is $K=1$ with identity boundary operators. Between core passes and before the coda, we instead apply
\begin{equation}
    \mathrm{BO}(h,e) = \mathrm{Norm}(h) + \alpha\,e,
    \label{eq:boundary-operator}
\end{equation}
which sets $\phi = \rho = \mathrm{BO}$ in Equation~\ref{eq:looped-general}. The operator has two parts, each with its own purpose. In a pre-norm block the residual stream grows with depth, so each update is a shrinking fraction of the stream, and normalizing lets every pass write at full weight. Adding back the prelude output $e$ keeps every pass conditioned on the input. Normalizing between passes is standard in looped transformers \citep{geiping2026scaling,wang2026hrm} and a known remedy for the curse of depth. Likewise, re-injecting the input appears in recurrent models \citep{geiping2026scaling,prairie2026parcae,schwethelm2026much} and in fixed-depth transformers such as nanochat and modded-nanogpt \citep{nanochat,moddednanogpt}. Prior looped transformers normalize before the coda but re-inject $e$ only between core passes. We instead apply the same normalize-and-inject map before the coda as well, and find it to be important (Table~\ref{tab:base-tuning-operators}, Figure~\ref{fig:ladder-arch}).

\paragraph{Looping.}
With the boundary operator fixed, the variants differ only in the core weights $\theta_k$ and in when the passes are active. Tying the weights, $\theta_1 = \cdots = \theta_K$, gives a looped transformer: one core is stored and applied $K$ times, so the model stores $P + C + D$ blocks and executes $P + KC + D$. Untying the weights gives each pass its own core. The untied model has the same computation graph and the same FLOPs as the tied one, but stores $P + KC + D$ blocks, so it is a deep transformer with the boundary operator. The untied model is therefore our control for separating the effect of depth from the effect of weight sharing.

\paragraph{Model growth.}
Using model growth, we start training at a small loop count $K$ and raise it partway through. In the tied case the existing core is simply applied more times, adding no weights. In the untied case the trained core is copied and the copies are then trained separately, which is the depthwise stacking of \citet{du2024stacking}. Growth changes only $K$, and only within a single run, whereas along a scaling ladder the prelude, core, and coda all grow together with model size. Finally, we fix the remaining choices that vary across prior looped transformers: the initial state is $h_0 = 0$, the loop count is fixed rather than sampled, and gradients flow through every pass.

\paragraph{Architecture variants.}

Figure~\ref{fig:arch} draws the six models we train at the smallest model size, arranged as three matched comparisons, one per axis. The first isolates the boundary operator: Vanilla is the standard pre-norm transformer, and \Kone matches it in parameters and FLOPs but adds the operator between the core passes and before the coda. The second isolates weight sharing: \Ktwo and \Dep both apply the core twice with the operator between passes and are matched in FLOPs and depth. The third isolates model growth: \KtwoGrow and \DepGrow each start as their fixed counterpart and double the core passes partway through training. Two operator-free controls, not shown in the figure, complete the family. Deep Vanilla is \Dep without the operator, and Deep Vanilla Grow is \DepGrow without the operator, so the pair isolates growth without the operator. We give the full training procedure for \DepGrow in Algorithm~\ref{alg:untied-grow}.

For each model along the compute-optimal scaling ladders, we use the notation $d\ell$ to denote a nominal depth of $\ell$ transformer blocks at our fixed width-depth ratio of 128, so $d8$ has a width of 1024. Vanilla, \Kone, and the tied variants at $d\ell$ store exactly $\ell$ blocks, split across prelude, core, and coda, whereas the untied variants store an additional copy of the core for each extra pass, so \Dep at $d8$ stores 11 blocks (Table~\ref{tab:model-sizes}). With the family fixed, what remains is how to train each member compute-optimally, which is the subject of the next section.

\begin{algorithm}[t]
\caption{Untied Grow: \ugboundary{normalize and re-inject} with \uggrowth{untied core growth}}
\label{alg:untied-grow}
\begin{algorithmic}[1]
\Require Training batches $\{(x_t,y_t)\}_{t=1}^{T}$; injection scale $\alpha$
\Require Growth step $g$; initial core count $K_0$; final core count $K_f=mK_0$, $m\ge2$
\State Initialize prelude $\mathcal{P}$, coda $\mathcal{C}$, and independent cores $\{\mathcal{R}_{\theta_k}\}_{k=1}^{K_0}$
\State $K\gets K_0$
\For{$t=1,\ldots,T$}
    \If{$t=g+1$} \Comment{Grow after $g$ training steps}
        \For{$r=1,\ldots,m-1$}
            \For{$j=1,\ldots,K_0$}
                \State \uggrowth{$\theta_{rK_0+j}\gets\operatorname{copy}(\theta_j)$} \Comment{Stack a copy of the core}
            \EndFor
        \EndFor
        \State $K\gets K_f$ \Comment{Activate new untied cores}
    \EndIf
    \State $e\gets\mathcal{P}(x_t),\quad h\gets0$
    \For{$k=1,\ldots,K$}
        \State $h\gets$ \ugboundary{$\operatorname{RMSNorm}(h)+\alpha e$} \Comment{Boundary before each core}
        \State $h\gets\mathcal{R}_{\theta_k}(h)$ \Comment{Distinct weights for every pass}
    \EndFor
    \State $h\gets$ \ugboundary{$\operatorname{RMSNorm}(h)+\alpha e$} \Comment{\textbf{Also before the coda}}
    \State $\hat y_t\gets\mathcal{C}(h)$
    \State $\mathcal{L}_t\gets\operatorname{CrossEntropy}(\hat y_t,y_t)$
    \State Update all active parameters using $\nabla\mathcal{L}_t$ \Comment{Backpropagate through all $K$ passes}
\EndFor
\State \Return Trained model
\end{algorithmic}
\end{algorithm}

\section{Improving the Compute-Optimal Scaling Exponent}
\label{sec:compute-optimal}
In this section we compare the compute-optimal scaling laws of the model families of Section~\ref{sec:arch} on fresh tokens, paying special attention to whether the gaps between them widen with scale or stay constant. Throughout, we train on FineWeb \citep{fineweb} with the GPT-2 tokenizer \citep{gpt2}. We describe the main setup here and defer details to Appendix~\ref{app:comp-opt}.

\subsection{Compute-Optimal Recipe}
\label{sec:recipe}

A gap in scaling exponents is only meaningful if every architecture is well tuned, since otherwise a difference in exponents could be a difference in tuning \citep{qiu2026hyperparameter}. Indeed, transferring Vanilla's recipe to \Kone costs $7.8\times10^{-3}$ loss at a depth of 8 transformer blocks ($d8$) and erases its exponent improvement along the ladder (Figure~\ref{fig:ladder-inject-1-van-recipe}). We therefore fit a complete compute-optimal recipe for every architecture, in the same four stages. In order, these are base hyperparameters at $d8$, tokens per stored parameter, growth timing (if applicable), and a learning-rate scaling rule. The ladders themselves run to about $10^{20}$ FLOPs, and we compare at equal compute throughout. Here we state only what the ladders depend on, and refer the reader to Appendix~\ref{app:recipe-details} for the grids, sweeps, and fits behind each stage.

Before any of the four stages, three choices are made once for the whole family, using the tied variants at 1B tokens and matched parameter counts (Appendix~\ref{app:family-design}, Figure~\ref{fig:recurrence-design}). The first is how to split the blocks across the prelude, core, and coda. The best fraction of blocks in the core is roughly constant across depth, so we scale the three stages evenly, giving leftover blocks first to the core and then to the coda. The second is the number of core passes. On fresh data the optimum lies between one and two at every budget, so the fixed variants use two core passes. The third is how many core passes to grow to. Starting from two passes, we find that four is best at every budget we tried, so the growth variants go from two to four core passes. When to grow, and how the token budget should shift for a model that will grow, are fitted along with the rest of the recipe, which we turn to next.

The four stages are then fitted separately for every architecture, since the optimal values differ across families. First, base hyperparameters are tuned at $d8$ on 1B tokens, one architecture at a time (Appendix~\ref{app:base-tuning}, Table~\ref{tab:tuned-hyperparameters}). Second, we fit the optimal tokens per stored parameter at five compute budgets and find that it does not drift with scale for any family, as in \citet{hoffmann2022training}. We therefore adopt one value per architecture, the rounded mean across budgets: $5$ for Vanilla, $6$ with the boundary operator, and $7$--$8$ with growth (Table~\ref{tab:arch-optimal-tpp}). Third, we tune a transition point for model growth, which lands at one-half to four-fifths of training (Appendix~\ref{app:growth-fitting}). Fourth, we fit a scaling rule for the learning rate at $d8$--$d10$, under which the learning rate decreases with model size. This single rule is enough, since re-sweeping and scaling the remaining hyperparameters moves the compute multiplier by at most $3\%$ (Figure~\ref{fig:ladder-vanilla-scaling}). With the recipe fixed, the differences we report in Section~\ref{sec:ladder-results} reflect the architecture rather than its tuning.

\subsection{Improving the Exponent of the Scaling Laws}
\label{sec:ladder-results}

We run the compute-optimal recipes for each architecture independently. We fit the irreducible loss $E$ in Equation~\ref{eq:compute-optimal} using a Huber loss for Vanilla \citep{hoffmann2022training}. For the remaining architectures, we fit an affine relationship in log--log space, $\log (L - E) = - \gamma \log (C/C_0) + \log A$, to obtain the exponent $\gamma$ and the constant shift $\log A$. For compute multipliers, we estimate the compute required to reach Vanilla's loss by linearly interpolating between the nearest two points in log-loss and log-compute space. On the x-axis, we plot the compute of the Vanilla run rather than the validation loss. We do not see a discrepancy between downstream metrics and validation loss, and in fact, \DepGrow has slightly better downstream metrics when validation loss is controlled (Figure~\ref{fig:downstream-summary}). We defer details to Appendix~\ref{app:downstream-vs-loss}.

\begin{figure}[t]
\centering
\includegraphics[width=\linewidth]{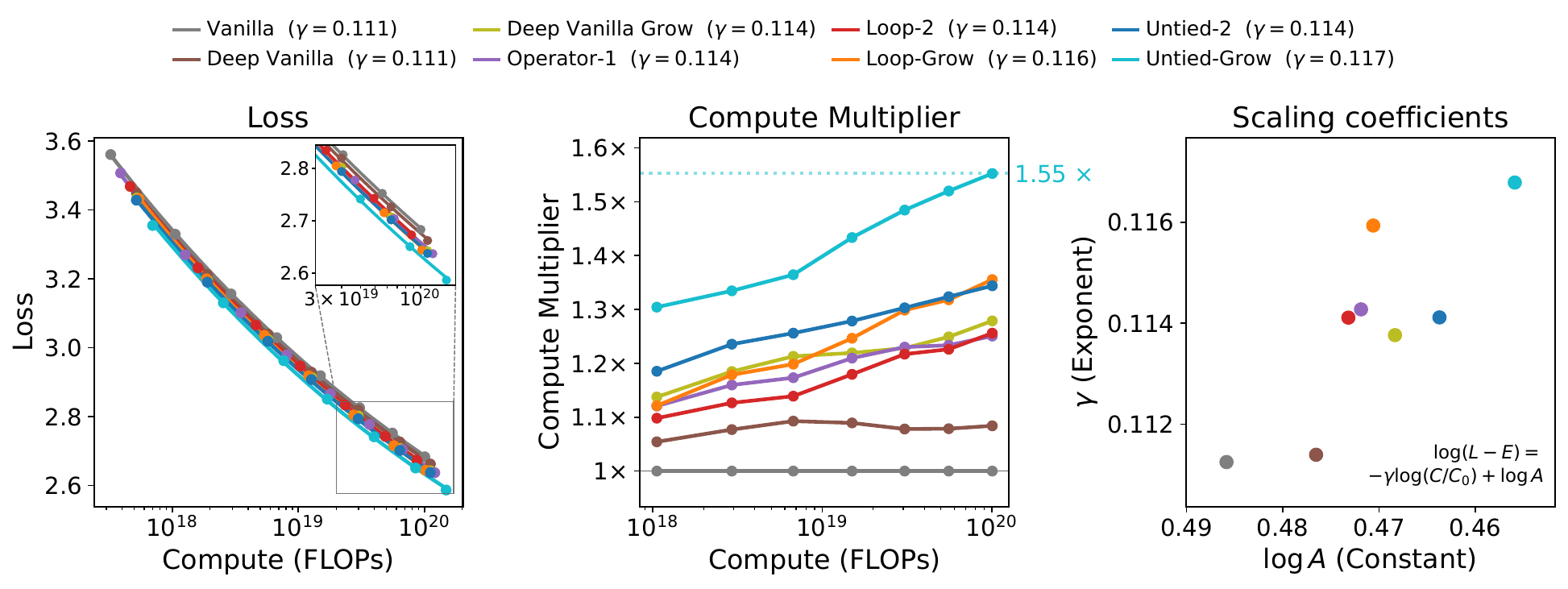}
\caption{\textbf{Model growth and the boundary operator improve the scaling exponent, while untying improves only the constant.} \textbf{Left:} validation loss on FineWeb against training compute for eight compute-optimal ladders, one dot per trained model, with fitted power laws of Equation~\ref{eq:compute-optimal} sharing an irreducible loss fitted on Vanilla. The legend gives each arm's fitted exponent. We find the regression standard error of the log--log slope to be lower than $10^{-3}$ in all cases. \textbf{Middle:} compute multiplier over Vanilla at the Vanilla budgets, interpolated in log compute without extrapolation. A flat curve is a constant improvement and a rising curve is an exponent improvement. \Kone matches Vanilla in parameters and FLOPs and rises from $1.12\times$ to $1.25\times$, so the operator alone improves the exponent. The grown families rise fastest, with \DepGrow reaching $1.55\times$ at $10^{20}$ FLOPs. \Dep sits above \Ktwo by a factor that does not widen with scale, so untying moves only the constant. Deep Vanilla stays flat near $1.08\times$, a constant gain from a more favorable width--depth ratio rather than from the operator or growth. \textbf{Right:} fitted exponent $\gamma$ against constant $\log A$ for each arm, where $\gamma$ controls the slope and $\log A$ controls the vertical translation of scaling curves; better is up and to the right. The operator and growth each move arms up and right, so they improve exponent and constant independently, whereas untying and added depth move arms only rightward. Figure~\ref{fig:ladder-runtime} shows the same ladders against wall-clock time.}
\label{fig:loop-ladder}
\end{figure}

\paragraph{The boundary operator improves the exponent.}
\Kone matches Vanilla in parameters and FLOPs and differs only by the boundary operator, and its multiplier over Vanilla increases with scale, from $1.12\times$ at $10^{18}$ FLOPs to $1.25\times$ at $10^{20}$ FLOPs. By simply adding a boundary operator, which has minimal effect on runtime, we improve the exponent of the scaling law. The same holds from Deep Vanilla to \Dep, where the only difference is the boundary operator.

We hypothesize that the operator improves the exponent because the fraction of blocks it recovers grows with depth, and compute-optimal models get deeper with compute (Section~\ref{sec:discussion}).

\paragraph{Model growth improves the exponent.}
\DepGrow starts as \Dep and differs only by doubling the core passes partway through training. Its multiplier over Vanilla widens from $1.30\times$ at $10^{18}$ FLOPs to $1.55\times$ at $10^{20}$, whereas \Dep's widens from $1.19\times$ to $1.34\times$ over the same range. Model growth's multiplier over \Dep therefore grows from $1.09\times$ to $1.16\times$, so growth improves the exponent rather than only the constant. We hypothesize that growth improves the exponent because the fraction of depth a network cannot yet use early in training grows with depth, and compute-optimal models get deeper with compute (Section~\ref{sec:discussion}). Growth does not depend on the operator. From the right panel of Figure~\ref{fig:loop-ladder}, Deep Vanilla Grow, which duplicates blocks mid-training without the boundary operator, shifts the constants and exponents over Deep Vanilla by about the same factor that \DepGrow shifts over \Dep, so the two improvements add rather than one enabling the other. With weight tying, \KtwoGrow achieves a slightly smaller gain over \Ktwo in both the exponent and the constant.

\paragraph{Untying improves the constant.}
The untied variants sit a fixed factor above their tied counterparts. \Dep is $1.08\times$ above \Ktwo at $10^{18}$ FLOPs and $1.06\times$ at $10^{20}$, and \DepGrow is $1.16\times$ and $1.14\times$ above \KtwoGrow at the same budgets. Weight sharing therefore costs a fixed factor of compute at every scale, rather than a factor that grows with the budget. Tying the weights and growing the model thus retains the exponent improvement of model growth at Vanilla's parameter count, trailing untied growth only by a constant factor. 

\paragraph{Deeper shapes improve the constant}
Width-to-depth ratios move the constant, contrary to what \citet{kaplan2020scaling} finds. A compute-optimal ladder fixes the split of compute between parameters and tokens, but not between width and depth, which we hold at a ratio of 128. Deep Vanilla, which executes the same blocks as \Dep with no operator, improves over Vanilla by a flat $1.08\times$, indicating a smaller optimal aspect ratio. However, decreasing width-to-depth ratios further doesn't bring further improvements to the scaling constant (Figure~\ref{fig:ladder-shape}).

\subsection{Extrapolation of Compute-Optimal Scaling Law}
\label{sec:extrap}

The exponents of Section~\ref{sec:ladder-results} were fitted on ladders spanning $10^{18}$ to $10^{20}$ FLOPs. To test whether the exponent gap persists to larger scales through extrapolation, we train \DepGrow at $8\times$ the largest fitted compute and ask whether the fitted law predicts the loss of that run. The law does: the run lands on the extrapolated curve, and the gain carries over to downstream performance.

We rerun the \DepGrow and Vanilla ladders on FineWeb-Edu using the same recipe fitted on FineWeb (Appendix~\ref{app:transfer-evaluation}). The gains in scaling exponents from Vanilla to \DepGrow are similar across the two datasets, while switching from FineWeb to FineWeb-Edu yields a constant multiplicative improvement in downstream performance (Figure~\ref{fig:corpus-transfer-summary}). Building on the downstream scaling law in \citet{llama3}, we develop a scaling law to predict downstream performance (Appendix~\ref{app:downstream-perf}) and set our training target to match the downstream capability of GPT-3 13B as estimated by \citet{miniseries}. Our scaling law predicts that \DepGrow $d26$, a $7.4$B model that starts at $5$B parameters, can reach this target using the compute-optimal recipe (Table~\ref{tab:full-ladder-recipes}).

\begin{figure}[h]
\centering
\includegraphics[width=\linewidth]{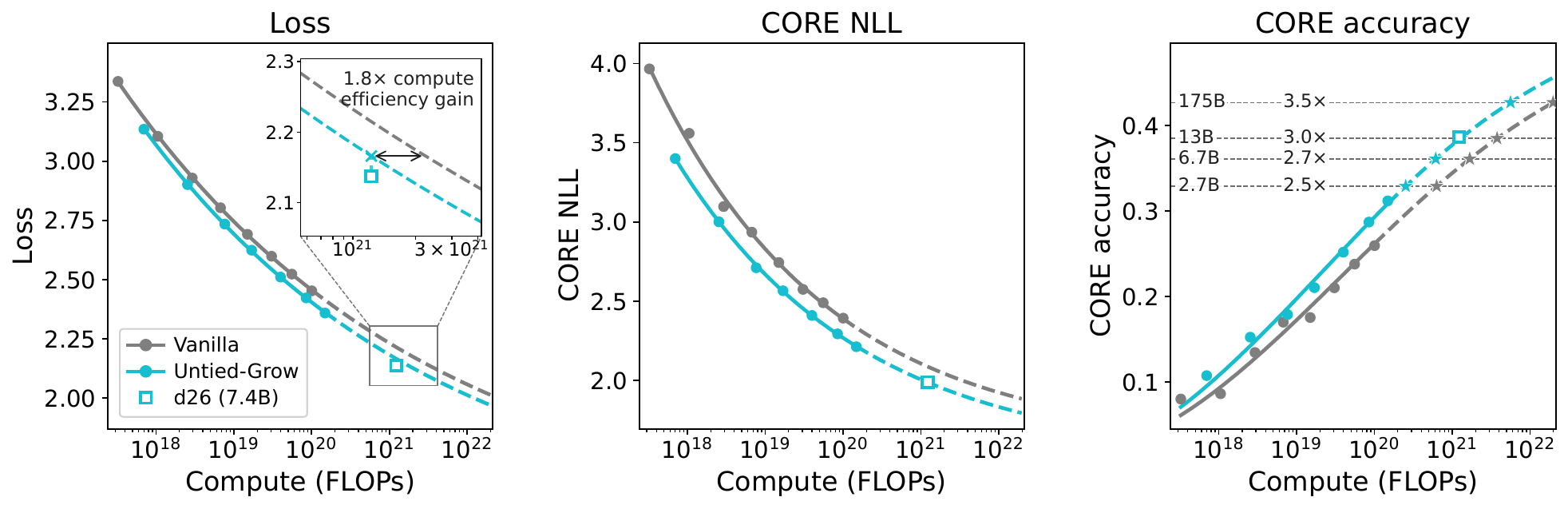}
\caption{\textbf{The scaling law fitted on $10^{18}$--$10^{20}$ FLOPs predicts an \DepGrow run at $8\times$ the fitted compute, and the compute-efficiency advantage over Vanilla widens with scale.} Ladders are trained on FineWeb-Edu, solid lines are fits, dashed lines are their extrapolation, and the square is a $7.4$B \DepGrow run not used in any fit. \textbf{Left:} validation loss on FineWeb-Edu, with a shared irreducible loss fitted on Vanilla. The run lands $0.03$ below the forecast. The inset arrow marks the compute multiplier at the run's budget: \DepGrow reaches Vanilla's loss with $1.8\times$ less compute, and the fitted laws project $2.7\times$ at $10^{25}$ FLOPs. \textbf{Middle:} answer NLL, where the run lands within $0.002$ of the forecast. \textbf{Right:} CORE accuracy predicted from compute (Appendix~\ref{app:downstream}), where the run reaches $0.387$ against a forecast of $0.384$. Horizontal lines are GPT-3 CORE scores from 2.7B to 175B, and stars mark where each curve crosses them. The Vanilla-to-\DepGrow compute ratio at those crossings grows from $2.5\times$ to $3.5\times$, so the exponent improvement of Figure~\ref{fig:loop-ladder} carries over downstream.}
\label{fig:dep-grow-extrap}
\end{figure}

Figure~\ref{fig:dep-grow-extrap} (left) shows that the $7.4$B run lands on, and in fact slightly below, the extrapolated curve, so the exponent improvement of Section~\ref{sec:ladder-results} extend to larger scales. The same holds for the downstream scaling law: the run's answer NLL and CORE score land on the curves predicted from the small-scale fits (Figure~\ref{fig:dep-grow-extrap}, middle and right).

The $7.4$B \DepGrow model reaches $0.3865$ CORE, on par with the GPT-3 13B reference of $0.3852$ \citep{miniseries}, at $1.23\times10^{21}$ FLOPs against $2.31\times10^{22}$ for GPT-3 13B \citep{GPT3}. The two models were trained on different data, and the GPT-3 reference is an estimate from a separate evaluation pipeline, so the roughly $20\times$ gap is indicative rather than a controlled comparison. 

The compute-efficiency advantage of \DepGrow over Vanilla widens with scale, as an exponent improvement predicts. Using the fitted FineWeb-Edu laws with a shared irreducible loss, the multiplier is $1.6\times$ at $10^{20}$ FLOPs, the end of the measured ladders, $1.8\times$ at $1.23\times10^{21}$ FLOPs, the compute of the $7.4$B run (Figure~\ref{fig:dep-grow-extrap}), and $2.7\times$ at $10^{25}$ FLOPs, the budget of a modern pretraining run.

\subsection{Prescriptions for Practitioners}
\label{sec:prescription}

For a standard transformer, a practitioner chooses model size and token count at a given budget. The family we study adds three choices: what fraction of the blocks goes in the core, how many times the core runs, and how much to grow and when. Fortunately, these choices can be kept fixed as we scale model size and tokens proportionally in the compute-optimal setting (Appendix~\ref{app:family-design}). This makes the recipe straightforward to use: calibrate the architecture and growth schedule at small scale, then reuse them at larger budgets. In particular, retuning the token allocation or growth fraction at each size brings little benefit in our sensitivity tests (Appendix~\ref{app:growth-sensitivity}). The full tuning procedure is in Appendix~\ref{app:recipe-details}, and Table~\ref{tab:full-ladder-recipes} lists the recipes used in our experiments.

We recommend careful, architecture-specific tuning and scaling of hyperparameters, as these are crucial to realizing the full scaling improvements (Appendices~\ref{app:recipe-details} and~\ref{app:scaling-ablations}). In our setup, the learning-rate scaling rule in Equation~\ref{eq:glr-scaling} works well while the other hyperparameters remain at their base-tuned values. Reusing a standard transformer's hyperparameters without tuning them for the new architecture can hide an exponent improvement (Figure~\ref{fig:ladder-inject-1-van-recipe}).

For single-epoch, compute-optimal pretraining, we recommend untied weights when memory is not a constraint: the untied variants reach the same loss with less compute than their tied counterparts (Section~\ref{sec:ladder-results}). Weight tying remains useful when parameter storage is the priority, retaining the exponent improvement from growth while trading a constant factor of compute efficiency for fewer parameters. In the data-constrained, multi-epoch setting, tying also provides a regularization benefit as we will see next.

\section{Looped Transformers in the Data-Constrained Regime}
\label{sec:data-constr}

\begin{figure}[t]
\centering
\includegraphics[width=\linewidth]{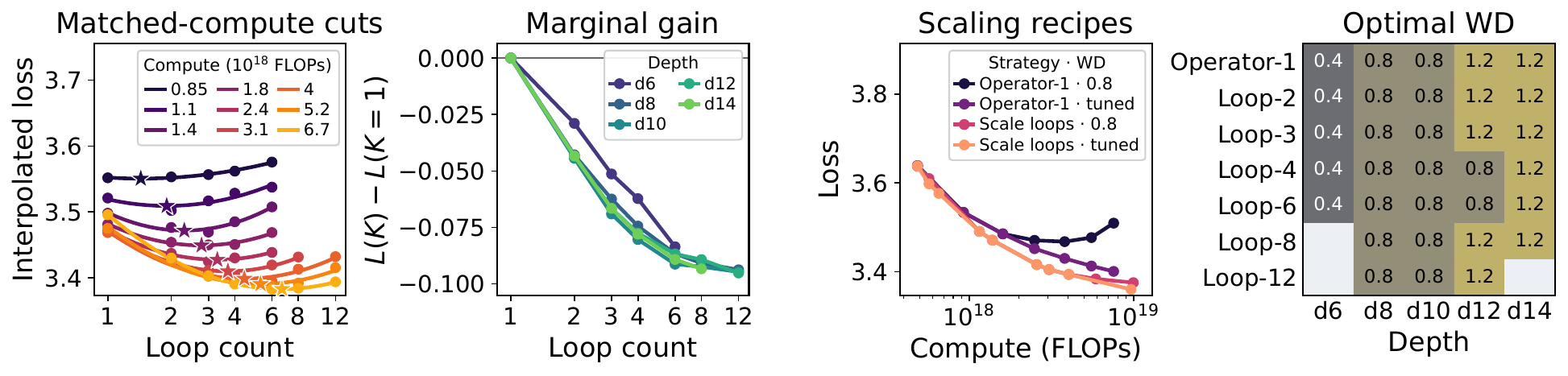}
\caption{\textbf{In multi-epoch training, the optimal loop count grows with compute, scaling the loop count beats scaling model size even against tuned weight decay, and looping leaves the optimal weight decay nearly unchanged.} Models are trained on 100M unique FineWeb tokens for 10 epochs, and all losses shown are validation losses on FineWeb. \textbf{First: matched-compute cuts.} The loss of each loop count is interpolated in compute between neighbouring depths, curves are quadratics in log loss against log loop count, and stars mark their minima. The optimal loop count rises from 1.4 to 6.7 across budgets, the gap to \Kone grows from 0.00 to 0.11. \textbf{Second: marginal gain from looping.} For each depth, we plot the loss at loop count $K$ minus the \Kone loss at the same depth. Every curve decreases monotonically, and from $d8$ upward the curves lie within 0.006 of one another at every loop count, so the gain from looping appears nearly independent of depth. \textbf{Third: weight decay when scaling loops versus model size.} Scaling model size at fixed weight decay overfits, and retuning weight decay at every model size only reaches 3.40 loss at $7.5\times10^{18}$ FLOPs. Scaling the loop count at fixed model size and fixed weight decay matches that loss with $2.2\times$ less compute, and tuning weight decay on top gains at most 0.02, so loop-count scaling is more compute-efficient and nearly free of tuning. \textbf{Fourth: optimal weight decay by depth and loop count.} The optimum rises with depth but is nearly flat in loop count, so overfitting tracks stored parameters rather than executed depth.}
\label{fig:data-constr-main}
\end{figure}

Section~\ref{sec:compute-optimal} showed that on fresh data, weight sharing is not compute-optimal, nor is increasing the loop count across scales. However, when data are repeated, we observe that the optimal loop count grows with compute, and tied weights become the better way to add capacity. This regime matters because compute is growing far faster than high-quality text \citep{villalobos2024rundatalimitsllm}, so pretraining will increasingly repeat data over multiple epochs \citep{muennighoff2023scaling,kim2026pre,lovelace2026prescriptive, slowrun2026}. Repeated data leads to overfitting, and the standard mitigation is weight decay retuned at every model size \citep{kim2026pre}. Looping offers a different mitigation: adding depth without adding parameters to overfit.

We fix a pool of 100M unique FineWeb tokens and train every model for ten epochs, a budget of 1B tokens. We use the \Kone family from Section~\ref{sec:compute-optimal}, with all hyperparameters other than weight decay fixed at the tuned values. There are then two ways to spend more compute: add parameters at a fixed loop count, or add loops at a fixed parameter count. We train a grid over both, \Kone through Loop-12 at 120M to 1.4B parameters, and sweep weight decay over the same grid.

The optimal loop count increases with compute (Figure~\ref{fig:data-constr-main}, first panel). We observe that every fixed-loop-count curve eventually overfits and turns upward, but the turn comes later and the minimum is lower for higher loop counts (Figure~\ref{fig:main-results}, upper right). At matched compute, the optimal loop count therefore rises from about one at the smallest budget to about seven at the largest, and the loss gap to \Kone grows from 0.00 to 0.11. Prior work shows that the optimal weight decay increases with parameter count on repeated data, because larger models overfit more and need more regularization \citep{kim2026pre}. Looping appears to act as a similar regularizer: it delays overfitting without adding parameters, and its optimal strength similarly grows with compute.

In addition, we find that scaling the loop count is more compute-efficient than scaling model size, even when weight decay is tuned for model-size scaling (Figure~\ref{fig:data-constr-main}, third panel). Here, scaling model size increases width and depth together at our fixed width--depth ratio. Retuning weight decay at every model size mitigates overfitting, but scaling the loop count at fixed model size and fixed weight decay matches the best tuned model-size-scaling loss with $2.2\times$ less compute. Moreover, weight-decay tuning on top of loop-count scaling reduces loss by at most 0.02, because the optimal weight decay of the \Kone model changes. This effect is visible in Figure~\ref{fig:data-constr-main} (fourth panel): the optimal weight decay rises with depth but is nearly flat in loop count, so the value tuned at one loop transfers to larger loop counts. Finally, untied looping adds the same depth with more parameters and does not beat \Kone (Figure~\ref{fig:data-constr-arch}), so weight sharing is an effective regularization technique in multi-epoch training.

We hypothesize that a model overfits with the parameters it stores, not with the blocks it executes. At small budgets the model is compute-limited, so a new parameter is preferred to a reused one, as on fresh data. Once the stored parameters begin to overfit the corpus, loops become the better way to add capacity, so the optimal loop count grows with compute. The last two panels of Figure~\ref{fig:data-constr-main} support the hypothesis directly: the marginal gains of looping are independent of the stored parameters and the optimal weight decay tracks stored parameters, not executed depth. We return to this picture in Section~\ref{sec:discussion} through the lens of computational depth.

\section{Computational Depth}
\label{sec:discussion}

Our motivation is \emph{computational depth}: for a fixed compute budget, we want a model to have as much usable depth as possible, since depth is what allows a network to compose many steps of computation. In particular, we define computational depth as the number of blocks that meaningfully influence the predictive distribution. Executed depth and computational depth need not coincide, however, and we see two ways in which compute spent on depth could go to waste. First, a block can execute without changing the prediction, which is the curse of depth \citep{sun2025curse}. Second, a block can consume compute budget throughout training even though the network may only need the block near the end. A waste of either kind would only change the exponent if it grew with scale. We hypothesize that both wastes take up a larger fraction of depth at larger depths, and the compute-optimal depth rises with budget, so both should grow with scale. Each of our interventions plausibly removes one of these wastes, which would explain why both change the exponent in Section~\ref{sec:ladder-results}: the boundary operator keeps every executed block contributing to the prediction, and model growth keeps the model shallow until the added depth is needed.

\begin{figure}[h]
\centering
\includegraphics[width=0.8\linewidth]{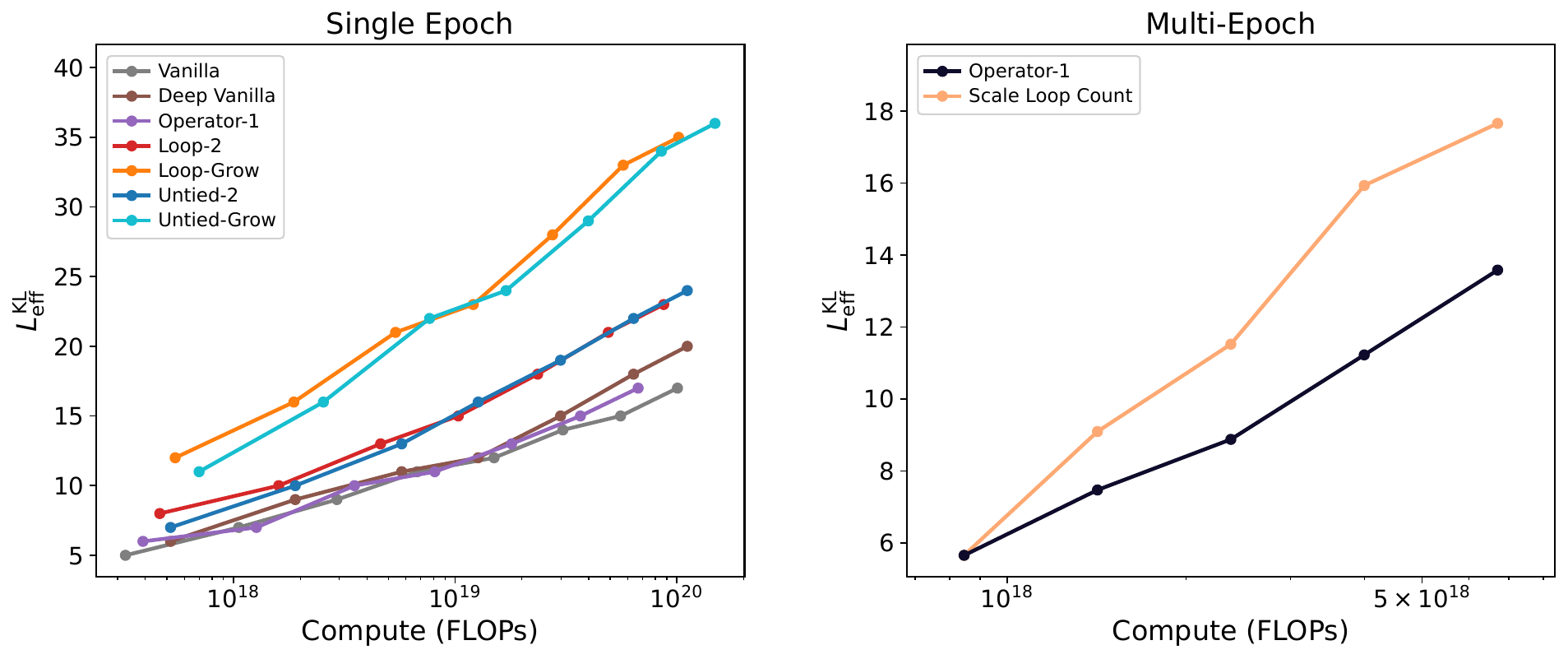}
\caption{\textbf{Growth and looping raise KL effective depth along the compute-optimal single-epoch ladders, and in multi-epoch training, scaling the loop count raises KL effective depth faster than scaling depth.} KL effective depth is measured with the logit lens: we decode the residual stream after each block and record the first block after the KL peak at which the decoded prediction is within 2 nats of the model's final output. \textbf{(Left)} Single-epoch scaling ladders. KL effective depth rises with compute for every family. Moreover, the curves group by depth multiplier: the one-pass models sit lowest, the two-pass models above them, and the models with model growth highest, reaching roughly twice Vanilla's KL effective depth at $10^{20}$ FLOPs. Within each group, the tied and untied curves coincide, so weight sharing does not change KL effective depth. Deep Vanilla, however, executes as many blocks as \Dep yet has a smaller KL effective depth (20 versus 24 at $10^{20}$), whereas \Kone lies on the Vanilla curve, so the boundary operator matters more at depth. \textbf{(Right)} Multi-epoch training on 100M unique tokens for 10 epochs, where each point is the best configuration at that budget (interpolated between neighboring checkpoints). Scaling the loop count at fixed weight decay gives a larger KL effective depth than scaling depth at $K=1$ with tuned weight decay, and the gap widens with compute.}
\label{fig:ladder-kl-eff-depth}
\end{figure}

\paragraph{The boundary operator increases computational depth.}
We hypothesize that the boundary operator improves the exponent because it closes the gap between executed and computational depth. In a pre-norm transformer the residual stream grows with depth, so a growing share of the blocks a token passes through does little to change it. Normalizing the stream and re-injecting the input lets every block write at full relative weight. To measure how much depth a model uses, we compute its \emph{KL effective depth} \citep{logit2020lens, csordas2026language}: we decode the residual stream after each block with the logit lens and record the first block after the KL peak at which the decoded prediction is within a KL threshold of the model's final output. KL effective depth is only a proxy for computational depth, since falling within this threshold does not mean later blocks stop changing the prediction, and unused blocks in the middle of the network go undetected. \Dep and Deep Vanilla execute the same blocks, but \Dep reaches a KL effective depth of $24$ against $20$ at $10^{20}$ FLOPs (Figure~\ref{fig:ladder-kl-eff-depth}, left), so the operator appears to improve computational depth. \Kone sits only slightly above Vanilla, so the operator changes little at the shallowest depths, and the gap widens as models get deeper, as the hypothesis predicts. Another evidence comes from width-only scaling. Scaling only the width, not the depth, turns the exponent improvement from \Kone to Vanilla to a constant one (Figure~\ref{fig:ladder-width-only}). This suggests depth scaling is necessary for the exponent improvement with the boundary operator.

\paragraph{Model growth adds depth when it is needed.}
We hypothesize that model growth improves the exponent because a fixed-depth model pays for depth it does not yet need. Networks fit simple structure early in training and more sophisticated structure only toward the end \citep{nakkiran2019sgd,rahaman2019spectral}, so a fixed-depth model may not need all of its blocks early in training. We further hypothesize that the fraction of its depth a model cannot yet use grows with depth. Since compute-optimal models get deeper with scale, the compute a fixed-depth model spends on such depth is then a growing fraction of the budget. Three observations support this hypothesis. Growth timing matters: the transition sweeps of Appendix~\ref{app:grow} have interior minima at fixed model size and budget, so paying for depth too early or too late both cost loss. Grown models prefer a smaller starting model trained on more tokens, with the optimal tokens per parameter rising from $6$ to $7$--$8$ (Table~\ref{tab:arch-optimal-tpp}), which is what the hypothesis predicts: spend most of the budget shallow and add depth late. Finally, \DepGrow reaches a KL effective depth of $36$ against $24$ for \Dep at $10^{20}$ FLOPs (Figure~\ref{fig:ladder-kl-eff-depth}, left), so growth appears to raise computational depth.

\paragraph{Looping adds depth without adding parameters to overfit.}
Under repeated data, we observe that adding depth through looping is better than adding parameters directly (Section~\ref{sec:data-constr}). We hypothesize that this is because a model overfits with the parameters it stores, not the blocks it executes, so tied looping adds depth without adding anything to overfit. The control supports this hypothesis: untied looping adds the same depth with more parameters and does not beat tuned \Kone (Appendix~\ref{app:data-const}), so weight sharing, not depth, is what helps under repetition. KL effective depth tells the same story: scaling the loop count raises KL effective depth faster than scaling depth at $K=1$ with tuned weight decay, reaching $18$ against $14$ layers at the largest budget (Figure~\ref{fig:ladder-kl-eff-depth}, right).

\section{Discussion}
\label{sec:conclusion}

For many years, \emph{data} interventions have been the main driver of advances in pretraining efficiency. By contrast, architectural innovations at pretraining have largely been absent, with the common belief that they can at best influence only scaling constants. However, we may be entering a new era, where the methodological landscape of research undergoes great change. We are now starting to see Muon challenge Adam as the default optimizer, after nearly a decade in which Adam dominated and hundreds of proposed alternatives never achieved broad adoption. Similarly, recursive depth, or looping, has recently been gaining mainstream traction for parameter-efficient representations, inference-time scaling, and reasoning. Moreover, as we become more data constrained, it will become increasingly natural to look to methodological interventions for further performance gains.

Contrary to the conventional wisdom, we have shown that architectural interventions at pretraining can influence scaling exponents. Both model growth, and even a simple boundary operator, 
provide compute efficiency gains that \emph{increase} with scale. Selecting for computational depth, reaching the largest usable depth for a given computational budget, can explain the effect of these interventions on scaling laws. Furthermore, in the data-constrained setting, where we train for multiple epochs, increasing the loop count with scale becomes compute-optimal. This finding may be particularly salient as data becomes a more constrained resource in the future. 

Going forward, proposing architectural interventions that increase computational depth could lead to further exponent improvements. 
Context length, the number of experts in a mixture-of-experts model, and width all grow with compute, yet all are fixed before training begins, so growing them on the schedule the network needs may change the exponent. Within depth itself, staged schedules (two passes, then four, then six) may increase the exponent further still. More broadly, the distinction between constant and exponent improvements deserves to be a standard part of how new architectures and training recipes are evaluated: a better constant saves the same factor of compute at every scale, while a better exponent saves a factor that compounds as budgets grow.

\vspace{5mm}
\textbf{Acknowledgements.} The work is partially inspired by the slowrun record \citet{slowrun26pr}. We thank Jonas Geiping and Shikai Qiu for helpful discussions.

\input{loop.bbl}
\newpage
\appendix

\setcounter{topnumber}{4}
\setcounter{bottomnumber}{4}
\setcounter{totalnumber}{6}
\renewcommand{\bottomfraction}{0.9}
\renewcommand{\floatpagefraction}{0.75}
\setlength{\textfloatsep}{12pt plus 2pt minus 2pt}
\setlength{\floatsep}{10pt plus 2pt minus 2pt}
\setlength{\intextsep}{10pt plus 2pt minus 2pt}
\makeatletter
\setlength{\@fptop}{0pt}
\setlength{\@fpsep}{10pt plus 2pt minus 2pt}
\setlength{\@fpbot}{0pt plus 1fil}
\makeatother
\raggedbottom

\section*{Appendix Outline}

The appendix provides the experimental details, derivations, and additional results supporting the main text.

\noindent\textbf{Appendix~\ref{app:comp-opt}} describes the common training setup, hardware, and architecture definitions for our compute-optimal experiments. It details the choices of block allocation, core-pass count, growth target, and growth timing, and derives the relationships between recurrence, compute, and tokens per parameter. It also gives the full recipe-fitting procedure: tuning a base model, selecting the token allocation and growth fraction, fitting the learning-rate scaling rule, and training the scaling ladders. The settings used for each architecture are listed in Table~\ref{tab:full-ladder-recipes}.

\noindent\textbf{Appendix~\ref{app:comp-opt-results}} presents additional compute-optimal scaling results and ablations. These isolate the components of the boundary operator, test block allocation, and compare efficiency in FLOPs and training time. Further experiments examine architecture-specific hyperparameter tuning, model shape, width-only scaling, random recurrence, test-time core passes, and optimizer choice, showing how these decisions affect constant-factor and exponent improvements.

\noindent\textbf{Appendix~\ref{app:transfer-evaluation}} examines corpus transfer and downstream performance. It compares scaling on FineWeb and FineWeb-Edu, specifies the CORE accuracy and answer-NLL evaluation protocols, and analyzes downstream scaling and task-level differences at matched pretraining loss. It then describes the taskwise calibration used to forecast CORE accuracy and evaluate the held-out large-model extrapolation, together with the scope and limitations of these comparisons.

\noindent\textbf{Appendix~\ref{app:data-const}} extends the data-constrained experiments across data-repetition levels, weight decay, and architectural controls. It examines how repetition changes the preferred number of core passes, how weight decay and recurrence interact, and how hyperparameters transfer across recurrence counts. Comparisons of tied and untied models, with and without the boundary operator, distinguish the roles of weight sharing and additional depth.
\newpage

\section{Experiment Details for Compute-Optimal Scaling}
\label{app:comp-opt}

We explain the training protocols we use for Section~\ref{sec:compute-optimal} and ~\ref{sec:data-constr}. The common setup is in Table~\ref{tab:common-setup}.

\paragraph{Hardware.}
Most runs use one node with eight H100 GPUs. The largest $d26$ extrapolation run uses two such nodes.

\begin{table}[!htbp]
\centering
\small
\begin{tabular}{lp{0.72\linewidth}}
\hline
Setting & Protocol \\
\hline
Context length & $2048$ tokens \\
Tokenizer & GPT-2 vocabulary of $50{,}257$, padded to $50{,}304$ \\
Global batch & $524{,}288$ tokens \\
Shape rule & Model name $d\ell$ denotes nominal/reference depth $\ell$ and width $w=128\ell$ \\
Optimizer & Muon for matrix parameters; AdamW for embeddings and the language-model head. \\
\hline
\end{tabular}
\caption{Fixed architecture and training protocol. Architecture-specific hyperparameters and token budgets are reported in the following subsections.}
\label{tab:common-setup}
\end{table}

\subsection{Architecture}
\label{app:architecture}

All models use the pre-norm decoder-only transformer, with RoPE, SwiGLU, and QK normalization. We use no biases or learned normalization gains, so trainable matrices are all two-dimensional. Additional RMS normalizations follow the token embedding and precede \texttt{lm\_head}. Attention and MLP output projections and \texttt{lm\_head} are initialized to zero; the token embedding is initialized normally, while the attention Q/K/V and MLP input matrices use uniform initialization.

Following \citet{nanochat}, $d\ell$ names a model by nominal/reference depth $\ell$. Vanilla is an unsplit dense stack, so its stored and executed depths are both $\ell$. The architectures are defined in Section~\ref{sec:arch}. Parameters for different architectures and reference depths are shown in Table~\ref{tab:model-sizes}.

\begin{table}[!htbp]
\centering
\small
\setlength{\tabcolsep}{4pt}
\begin{tabular}{lrrrrrrr}
\hline
Depths & P/C/D & Exec. Depth & Width & Vanilla & Loop-Grow & Untied-2 & Untied-Grow \\
\hline
$d6$  & $2/2/2$ & $8\to12$  & $768$     & $120\,\mathrm{M}$ & $120\,\mathrm{M}$ & $130\,\mathrm{M}$ & $160\,\mathrm{M}$ \\
$d8$  & $2/3/3$ & $11\to17$ & $1{,}024$ & $210\,\mathrm{M}$ & $210\,\mathrm{M}$ & $240\,\mathrm{M}$ & $320\,\mathrm{M}$ \\
$d10$ & $3/4/3$ & $14\to22$ & $1{,}280$ & $330\,\mathrm{M}$ & $330\,\mathrm{M}$ & $410\,\mathrm{M}$ & $580\,\mathrm{M}$ \\
$d12$ & $4/4/4$ & $16\to24$ & $1{,}536$ & $490\,\mathrm{M}$ & $490\,\mathrm{M}$ & $610\,\mathrm{M}$ & $830\,\mathrm{M}$ \\
$d14$ & $4/5/5$ & $19\to29$ & $1{,}792$ & $730\,\mathrm{M}$ & $730\,\mathrm{M}$ & $920\,\mathrm{M}$ & $1.3\,\mathrm{B}$ \\
$d16$ & $5/6/5$ & $22\to34$ & $2{,}048$ & $1.0\,\mathrm{B}$ & $1.0\,\mathrm{B}$ & $1.3\,\mathrm{B}$ & $2.0\,\mathrm{B}$ \\
$d18$ & $6/6/6$ & $24\to36$ & $2{,}304$ & $1.4\,\mathrm{B}$ & $1.4\,\mathrm{B}$ & $1.8\,\mathrm{B}$ & $2.5\,\mathrm{B}$ \\
$d20$ & $6/7/7$ & $27\to41$ & $2{,}560$ & $1.8\,\mathrm{B}$ & $1.8\,\mathrm{B}$ & $2.4\,\mathrm{B}$ & $3.5\,\mathrm{B}$ \\
$d22$ & $7/8/7$ & $30\to46$ & $2{,}816$ & $2.4\,\mathrm{B}$ & $2.4\,\mathrm{B}$ & $3.2\,\mathrm{B}$ & $4.7\,\mathrm{B}$ \\
$d24$ & $8/8/8$ & $32\to48$ & $3{,}072$ & $3.0\,\mathrm{B}$ & $3.0\,\mathrm{B}$ & $3.9\,\mathrm{B}$ & $5.7\,\mathrm{B}$ \\
$d26$ & $8/9/9$ & $35\to53$ & $3{,}328$ & $3.8\,\mathrm{B}$ & $3.8\,\mathrm{B}$ & $5.0\,\mathrm{B}$ & $7.4\,\mathrm{B}$ \\
\hline
\end{tabular}
\caption{\textbf{Model depths, layer splits, executed depths, and stored parameters.} P/C/D gives the prelude/core/coda layer split. Executed depth shows the transition from $K=2$ to $K=4$; fixed Loop-2 and Untied-2 use the first value and \KtwoGrow and \DepGrow use the second. Vanilla has executed depths equal to physical depths. Operator-1 and Loop-2 have the same stored parameters as Vanilla, while Loop-2 has the same effective parameters as Untied-2. Loop-Grow shares the Vanilla stored parameters, whereas Untied-Grow allocates all four untied core copies from the start.}
\label{tab:model-sizes}
\end{table}

\paragraph{Compute budget estimation}
Plotted compute comes from the model FLOP estimator, which includes matrix multiplications and attention computation; expressions of the form $6TN_{\eff}$ below are leading-order allocation identities.

\paragraph{Hardware consideration}
Operator-1 differs from Vanilla only with the boundary operator, adding one RMS Norm and vector additions.  Untied-2 and Loop-2 have the same computational graph. The added arithmetic and memory communication is relatively small. For a runtime comparison, see Figure~\ref{fig:ladder-runtime}.

\paragraph{Multipliers}
Suppose $\Theta$ is parameter and $F$ is one layer (e.g. MLP, attention). Multiplier $\alpha$ for $\Theta$ is defined as $F(\alpha \Theta)$. Multiplier is not trainable. Multipliers follow an equivalence relationship with learning rate and initialization scales, known as ABC-parameterization \citep{yang2022tensorprogramsvtuning}. In this paper, we define two multipliers of interest. Output multiplier (OM) is the multiplier to the unembedding or readout layer. Residual multiplier (RM) is the shared weight multiplier to the MLP down projection and attention output projection.

\input{sections/looping_decisions}

\input{sections/tpp_derivations}

\subsection{Compute-Optimal Recipe: Staged Sweeps and Fits}
\label{app:recipe-details}

The recipe follows the four stages of Section~\ref{sec:recipe}: tune base hyperparameters at $d8$ (Stage 1), fit tokens per stored parameter (Stage 2), choose a fixed growth fraction $\rho$ for growth variants (Stage 3), and fit the learning-rate scaling rule (Stage 4). Fixed-recurrence models skip Stage 3. The block allocation, recurrence count, and growth target are chosen as described in Appendix~\ref{app:family-design}. For growth, we recommend retaining the Stage 2 TPP and calibrating $\rho$ once: the sensitivity tests in Appendix~\ref{app:growth-sensitivity} show little benefit from retuning TPP or refitting the transition across sizes.

\subsubsection{Stage 1: Tune a Base Model}
\label{app:base-tuning}
\begin{figure}[!htbp]
\centering

\begin{minipage}[t]{0.52\linewidth}
\vspace{0pt}
\centering
\includegraphics[width=\linewidth]{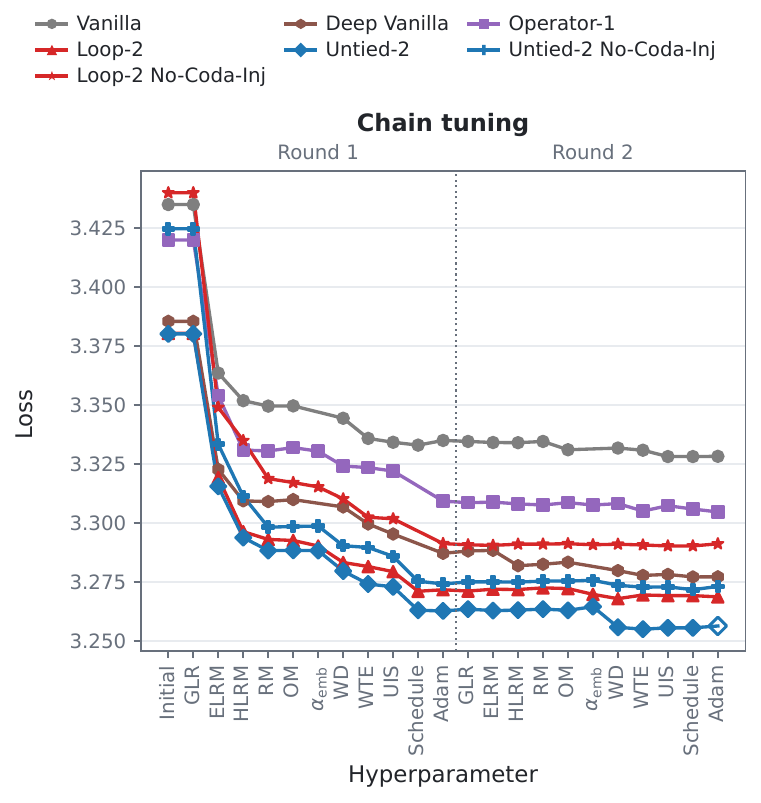}
\captionof{figure}{\textbf{Chain tuning at $d8$ on 1B tokens.}
Each curve is one architecture's tuning chain. We adopt the new hyperparamters if it's $10^{-3}$ better than the current best. Loss continuously drops with more tuning. ELRM, HLRM, WD, Schedule, WTE, Adam bring most drops. The second round still improves loss at some HP because the optimum changes as other HPs change. The loss plateaus near the end of second round. The hollow marker denotes that more than one discrete HPs improve the performance and stacking both can potentially improve performance more.}
\label{fig:base-tuning}
\end{minipage}
\hfill
\begin{minipage}[t]{0.45\linewidth}
\vspace{0pt}
\centering
\scriptsize
\setlength{\tabcolsep}{2.5pt}
\renewcommand{\arraystretch}{0.95}

\begin{tabularx}{\linewidth}{lXr}
\toprule
HP & Definition & Initial HP \\
\midrule

\multicolumn{2}{l}{\emph{$\times\{1/16,1/8,\ldots,8\}$}} & \\
GLR & global learning rate & $0.04$ \\
RM & residual multiplier & $0.5$ \\
OM & output multiplier & $1$ \\
$\alpha_{\rm emb}$ & injection weight & $1$ \\

\midrule
\multicolumn{2}{l}{\emph{$\times\{1/8,1/4,\ldots,16\}$}} & \\
WTE init. & embedding init.\ scale & $0.08$ \\
UIS & input-matrix init.\ scale & $0.25$ \\

\midrule
\multicolumn{2}{l}{\emph{$\times\{1,2,\ldots,128\}$}} & \\
ELRM & embedding LR multiplier & $0.02$ \\
HLRM & head LR multiplier & $0.02$ \\

\midrule
\multicolumn{2}{l}{\emph{$\{0\}\cup\times\{1/32,\ldots,2\}$}} & \\
WD & weight decay & $0.1$ \\

\midrule
\multicolumn{2}{l}{\emph{Discrete}} & \\

Schedule &
warmup $\{0,5,10,20\}$;\newline
warmdown $\{.2,.6,.8,1\}$ &
\begin{tabular}[t]{@{}r@{}}
$40$;\\
$.4$
\end{tabular}
\\[1pt]

Adam &
$\beta_1\{.9,.95\}$;\newline
$\beta_2\{.90,.98,.99\}$;\newline
$\epsilon\{10^{-8},10^{-6}\}$ &
\begin{tabular}[t]{@{}r@{}}
$(.8,.95)$;\\
$10^{-10}$
\end{tabular}
\\

\bottomrule
\end{tabularx}

\captionof{table}{\textbf{Tuning Grid for Chain Tuning.}
There's nine sweeps (eight for Vanilla) in each round. Each sweep has eight parallel runs. For numerical values, grid is centered at the optimal values and span 2x grid in the first round and $\sqrt{2}$x grid in the second round. For discrete values, there are two or three HPs. We change one HP in each run. If more than one HPs are better than the baseline, we apply both HPs in the next sweep. Discrete sweeps use the same candidates in both rounds.}
\label{tab:base-tuning-grid}
\end{minipage}

\end{figure}

We first tune every hyperparameter at base size, $d8$, on a fixed budget of 1B tokens. We sweep one hyperparameter at a time and hold the local optimum into next hyperparameter sweep, similar to \citep{wen2026fantastic}. We show the hyperparameter grids in Table~\ref{tab:base-tuning-grid}. Partial tuning trajectories are shown in Figure~\ref{fig:base-tuning}.

Table~\ref{tab:base-tuning-operators} lists the tuned losses of compute-matched arms and justifies our architecture choices. The final mixing operator $\rho$, which is missing in previous looped architectures, improves the loss by a fair amount. In addition, we find that untying the weights have a benefits over tying.

Tuning each architecture separately matters: transferring Vanilla's tuned recipe to the looped variants costs $7$--$26\times10^{-3}$ in loss (Table~\ref{tab:base-tuning-transfer}), emphasizing the importance of separate tuning for each architecture.

\begin{table}[!htbp]
\centering
\small
\begin{tabular}{@{}lccrr@{}}
\toprule
Architecture & $\phi$ (core boundary) & $\rho$ (before coda) & Tuned loss & Runtime (min) \\
\midrule
Deep Vanilla & $h$ & $h$ & 3.2772 & 6.49 \\
Loop-2 & $\mathrm{BO}(h,e)$ & $\mathrm{BO}(h,e)$ & 3.2704 & 6.54 \\
Loop-2-no-coda-inj & $\mathrm{BO}(h,e)$ & $\mathrm{Norm}(h)$ & 3.2912 & 6.45 \\
Untied-2 & $\mathrm{BO}(h,e)$ & $\mathrm{BO}(h,e)$ & \textbf{3.2563} & 6.54 \\
Untied-2-no-coda-inj & $\mathrm{BO}(h,e)$ & $\mathrm{Norm}(h)$ & 3.2731 & 6.53 \\
Deep Vanilla + norm & $\mathrm{Norm}(h)$ & $\mathrm{Norm}(h)$ & 3.2781 & 6.48 \\
Deep Vanilla + injection & $h+\alpha e$ & $h+\alpha e$ & 3.2690 & 6.52 \\
\bottomrule
\end{tabular}
\caption{\textbf{Boundary-operator ablations at the base size.} Architectures and boundary maps follow Equation~\ref{eq:looped-general} and Equation~\ref{eq:boundary-operator}. Each model is tuned independently and trained on 1B tokens at width $1024$ and executed depth $11$. Untied-2 achieves the lowest loss; removing either BO component or the coda injection increases loss. Runtime is the corresponding run's logged training time in minutes on eight H100 GPUs, excluding evaluation and initial compilation/warmup.}
\label{tab:base-tuning-operators}
\end{table}

\begin{table}[!htbp]
\centering
\scriptsize
\setlength{\tabcolsep}{3pt}
\resizebox{\linewidth}{!}{%
\begin{tabular}{@{}l*{14}{c}@{}}
\toprule
Architecture & GLR & ELRM & HLRM & RM & OM & $\alpha_{\rm emb}$ & WD & WTE & UIS & WU & WDR & $\beta_1$ & $\beta_2$ & $\epsilon$ \\
\midrule
Vanilla      & 0.04 & 0.453 & 0.113 & 0.25 & 0.5 & --- & 0.071 & 0.007 & 0.063 & 40 & 0.6 & 0.8 & 0.95 & $10^{-10}$ \\
Deep Vanilla & 0.04 & 0.16  & 0.057 & 0.5  & 1   & --- & 0.1   & 0.005 & 0.5   & 5  & 0.8 & 0.8 & 0.99 & $10^{-8}$  \\
Operator-1     & 0.04 & 0.905 & 0.08  & 0.5  & 1   & 1   & 0.05  & 0.113 & 0.354 & 0  & 0.8 & 0.8 & 0.98 & $10^{-10}$ \\
Loop-2       & 0.04 & 0.32  & 0.113 & 0.25 & 1   & 0.707 & 0.05 & 0.02 & 0.044 & 40 & 1 & 0.8 & 0.95 & $10^{-10}$ \\
Untied-2     & 0.04 & 0.16  & 0.16  & 0.25 & 1   & 1   & 0.071 & 0.01  & 0.354 & 40 & 1 & 0.8 & 0.99 & $10^{-8}$  \\
\bottomrule
\end{tabular}%
}
\caption{\textbf{Base-tuned hyperparameters by architecture.} Recipes are selected by two rounds of chain tuning at the base size on 1B tokens. Dashes indicate inapplicable hyperparameters.}
\label{tab:tuned-hyperparameters}
\end{table}

\begin{table}[!htbp]
\centering
\small
\begin{adjustbox}{max width=\linewidth}
\begin{tabular}{lrrrr}
\toprule
Architecture & Vanilla-recipe loss & $\Delta$ vs. Vanilla ($10^{-3}$) & Own-recipe loss & Transfer regret ($10^{-3}$) \\
\midrule
Vanilla & 3.3275 & $+0.0$ & 3.3279 & $-0.3$ \\
Deep Vanilla & 3.2869 & $-40.6$ & 3.2772 & $+9.8$ \\
Operator-1 & 3.3135 & $-14.1$ & 3.3057 & $+7.8$ \\
Loop-2 & 3.2777 & $-49.9$ & 3.2704 & $+7.2$ \\
Untied-2 & 3.2697 & $-57.9$ & 3.2563 & $+13.3$ \\
\bottomrule
\end{tabular}
\end{adjustbox}
\caption{\textbf{Transfer probe from the Vanilla recipe.} Each available transfer probe is trained once at $d8$ on 1B tokens with Vanilla's tuned recipe (injection variants keep the default $\alpha_{\text{emb}}$) and compared with its own tuned recipe. Transfer regret is the transferred-recipe loss minus the own-recipe loss, so positive values measure the benefit of variant-specific tuning; deltas and regrets use unrounded losses. Dashes indicate unavailable transfer measurements for Deep Vanilla, whose own-recipe loss is reported at matched executed depth $11$ and width $1024$. Vanilla's $-0.3\times10^{-3}$ is below the adoption threshold and reflects run-to-run noise.}
\label{tab:base-tuning-transfer}
\end{table}

\subsubsection{Stage 2: Fit the Optimal Tokens per Parameter}
\label{app:tpp-fitting}
With the base recipe frozen, we fit the split between model size and training tokens at fixed compute \citep{hoffmann2022training}. We express this split as tokens per stored parameter, $\mathrm{TPP}=T/N$. Compute depends instead on the compute-active count $N_c$, which counts a shared core once per application (Appendix~\ref{app:recurrence-choice}). At the same anchor size and recurrence, Loop-2 and Untied-2 have the same compute per token, but Untied-2 stores more parameters; its stored-parameter TPP is therefore lower at matched compute. We use stored-parameter TPP throughout this stage.

\begin{figure}[!htbp]
\centering
\includegraphics[width=\linewidth]{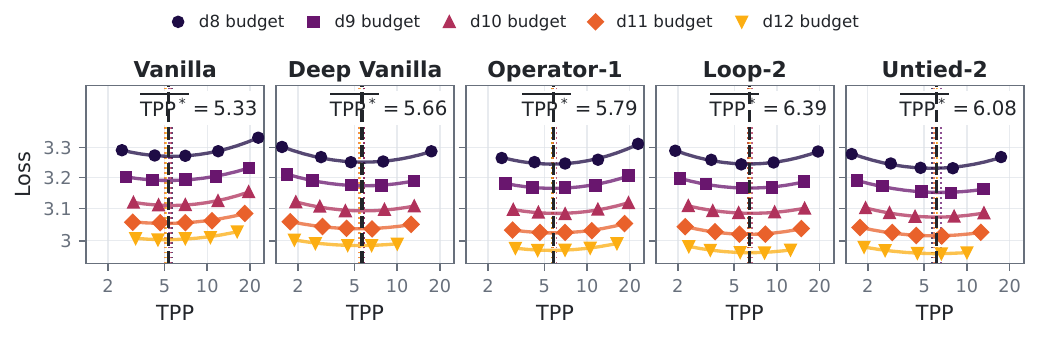}
\caption{\textbf{Iso-compute TPP fits and fitted scaling laws.}  Loss against tokens per parameter for Vanilla, Loop-2, and Untied-2 and a scaling law from fitted vertices. Each color is one fixed compute budget, the cost of the $d8$--$d12$ model at TPP $7$; curves are quadratic fits of log loss against log TPP, dotted lines mark the per-budget vertices, and the dashed line is their mean. Per-budget vertices for the three architectures are plotted in the last subfigure as a scaling law.}
\label{fig:tpp-curves}
\end{figure}

\textbf{Iso-compute sweeps.} We name each budget by the reference model size whose training at TPP $7$ defines its compute cost. We use five budgets, $d8$ through $d12$. At each budget, we train five model sizes around the anchor and adjust the token count to match compute. We fit a quadratic in log loss against log TPP and take its vertex as the budget's optimal TPP (Figure~\ref{fig:tpp-curves}). The vertex losses describe the fitted frontier under the base recipe. We defer scaling-law comparisons to the completed ladders, after fitting growth timing and learning-rate scaling.

\textbf{Is the optimal TPP stable across scale?} \citet{hoffmann2022training} found a roughly constant optimal TPP across compute, and we test the same property for each architecture. Table~\ref{tab:arch-optimal-tpp} reports the mean of the five vertices together with a scale-shift test. No variant rejects a scale-invariant TPP at $\alpha=0.05$. We adopt the rounded mean in Table~\ref{tab:arch-optimal-tpp}: $5$ for Vanilla and $6$ for Operator-1, Loop-2, and Untied-2. Every looped variant prefers a slightly higher TPP than Vanilla, indicating better parameter efficiency.

\begin{table}[!htbp]
\centering
\begin{tabular}{lrrrrrr}
\toprule
Architecture & Mean TPP$^\star$ & Rounded & $p$ & $L^\star_{d8}$ & $L^\star_{d10}$ & $L^\star_{d12}$ \\
\midrule
Vanilla & 5.33 & 5 & 0.087 & 3.270 & 3.110 & 3.003 \\
Deep Vanilla & 5.66 & 6 & 0.109 & 3.251 & 3.093 & 2.985 \\
Operator-1 & 5.79 & 6 & 0.923 & 3.244 & 3.085 & 2.971 \\ %
Loop-2 & 6.39 & 6 & 0.588 & 3.244 & 3.085 & 2.964 \\
Untied-2 & 6.08 & 6 & 0.760 & \textbf{3.230} & \textbf{3.073} & \textbf{2.961} \\
\bottomrule
\end{tabular}
\captionof{table}{\textbf{Compute-optimal TPP by architecture.} Mean TPP$^\star$ averages the five iso-compute vertices; rounded TPP is the value used in the scaling ladders. The loss columns report the fitted vertex loss at budgets d8, d10, and d12; the best loss in each budget is bold. The scale-shift test fits $\log\mathrm{TPP}^\star=a+\beta\log C$ over all five budgets and reports the two-sided $p$-value of $H_0\!:\beta=0$. No variant rejects a scale-invariant TPP at $\alpha=0.05$.}
\label{tab:arch-optimal-tpp}
\end{table}

\subsubsection{Stage 3: Choose a Fixed Growth Fraction}
\label{app:growth-fitting}

\begin{figure}[!htbp]
    \centering
    \includegraphics[width=\linewidth]{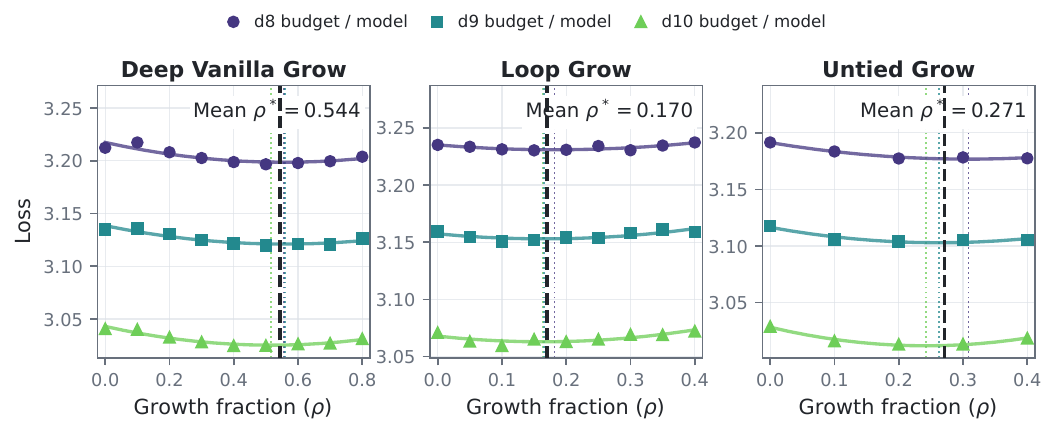}
    \caption{\textbf{Choosing a fixed growth fraction.} Left to right: Deep Vanilla Grow, Loop Grow, and Untied Grow. Each series shows validation loss versus the post-growth token fraction $\rho$ for a $d8$, $d9$, or $d10$ model at its corresponding fixed-compute budget; tokens are adjusted as $\rho$ varies. Points are measured runs and curves are quadratic fits. Colored dotted lines mark the fitted optima, and black dashed lines mark their arithmetic means: $0.544$, $0.170$, and $0.271$, respectively. The panels use different axis ranges.}
    \label{fig:opt-grow-fits}
\end{figure}

For the $2\to4$ growth schedule, let $\rho$ denote the fraction of training tokens processed \emph{after} growth; the transition therefore occurs after fraction $1-\rho$. We recommend reusing the Stage 2 TPP to set the reference compute budget and choosing one fixed $\rho$ for the ladder. At each of three small-model anchors ($d8$--$d10$), sweep $\rho$ while adjusting tokens to preserve that anchor's compute budget, then fit a quadratic to validation loss versus $\rho$ (Figure~\ref{fig:opt-grow-fits}). Average the three fitted optima and round to one decimal place. The broad minima support this simple choice without a growth-specific TPP refit or per-size transition rule.

We use this recipe for Deep Vanilla Grow: the mean optimum $0.544$ gives $\rho=0.5$, with the Deep Vanilla TPP-$6$ reference budget. The reported Untied Grow and Loop Grow ladders instead use fitted per-size fractions from Appendix~\ref{app:grow} and refitted reference TPP values of $8$ and $7$, respectively (Table~\ref{tab:full-ladder-recipes}). We expect little benefit from these refinements: the Untied Grow ablation finds less than about $5\%$ change in matched-loss compute when reusing TPP $6$, and loss changes of $-0.0009$ to $+0.0019$ when replacing the fitted fractions with $\rho=0.30$ (Figure~\ref{fig:opt-grow-regret}, right).

\subsubsection{Stage 4: Fit the Learning-Rate Rule}
\label{sec:glr-rule}

\begin{figure}[!htbp]
\centering
\begin{subfigure}[t]{\linewidth}
\centering
\includegraphics[width=\linewidth]{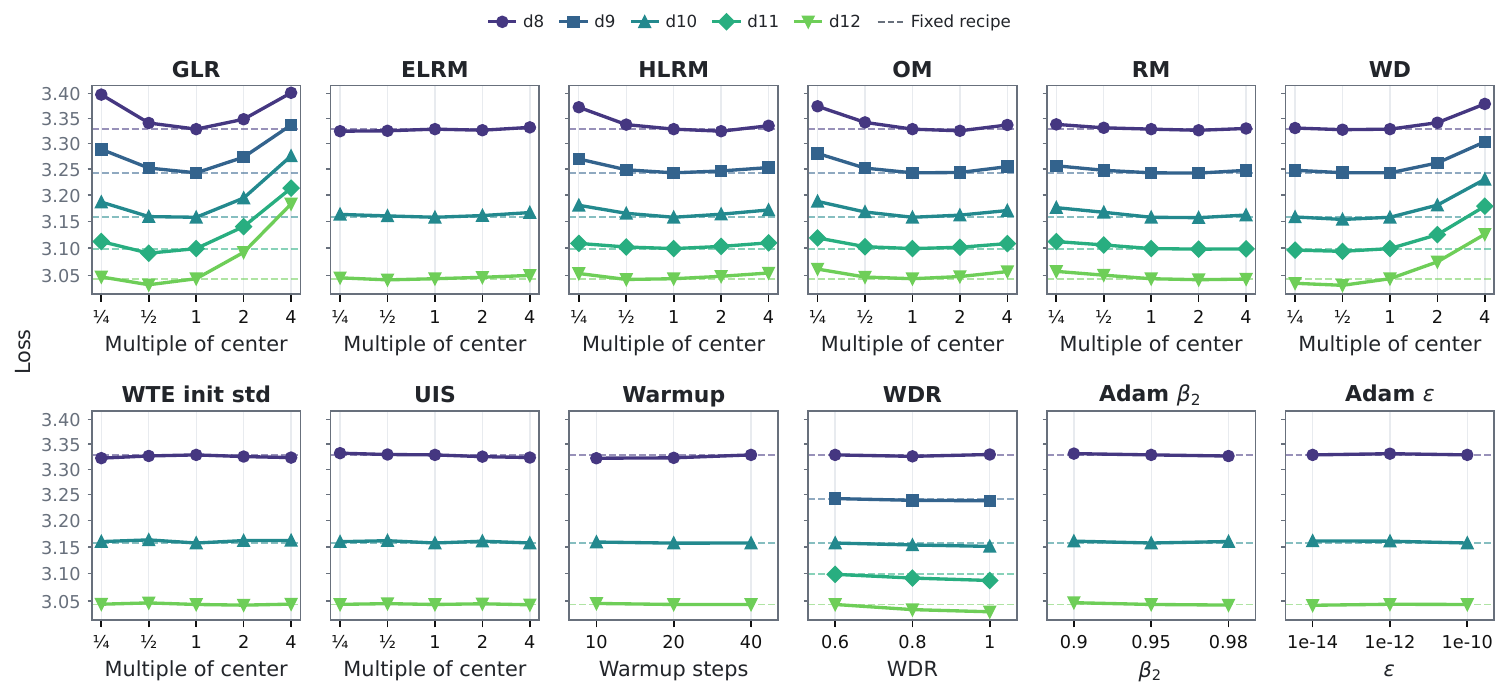}
\caption{Vanilla hyperparameter sensitivity across scale}
\label{fig:hyper1d}
\end{subfigure}

\medskip
\begin{subfigure}[t]{\linewidth}
\centering
\includegraphics[width=\linewidth]{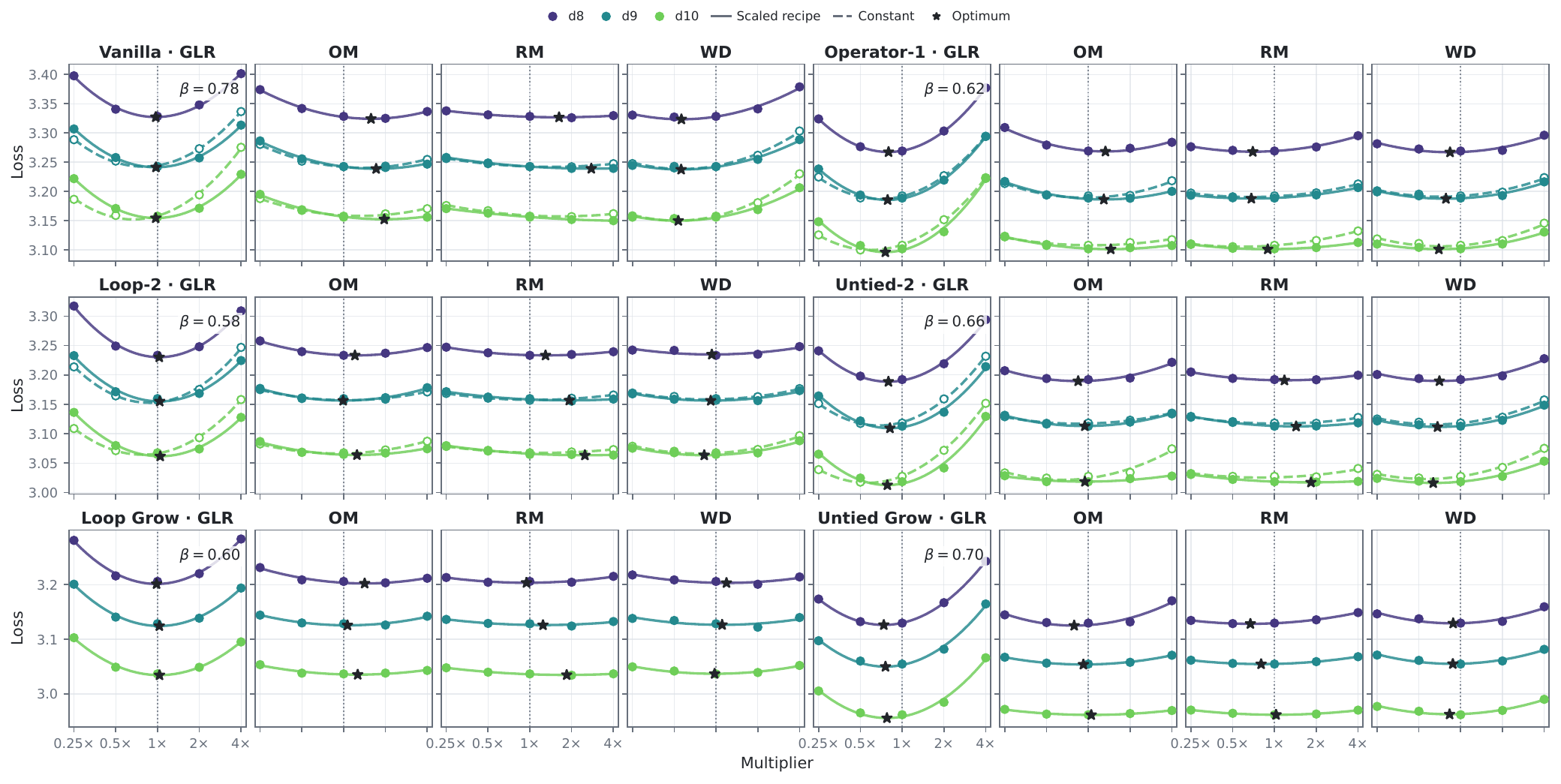}
\caption{Architecture-specific sweeps under constant and scaled recipes}
\label{fig:hp-slice}
\end{subfigure}
\caption{\textbf{Hyperparameter sensitivity and transfer across scale.} \textbf{(a)} One-dimensional Vanilla sweeps at $d8$--$d12$; dashed lines show the fixed-recipe loss. Scalar settings use $\{1/4,1/2,1,2,4\}$ times the base value; schedule and Adam settings use the displayed grids. ELRM and HLRM are embedding and head learning-rate ratios, WTE init std is the embedding initialization standard deviation, UIS is the uniform initialization scale, and WDR is the warmdown ratio. \textbf{(b)} Sweeps of global learning rate (GLR), output multiplier (OM), residual multiplier (RM), and weight decay (WD) at $d8$--$d10$. Rows pair Vanilla with \Kone, \Ktwo with \Dep, and \KtwoGrow with \DepGrow. Solid and dashed curves are quadratic fits for scaled and constant recipes; stars mark scaled-recipe optima. Positive $\beta$ annotations use $\mathrm{GLR}\propto N^{-\beta}$, opposite to the signed convention in Equation~\ref{eq:glr-scaling}. Fixed-recurrence panels report constant-recipe fits; growth panels show inherited rule magnitudes.}
\label{fig:hp-sensitivity-sweeps}
\end{figure}

With token allocation and growth timing fixed, we determine which base hyperparameters need to change with model size. We first screen Vanilla with one-dimensional sweeps over $d8$--$d12$, using the Stage 2 token allocation at each size (Figure~\ref{fig:hyper1d}). All hyperparameters are swept at $d8$, $d10$, and $d12$; the intermediate sizes cover the global learning rate (GLR), head learning-rate ratio, output and residual multipliers, weight decay, and warmdown ratio. Scalar settings use $\{1/4,1/2,1,2,4\}$ times their base values, while schedule and Adam settings use discrete grids. The preferred GLR falls from $0.04$ at $d8$--$d10$ to $0.02$ at $d11$--$d12$, whereas several other settings have nearly flat loss curves.

We then sweep the global learning rate (GLR), output multiplier (OM), residual multiplier (RM), and weight decay (WD) at $d8$--$d10$ for the architecture families (Figure~\ref{fig:hp-slice}). For each hyperparameter and size, we fit a quadratic to log validation loss against the log multiplier and take its minimum as the estimated optimum. Regressing the log optima against log stored parameter count gives the drift exponent $\beta$ in Table~\ref{tab:hp-drift}. The table also reports the loss penalty for a factor-of-two change and the full loss range of each sweep. Among these four hyperparameters, GLR has the largest sensitivity under the constant recipe. We therefore use the anchored learning-rate rule
\begin{equation}
\mathrm{GLR}(N)=\mathrm{GLR}_{d8}\left(\frac{N}{N_{d8}}\right)^{\beta},
\label{eq:glr-scaling}
\end{equation}
where $N$ is the initial stored parameter count and $N_{d8}$ is its value for that architecture's base model. The exponent is negative, so the learning rate decreases as models grow. For growth variants, we center the learning-rate sweeps on the corresponding fixed-recurrence rule and measure the remaining size dependence. The exponents and base learning rates used in the reported ladders are listed in Table~\ref{tab:full-ladder-recipes}; the exponents are specified to one decimal place.

We repeat the hyperparameter sweeps around the scaled recipe to check transfer across sizes (Figure~\ref{fig:hp-slice} and Table~\ref{tab:hp-drift}). Scaling GLR also changes the preferred values of the other hyperparameters, so drift measured under a constant recipe need not imply that an additional scaling rule is useful. In the Vanilla ladder ablation, adding output-multiplier or weight-decay scaling changes the matched-loss compute multiplier by at most about $3\%$ relative to GLR scaling alone (Figure~\ref{fig:ladder-vanilla-scaling}). We therefore use the fitted GLR rule and keep the other hyperparameters at their base-tuned values. This gives a simple recipe for comparing architectures without carrying a separate scaling law for every hyperparameter.

\begin{table}[!htbp]
\centering
\small
\setlength{\tabcolsep}{4pt}
\begin{adjustbox}{max width=\linewidth}
\begin{tabular}{llcrcrcrcr}
\toprule
& & \multicolumn{2}{c}{GLR} & \multicolumn{2}{c}{OM} & \multicolumn{2}{c}{RM} & \multicolumn{2}{c}{WD} \\
\cmidrule(lr){3-4}\cmidrule(lr){5-6}\cmidrule(lr){7-8}\cmidrule(lr){9-10}
Variant & Recipe & $\beta$ & \multicolumn{1}{c}{regret / range} & $\beta$ & \multicolumn{1}{c}{regret / range} & $\beta$ & \multicolumn{1}{c}{regret / range} & $\beta$ & \multicolumn{1}{c}{regret / range} \\
\midrule
Vanilla & constant & {\boldmath$-0.78\pm0.06$} & 18.7 / 150.0 & {\boldmath$-0.37\pm0.10$} & 3.5 / 17.6 & $+0.06\pm0.37$ & 2.3 / 15.2 & $-0.17\pm0.11$ & 9.4 / 94.0 \\
 & $+$GLR ($\beta{=}{-}0.8$) & $-0.00\pm0.01$ & 15.0 / 73.3 & {\boldmath$+0.46\pm0.05$} & 5.9 / 39.8 & {\boldmath$+2.73\pm0.23$} & 1.4 / 20.7 & $-0.11\pm0.05$ & 4.9 / 53.3 \\
\midrule
\Kone & constant & {\boldmath$-0.62\pm0.02$} & 17.9 / 122.4 & $-0.43\pm0.55$ & 3.0 / 14.4 & $-0.41\pm0.15$ & 3.0 / 26.9 & $-0.32\pm0.20$ & 5.6 / 37.8 \\
 & $+$GLR ($\beta{=}{-}0.6$) & {\boldmath$-0.11\pm0.02$} & 17.4 / 121.6 & $+0.19\pm0.17$ & 3.8 / 21.5 & $+0.53\pm0.35$ & 1.4 / 10.8 & {\boldmath$-0.39\pm0.05$} & 5.4 / 28.6 \\
\midrule
\Ktwo & constant & {\boldmath$-0.58\pm0.08$} & 16.8 / 110.1 & {\boldmath$-0.60\pm0.04$} & 2.3 / 14.9 & $-0.09\pm0.38$ & 2.1 / 11.6 & {\boldmath$-0.40\pm0.06$} & 7.0 / 41.1 \\
 & $+$GLR ($\beta{=}{-}0.6$) & {\boldmath$+0.03\pm0.00$} & 12.2 / 71.5 & $+0.08\pm0.51$ & 3.0 / 21.9 & {\boldmath$+1.36\pm0.19$} & 3.4 / 16.1 & $-0.26\pm0.10$ & 2.7 / 23.2 \\
\KtwoGrow & $+$GLR ($\beta{=}{-}0.6$) & $+0.10\pm0.05$ & 12.2 / 66.3 & $-0.23\pm0.55$ & 1.4 / 16.9 & {\boldmath$+1.40\pm0.12$} & 0.4 / 13.6 & {\boldmath$-0.41\pm0.06$} & 3.9 / 15.5 \\
\midrule
\Dep & constant & {\boldmath$-0.66\pm0.02$} & 17.3 / 134.1 & $-0.53\pm0.26$ & 1.9 / 49.7 & {\boldmath$-0.73\pm0.11$} & $-0.4$ / 14.2 & {\boldmath$-0.43\pm0.02$} & 6.2 / 50.4 \\
 & $+$GLR ($\beta{=}{-}0.7$) & $-0.04\pm0.07$ & 14.8 / 111.2 & $+0.20\pm0.13$ & 2.7 / 10.1 & {\boldmath$+0.84\pm0.01$} & 1.7 / 12.8 & {\boldmath$-0.20\pm0.03$} & 5.1 / 35.0 \\
\DepGrow & $+$GLR ($\beta{=}{-}0.7$) & {\boldmath$+0.10\pm0.01$} & 13.1 / 104.3 & {\boldmath$+0.53\pm0.06$} & 1.6 / 9.8 & {\boldmath$+0.81\pm0.02$} & 2.0 / 8.5 & $-0.12\pm0.04$ & 7.2 / 28.2 \\
\bottomrule
\end{tabular}
\end{adjustbox}
\caption{\textbf{Fitted hyperparameter drift and sensitivity.} GLR, OM, RM, and WD denote global learning rate, output multiplier, residual multiplier, and weight decay. Each $\beta$ is fitted from the optima of quadratic slices at $d8$--$d10$, with one regression standard error; bold entries have $|\beta|\geq3$ standard errors. For constant recipes, the optimal value scales as $N^\beta$. For $+$GLR recipes, the GLR column measures residual drift relative to the applied exponent shown in the recipe column; the other columns still measure drift relative to constant base values. Regret is the mean loss of the $0.5\times$ and $2\times$ settings minus the center loss; range is the full loss span of the slice. Both are in units of $10^{-3}$ at the largest fully swept size. The $d8$ anchor is shared across constant and scaled recipes. Growth variants are swept around their fixed-recurrence GLR rule. These diagnostic fits are distinct from the deployed settings in Table~\ref{tab:full-ladder-recipes}.}
\label{tab:hp-drift}
\end{table}

\subsubsection{Train the Ladders with the Fitted Recipe}
Once the recipe is fixed, we train the scaling ladder. For a fixed-recurrence model, we count stored parameters $N$ and set the token budget to $T=\mathrm{TPP}\times N$. For a growth variant, the recommended recipe retains the Stage 2 allocation prescription and uses the fixed Stage 3 growth fraction, with tokens adjusted to preserve the prescribed compute budget across the two phases. We then apply the Stage 4 learning-rate rule and keep the other base hyperparameters unchanged. Table~\ref{tab:full-ladder-recipes} lists the settings used for the reported ladders.

The ladders span $d6$--$d20$ for Vanilla ($120$M--$1.8$B stored parameters) and $d6$--$d18$ for the looped variants ($120$M--$1.4$B for K2 and $130$M--$1.8$B for Dep; Table~\ref{tab:model-sizes}). Because the looped variants execute more blocks per token and train on more tokens per parameter, every ladder covers a similar range of compute, ending near $10^{20}$ FLOPs at its largest size despite the different stored-parameter counts. We compare at equal compute and fit Equation~\ref{eq:compute-optimal} to each ladder. First, we fit the irreducible loss $E$ from the Vanilla ladder with Huber loss minimization \citep{hoffmann2022training}. For the remaining ladders, we regress log reducible loss against log compute and report the exponent's regression standard error.

\begin{table}[!htbp]
\centering
\small
\begin{tabular}{lccc}
\toprule
Architecture & GLR exponent $\beta$ & Token allocation & Measured models \\
\midrule
Vanilla & $-0.8$ & TPP $5$ & $d6$--$d20$, even \\
Deep Vanilla & $-0.7$ & TPP $6$ & $d6$--$d18$, even \\
Operator-1 & $-0.6$ & TPP $6$ & $d6$--$d20$, even \\
Loop-2 & $-0.6$ & TPP $6$ & $d6$--$d18$, even \\
Untied-2 & $-0.6$ & TPP $6$ & $d6$--$d18$, even \\
\midrule
Deep Vanilla Grow & $-0.8$ & $t_{\mathrm{ref}}=6$ & $d6$--$d18$, even \\
Untied Grow & $-0.6$ & $t_{\mathrm{ref}}=8$ & $d6$--$d18$, even \\
Loop Grow & $-0.5$ & $t_{\mathrm{ref}}=7$ & $d6$--$d18$, even \\
\bottomrule
\end{tabular}
\caption{\textbf{Recipes used for the reported ladders.} Exponents apply in Equation~\ref{eq:glr-scaling} with $\mathrm{GLR}_{d8}=0.04$ for every family except Deep Vanilla Grow, whose fitted $d8$ anchor is $0.036$. TPP is $T/N$ over stored parameters; for the Grow variants, $t_{\mathrm{ref}}=T_0/N_{\mathrm{ref}}$ sets the pre-growth compute budget, with tokens recomputed after choosing the transition. Deep Vanilla Grow uses the Deep Vanilla TPP-$6$ reference budget and $\rho=0.5$. The Loop Grow and Untied Grow experiments include refitted growth allocations; the sensitivity results motivate the simpler prescription of reusing the fixed-recurrence allocation and a constant $\rho$ (Appendix~\ref{app:growth-fitting}). All other hyperparameters keep their base-tuned values.}
\label{tab:full-ladder-recipes}
\end{table}

\section{Additional Results for Compute-Optimal Scalings}
\label{app:comp-opt-results}

Figures~\ref{fig:ladder-ablation-summary} and~\ref{fig:ladder-recipe-shape-summary} test how the scaling results depend on architecture, training recipe, model shape, and optimizer. Figure~\ref{fig:ladder-ablation-summary} examines the boundary operator, runtime efficiency, and constant-recipe baselines; Figure~\ref{fig:ladder-recipe-shape-summary} separates the effects of tuning, depth, and recurrence. We compare the compute needed to reach the same loss, using interpolation within the measured ladders. Each ablation's reference is specified in its caption, so multiplier magnitudes should not be compared directly across panels.

\subsection{Architecture Choices and Runtime Efficiency}

\begin{figure}[!htbp]
\centering
\begin{subfigure}[t]{1.0\linewidth}
\vspace{0pt}
\centering
\includegraphics[width=\linewidth]{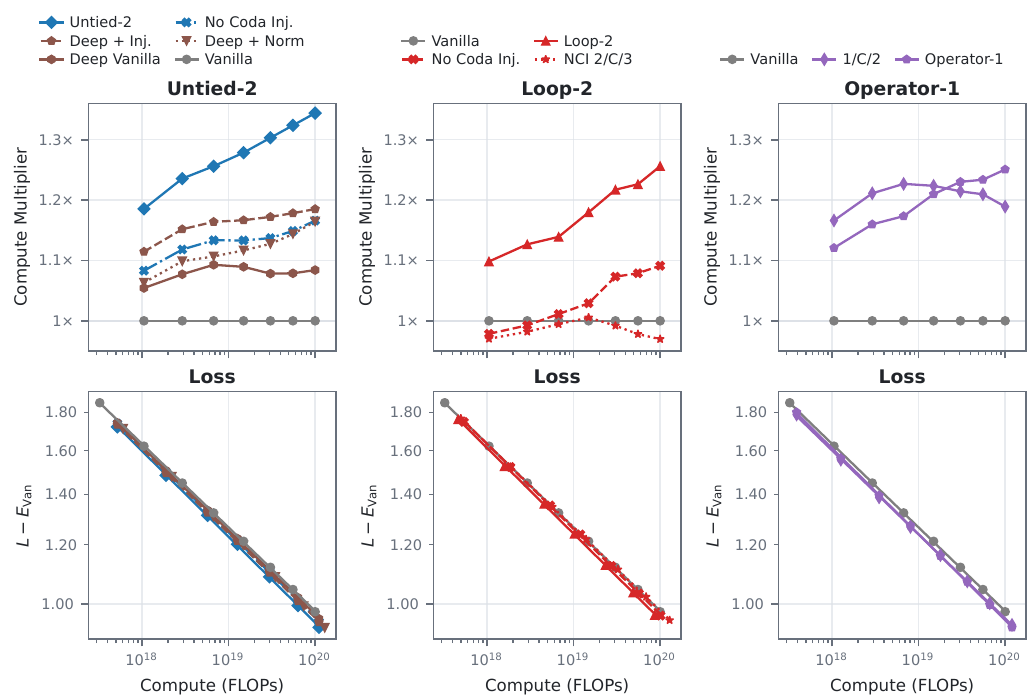}
\caption{Architecture ablations}
\label{fig:ladder-arch}
\end{subfigure}

\medskip
\begin{subfigure}[t]{0.5\linewidth}
\vspace{0pt}
\centering
\includegraphics[width=\linewidth]{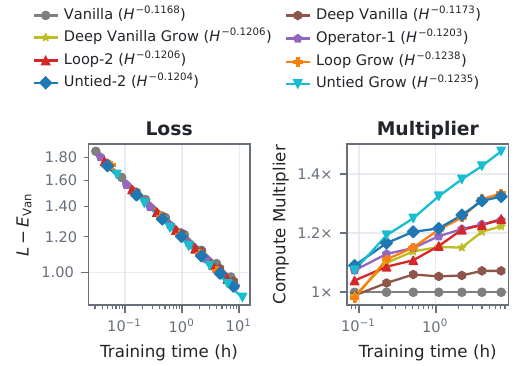}
\caption{Major variants by runtime}
\label{fig:ladder-runtime}
\end{subfigure}%
\begin{subfigure}[t]{0.5\linewidth}
\vspace{0pt}
\centering
\includegraphics[width=\linewidth]{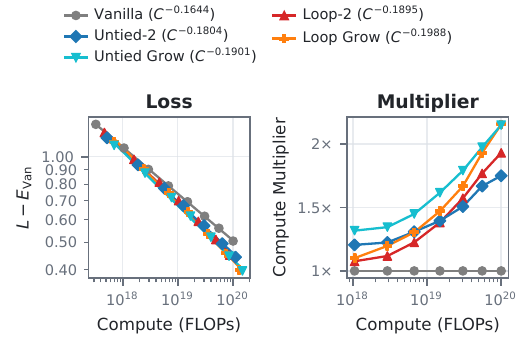}
\caption{Constant recipes}
\label{fig:ladder-constant}
\end{subfigure}
\caption{\textbf{Architecture, runtime, and constant-recipe ladders.} Points are measurements; loss curves fit power laws above a shared Vanilla-fitted floor. \textbf{(a)} Columns compare Untied-2 boundary-operator ablations, Loop-2 coda and allocation ablations, and Operator-1 block allocations. Rows show compute multipliers and reducible loss. NCI denotes no coda injection; $2/C/3$ and $1/C/2$ fix the prelude and coda depths while scaling the core. \textbf{(b)} Major variants against recorded optimization time, excluding evaluation and checkpoint overhead; the multiplier is Vanilla time divided by each variant's time at matched loss. \textbf{(c)} Ladders with constant base recipes; the multiplier uses constant-recipe Vanilla as the reference. Multipliers interpolate between measurements in log compute or log time without extrapolation.}
\label{fig:ladder-ablation-summary}
\end{figure}

\paragraph{The full boundary operator retains the largest gain.}
Figure~\ref{fig:ladder-arch} (left) compares Untied-2 with plain Deep Vanilla and three operator ablations. Near $10^{20}$ FLOPs, Untied-2 reaches roughly $1.34\times$ the compute efficiency of Vanilla, while using only normalization or injection, or omitting coda injection, reduces the multiplier to about $1.17$--$1.19\times$. Deep Vanilla remains near $1.08\times$. Thus, extra executed depth alone does not recover the full gain, and each of these operator choices contributes over the measured range.

\paragraph{Block allocation matters as models grow.}
Removing coda injection also reduces Loop-2's gain (Figure~\ref{fig:ladder-arch}, middle). Fixing its prelude and coda at two and three blocks while scaling only the core is worse still: the multiplier falls below one at the largest budgets. Operator-1 shows a similar limitation with a fixed $1/C/2$ allocation: its early advantage peaks and then declines, while the proportional allocation continues improving (right). These comparisons support scaling the prelude, core, and coda together.

\paragraph{The gains carry over to training time.}
Figure~\ref{fig:ladder-runtime} replaces FLOPs with recorded optimization time. The growth variants retain increasing time savings at matched loss, with Untied Grow giving the largest gain. The benefit therefore survives implementation costs on our hardware.

\subsection{Recipe Tuning Changes the Apparent Scaling Advantage}
\label{app:scaling-ablations}

Constant base recipes favor the looped and grown models over Vanilla (Figure~\ref{fig:ladder-constant}). This is consistent with the ordering of $\beta$ magnitudes in Table~\ref{tab:hp-drift}. Vanilla has the most negative $\beta$, and \Ktwo has the least negative values of $\beta$. As a result, compute multipliers under constant recipes are better than those under scaled ones, and in the long run, \KtwoGrow is the best constant recipe. This emphasizes the importance of hyperparameter optimality in scaling \citep{qiu2026hyperparameter,mlodozeniec2026completed}. Figure~\ref{fig:ladder-vanilla-scaling} reinforces this theme: scaling Vanilla's learning rate substantially improves its ladder. Adding output-multiplier or weight-decay scaling changes the multiplier by at most about $3\%$ relative to GLR scaling alone, whereas the tested muP output-multiplier rule without GLR scaling loses efficiency with scale. This supports the GLR-only scaling prescription in Appendix~\ref{sec:glr-rule}.

Tuning must also be architecture-specific. Simply ablating Operator-1's recipe with Vanilla's base tuned hyperparameters causes the exponent improvements to become constant ones (Figure~\ref{fig:ladder-inject-1-van-recipe}). Transferring a baseline recipe can therefore hide an architecture's scaling improvement even when it remains better at individual budgets.

\subsection{Model Shape and the Direction of Scaling}

Making a model deeper at a given width does not consistently improve its compute efficiency (Figure~\ref{fig:ladder-shape}). Deep Vanilla gives a modest gain over Vanilla, but the still-deeper $1{:}64$ depth-to-width ladders (Deeper series) do not consistently outperform their corresponding default shapes. In particular, Deeper Operator-1 loses the increasing advantage of Operator-1 over the measured range.

Holding executed depth fixed provides a more direct control (Figure~\ref{fig:ladder-width-only}). At depth 11, width-only Untied-2 remains more efficient than Deep Vanilla, but Untied-2 now has a better constant, not exponent, than Deep Vanilla. This supports that boundary operators become more useful as depths increase and the hypothesis that compute-optimal scaling exponents improve because of increased computational depths..

\subsection{Random Recurrence and Test-Time Passes}

We match the Loop Grow recipe but sample $K$ uniformly from $\{2,3,4,5,6\}$ at each optimizer step after growth, preserving mean recurrence four. Compute accounting uses the realized recurrence counts, and evaluation uses $K=4$. This random-recurrence ladder is slightly worse than fixed Loop Grow at matched compute (Figure~\ref{fig:ladder-rand}).

\begin{figure}[!htbp]
\centering
\includegraphics[width=\linewidth]{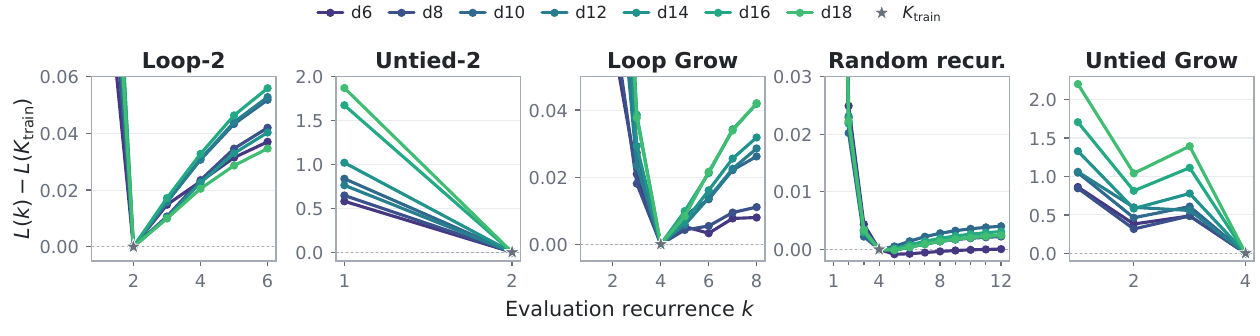}
\caption{\textbf{Sensitivity to evaluation recurrence.} Loss differences $L(k)-L(K_{\mathrm{train}})$ for the final-ladder checkpoints, with color indicating model size and stars marking the reference recurrence: $K_{\mathrm{train}}=2$ for Loop-2 and Untied-2, and $4$ for the growth variants. Tied models reuse their core for additional passes; untied models can only run their allocated cores. Fixed Loop-2 and Loop Grow are best at their trained recurrence. Random-recurrence training produces much flatter curves, with small gains at $k=5$ for five of seven checkpoints but no sustained improvement as more passes are added. Panels use different vertical scales.}
\label{fig:loop-exit}
\end{figure}

Random training does improve tolerance to extra test-time passes (Figure~\ref{fig:loop-exit}). At $k=8$, loss relative to $k=4$ changes by $-0.0003$ to $+0.0028$, compared with increases of $0.008$--$0.042$ for fixed Loop Grow. Five of seven random-recurrence checkpoints have a shallow minimum at $k=5$, but the improvement is below $0.001$ loss and does not continue with further passes. In this setting, random recurrence chiefly reduces the penalty for extra passes rather than providing sustained test-time scaling.

\subsection{Optimizer Choice Affects Both the Constant and the Exponent}

\begin{figure}[!htbp]
\centering
\begin{subfigure}[t]{0.5\linewidth}
\vspace{0pt}
\centering
\includegraphics[width=\linewidth]{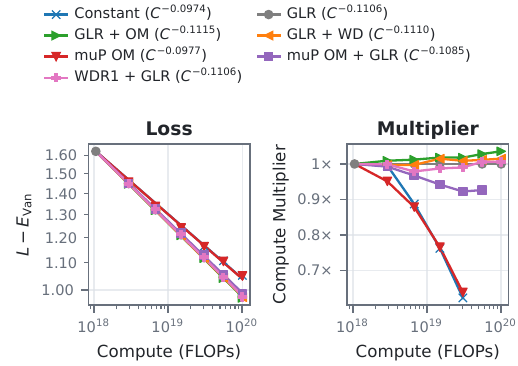}
\caption{Vanilla hyperparameter scaling}
\label{fig:ladder-vanilla-scaling}
\end{subfigure}%
\begin{subfigure}[t]{0.5\linewidth}
\vspace{0pt}
\centering
\includegraphics[width=\linewidth]{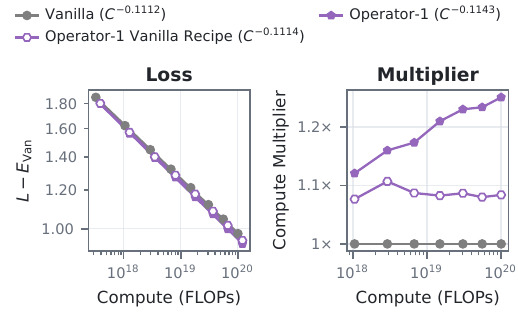}
\caption{Operator-1 recipe ablations}
\label{fig:ladder-inject-1-van-recipe}
\end{subfigure}

\medskip
\begin{subfigure}[t]{0.5\linewidth}
\vspace{0pt}
\centering
\includegraphics[width=\linewidth]{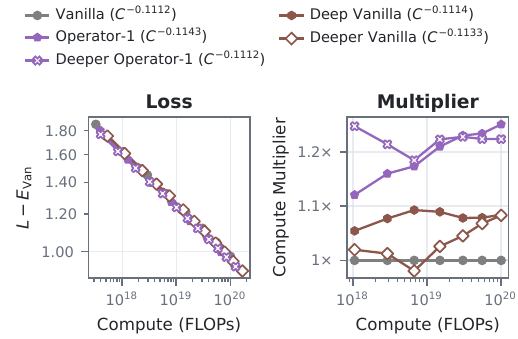}
\caption{Depth-to-width ratio}
\label{fig:ladder-shape}
\end{subfigure}%
\begin{subfigure}[t]{0.5\linewidth}
\vspace{0pt}
\centering
\includegraphics[width=\linewidth]{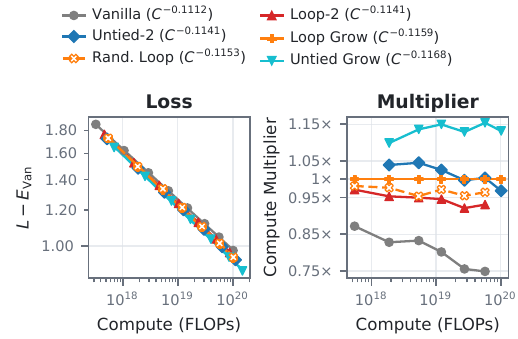}
\caption{Random recurrence}
\label{fig:ladder-rand}
\end{subfigure}

\medskip
\begin{subfigure}[t]{0.5\linewidth}
\vspace{0pt}
\centering
\includegraphics[width=\linewidth]{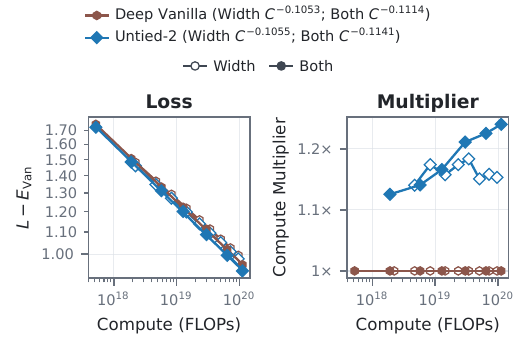}
\caption{Width-only versus width/depth scaling}
\label{fig:ladder-width-only}
\end{subfigure}%
\begin{subfigure}[t]{0.5\linewidth}
\vspace{0pt}
\centering
\includegraphics[width=\linewidth]{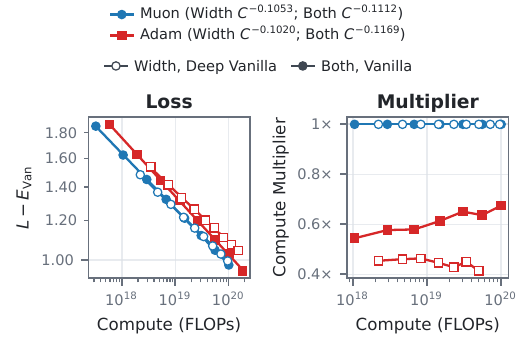}
\caption{Muon versus Adam}
\label{fig:ladder-optimizers}
\end{subfigure}
\caption{\textbf{Recipe, shape, recurrence, and optimizer ablations.} Each subfigure shows reducible loss and a loss-matched compute multiplier, interpolated without extrapolation. \textbf{(a)} Vanilla recipes share a fitted $d8$ anchor; multipliers are relative to constant + GLR. OM, WD, and WDR denote output multiplier, weight decay, and warmdown ratio. \textbf{(b)} Ablating Operator-1's recipe by using Vanilla's based tuned hyperparameters. \textbf{(c)} Shape comparisons; Deeper Vanilla and Deeper Operator-1 use depth-to-width ratio $1{:}64$. Panels (b,c) use Vanilla as the reference. \textbf{(d)} Random recurrence samples $K\in\{2,3,4,5,6\}$ after growth and evaluates at $K=4$; compute uses the realized recurrence counts, with fixed Loop Grow as the reference. \textbf{(e,f)} Open markers indicate width-only scaling at executed depth 11; filled markers indicate coupled width/depth scaling. Each regime uses its own reference: Deep Vanilla with Muon in (e), and the corresponding Muon ladder in (f). Each optimizer uses its fitted token budget.}
\label{fig:ladder-recipe-shape-summary}
\end{figure}

Muon has a constant improvement over Adam under width-only scaling, consistent with \citet{qiu2026hyperparameter}, but in coupled width/depth scaling, we find that compute efficiency gains \textit{shrink} with larger scales (Figure~\ref{fig:ladder-optimizers}). This suggests that Muon optimizer may not be as effective as Adam for deeper networks. This discrepancy between width and joint scaling motivates investigation on the interaction of architectures and optimizers.

\section{Corpus Transfer and Downstream Evaluation}
\label{app:transfer-evaluation}

We test whether the compute-optimal gains carry over to a different pretraining corpus and to downstream tasks. We first compare Vanilla and \DepGrow on FineWeb and FineWeb-Edu, then define the evaluation protocol and examine the eight-architecture downstream ladders. Finally, we describe the taskwise calibration used to extrapolate CORE accuracy beyond the measured compute range.

\subsection{Transfer from FineWeb to FineWeb-Edu}

\begin{figure}[!htbp]
\centering
\begin{subfigure}[t]{0.735\linewidth}
\vspace{0pt}
\centering
\includegraphics[width=\linewidth]{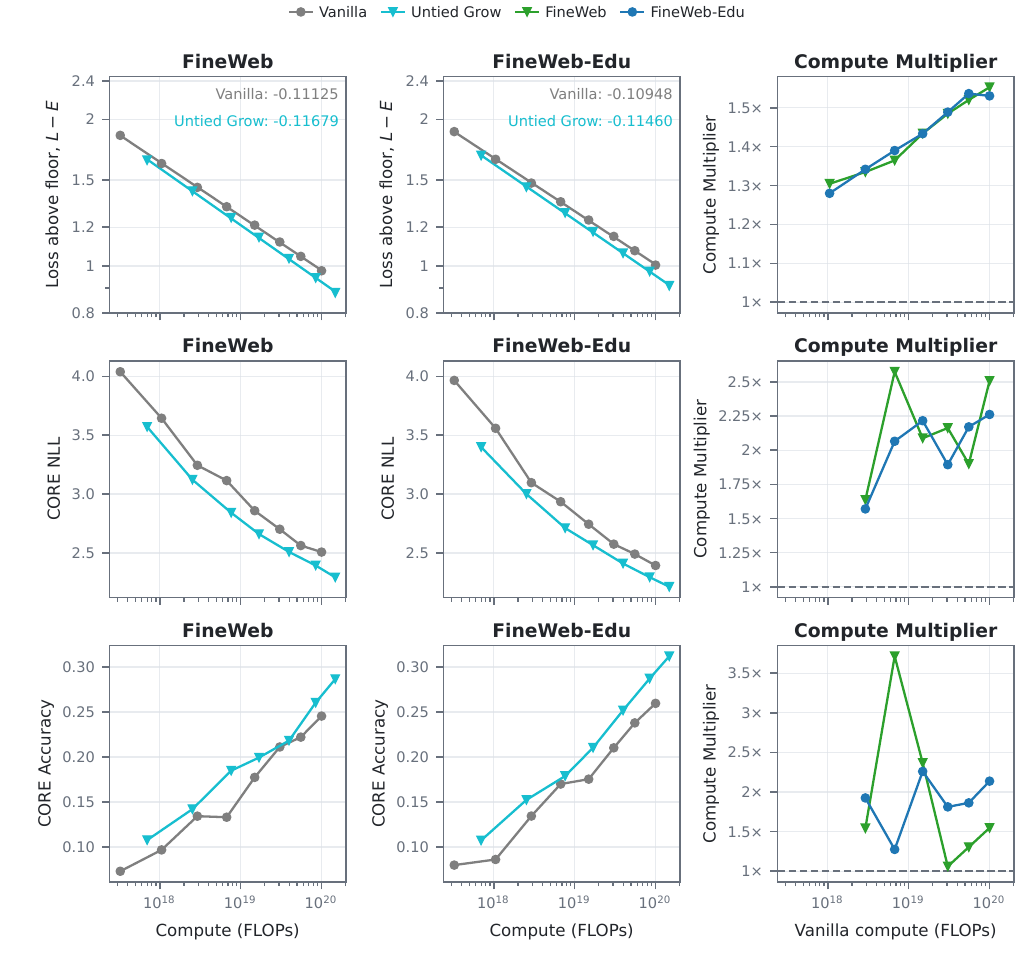}
\caption{Architecture gains on FineWeb and FineWeb-Edu}
\label{fig:data-compare}
\end{subfigure}\hfill
\begin{subfigure}[t]{0.245\linewidth}
\vspace{0pt}
\centering
\includegraphics[width=\linewidth]{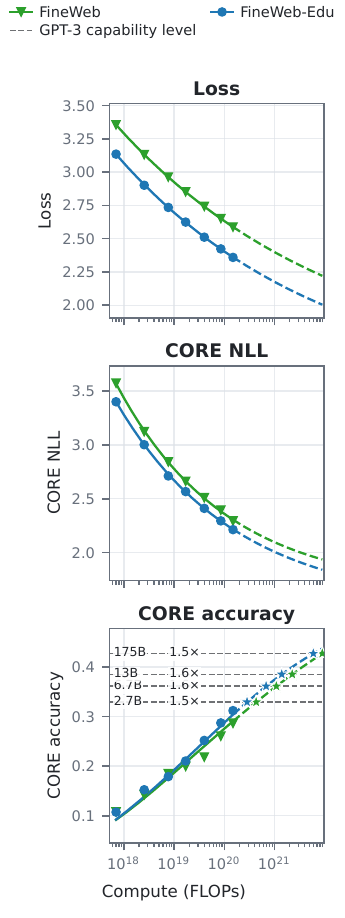}
\caption{Corpus comparison}
\label{fig:dataset-comparison}
\end{subfigure}
\caption{\textbf{Transfer from FineWeb to FineWeb-Edu.} \textbf{(a)} Vanilla and \DepGrow on each corpus: rows show loss above the Vanilla-fitted floor, CORE NLL, and CORE accuracy. The final column shows Vanilla-to-\DepGrow compute multipliers at matched metrics within each corpus, interpolated without extrapolation. \textbf{(b)} \DepGrow corpus comparison: loss, CORE NLL, and CORE accuracy from top to bottom. Solid curves cover measured domains and dashed curves show extrapolation. Horizontal references mark GPT-3 CORE scores; multipliers compare FineWeb compute with FineWeb-Edu compute at matched accuracy. The held-out $d26$ run is excluded from both subfigures.}
\label{fig:corpus-transfer-summary}
\end{figure}

We repeat the Vanilla and \DepGrow ladders on FineWeb-Edu with the same architectures and training recipes. Figure~\ref{fig:data-compare} separates the architectural advantage within each corpus from the effect of changing the corpus. On both datasets, \DepGrow's loss-matched compute multiplier increases with scale. Its fitted loss exponent exceeds Vanilla's by a similar amount: $0.1168$ versus $0.1113$ on FineWeb, and $0.1146$ versus $0.1095$ on FineWeb-Edu, using a separate Vanilla-fitted irreducible loss for each corpus. Downstream gains also persist, although accuracy-based multipliers fluctuate more than loss-based multipliers.

For \DepGrow itself, FineWeb-Edu improves downstream performance at matched training compute (Figure~\ref{fig:dataset-comparison}). The fitted CORE curves project that FineWeb requires about $1.5$--$1.6\times$ as much compute to reach the displayed GPT-3 reference scores. These are extrapolated corpus comparisons; the multipliers in Figure~\ref{fig:data-compare} instead interpolate between measured architectures within each corpus. Appendix~\ref{app:downstream-perf} specifies the forecasting procedure. The held-out $d26$ run is excluded from these fits.

\subsection{Downstream Evaluation Protocol}
\label{app:downstream}

We follow the 22-task DCLM CORE benchmark \citep{li2024datacomp} as implemented in \citet{nanochat}, evaluating all 91,037 examples at each checkpoint. Full CORE accuracy and answer negative log-likelihood (NLL) are the primary metrics. We also report a secondary accuracy score on a fixed subset of 17 tasks.

\paragraph{Full CORE accuracy.}
For multiple-choice and shared-ending tasks, the model chooses the candidate with the lowest mean token loss. For language-modeling tasks, an answer is correct only if every reference token is predicted correctly. Let $a_j$ be task $j$'s raw accuracy and $b_j$ its random-guess accuracy. We chance-center each task and average with equal task weight:
\begin{equation}
c_j=\frac{a_j-b_j}{1-b_j},
\qquad
\operatorname{CORE}_{22}=\frac{1}{22}\sum_{j=1}^{22}c_j.
\label{eq:downstream-core}
\end{equation}
Zero denotes chance, one denotes perfect accuracy, and negative values denote below-chance performance. Table~\ref{tab:core-eval-type} lists the tasks and scoring types.

\begin{table}[!htbp]
\centering
\small
\begin{tabular}{lll}
\toprule
Task name & Evaluation type & Filter \\
\midrule
HellaSwag (zero-shot) & MC & - \\
ARC-Easy & MC & - \\
ARC-Challenge & MC & - \\
COPA & MC & - \\
CommonsenseQA & MC & Kendall ($\tau=0.000$) \\
PIQA & MC & - \\
OpenBookQA & MC & - \\
HellaSwag (10-shot) & MC & - \\
AGI Eval LSAT-AR & MC & - \\
BoolQ & MC & Below chance ($-4.62$ pp); Kendall ($\tau=0.286$) \\
BIG-bench Language Identification & MC & Kendall ($\tau=0.143$) \\
Winograd & SE & - \\
WinoGrande & SE & - \\
Jeopardy & LM & - \\
BIG-bench QA Wikidata & LM & - \\
LAMBADA OpenAI & LM & - \\
BIG-bench Dyck Languages & LM & - \\
BIG-bench CS Algorithms & LM & Kendall ($\tau=0.500$) \\
BIG-bench Operators & LM & - \\
BIG-bench Repeat Copy Logic & LM & At chance ($+0.00$ pp); Kendall ($\tau=0.423$) \\
SQuAD & LM & - \\
CoQA & LM & - \\
\bottomrule
\end{tabular}
\caption{DCLM CORE tasks and evaluation types. MC denotes multiple choice, SE shared ending, and LM language modeling. A dash in the Filter column denotes a retained task; other entries list the failed criteria: the accuracy difference from random chance in percentage points (pp) and/or Kendall's $\tau_b$ across Vanilla checkpoints. Retention requires an accuracy gap of at least $+2$ pp and $\tau_b \geq 0.50$. Values are rounded; the unrounded $\tau_b$ for BIG-bench CS Algorithms is marginally below $0.50$.}
\label{tab:core-eval-type}
\end{table}

\paragraph{Answer NLL.}
Answer NLL measures the probability assigned to the correct reference answer, providing a continuous comparison even when two models select the same option \citep{llama3}. For example $i$ of task $j$, with prompt $x_{ij}$ and answer tokens $y_{ij1:T_{ij}}$, define
\begin{align}
\ell^{\mathrm{ans}}_{ij} &=-\frac{1}{T_{ij}}\sum_{t=1}^{T_{ij}}\log p\!\left(y_{ijt}\mid x_{ij},y_{ij,<t}\right), \notag\\
\operatorname{NLL}^{\mathrm{ans}}_{22} &=\frac{1}{22}\sum_{j=1}^{22}\left(\frac{1}{n_j}\sum_{i=1}^{n_j}\ell^{\mathrm{ans}}_{ij}\right).
\label{eq:downstream-answer-nll}
\end{align}
Here $n_j$ is the number of examples in task $j$. Prompt tokens are excluded; token averaging prevents longer answers from receiving larger losses merely because of their length. We average examples within each task and then average tasks equally. Lower values are better; the figures abbreviate this metric as CORE NLL.

\paragraph{Frozen Vanilla-only filter.}
Some tasks provide little signal at the scales studied, so we additionally report mean centered accuracy on a subset selected using only the eight final-GLR Vanilla checkpoints. Inspired by the evaluation-task selection criteria of \citet{fineweb}, a task is retained if the largest-token checkpoint's raw accuracy is at least two percentage points above chance and Kendall's $\tau_b$ between training amount and accuracy is at least $0.50$. This selects 17 tasks (Table~\ref{tab:core-eval-type}). The subset stays fixed across architectures, recipes, and corpora. It is a post-hoc sensitivity analysis, reported alongside the two full-suite metrics; every evaluation still scores all 22 tasks.

\paragraph{Replicates and reproducibility.}
We use evaluation seeds 0, 1, and 2, which change the few-shot demonstrations rather than the benchmark examples. Deterministic zero-shot results may be reused across replicates. We average replicates within each task before averaging tasks.

\subsection{Downstream Scaling and Task-Level Residuals}
\label{app:downstream-vs-loss}

\begin{figure}[!htbp]
\centering
\begin{subfigure}{\linewidth}
\centering
\includegraphics[width=\linewidth]{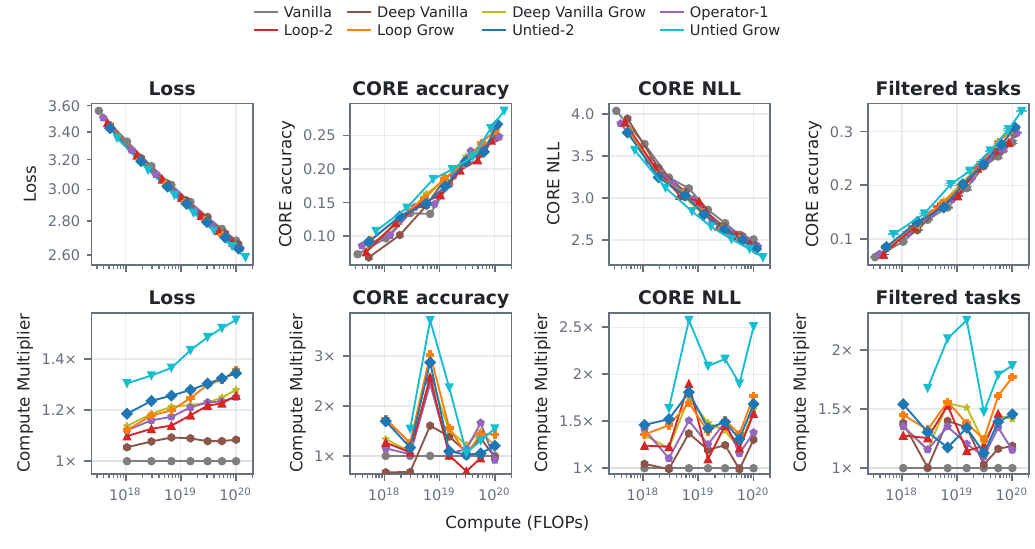}
\caption{Downstream scaling ladders}
\label{fig:downstream-major}
\end{subfigure}

\medskip
\begin{subfigure}[t]{0.245\linewidth}
\vspace{0pt}
\centering
\includegraphics[width=\linewidth]{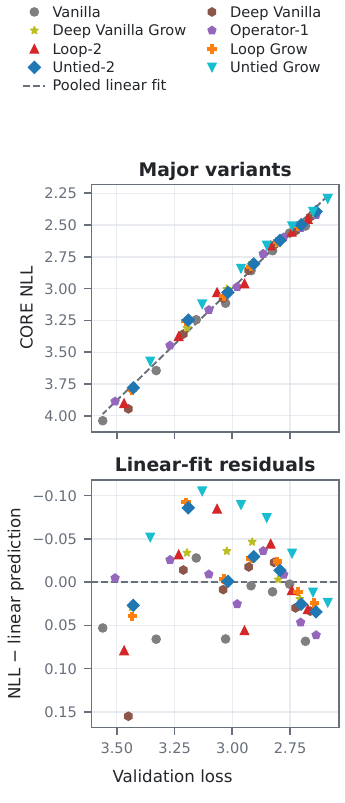}
\caption{CORE NLL fits}
\label{fig:core-vs-nll}
\end{subfigure}\hfill
\begin{subfigure}[t]{0.735\linewidth}
\vspace{0pt}
\centering
\includegraphics[width=\linewidth]{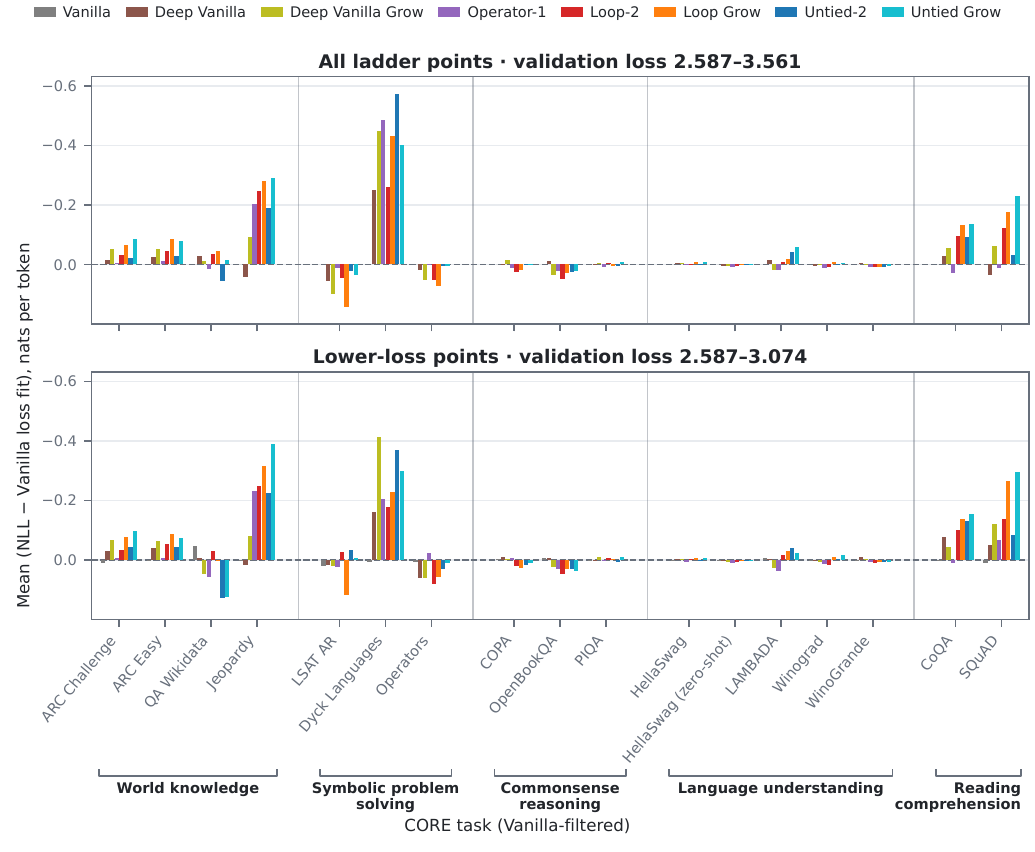}
\caption{Mean task residuals by architecture}
\label{fig:matched-loss-task-bars}
\end{subfigure}
\caption{\textbf{Downstream performance across architectures and tasks.} \textbf{(a)} Loss and downstream metrics for eight architectures; the lower row shows compute multipliers relative to Vanilla, interpolated between measurements without extrapolation. \textbf{(b)} CORE NLL versus validation loss (top) and residuals from one linear fit pooled across all architectures (bottom), with both axes reversed. \textbf{(c)} Mean task NLL residuals relative to a Vanilla-only linear fit against validation loss, grouped by CORE category, for all ladder points (top) and the lower-loss subset (bottom). Negative residuals indicate lower NLL and point upward.}
\label{fig:downstream-summary}
\end{figure}

Figure~\ref{fig:downstream-major} evaluates 58 checkpoints across eight architectures. The looped and grown variants generally improve downstream compute efficiency, especially on answer NLL and filtered CORE accuracy. Full CORE accuracy is less smooth across checkpoints, and its matched-score compute estimates fluctuate accordingly. Each multiplier uses interpolation between measurements without extrapolation.

Much of the downstream improvement tracks pretraining loss, but architecture-dependent residuals remain (Figure~\ref{fig:core-vs-nll}). A single linear fit pools all checkpoints with equal weight; \DepGrow generally has lower CORE NLL than this fit predicts. Figure~\ref{fig:matched-loss-task-bars} uses a different reference to resolve the task contributions: a separate Vanilla-only NLL-versus-validation-loss line for each task. Bars average measured NLL minus that prediction for each architecture, over all checkpoints or the lower half of the pooled validation-loss range. Negative residuals indicate better answer prediction than the Vanilla trend. Differences are uneven across tasks, with prominent gains on Dyck Languages and several reading-comprehension tasks. These residuals depend on the linear reference, which is also evaluated beyond Vanilla's measured loss range for some checkpoints.

\subsection{Forecasting Downstream Performance}
\label{app:downstream-perf}

To forecast the compute needed for a target CORE score \citep{llama3}, we fit a taskwise pipeline: compute to answer NLL, answer NLL to centered accuracy, and task accuracy to the full CORE score. Neither validation loss nor the aggregate CORE NLL is an intermediate in this forecast.

\begin{enumerate}
    \item \textbf{Compute to task NLL.} For series $a$ (an architecture or corpus) and each of the 17 selected tasks $t$, fit $B_{a,t}(C)=E_{a,t}+A_{a,t}(C/10^{18})^{-\alpha_{a,t}}$ with Huber loss. Each series has its own task-NLL scaling laws.
    \item \textbf{Task NLL to accuracy.} Fit one sigmoid per task, $\hat c_t(B)=[1+\exp(s_tB-b_t)]^{-1}$, shared across the series being compared. The target is the task's centered accuracy.
    \item \textbf{Aggregation.} Average the 17 predicted task scores and fit a shared, no-intercept coefficient $r$ to recover full CORE accuracy: $\widehat{\operatorname{CORE}}_{22,a}(C)=\frac{r}{17}\sum_t\hat c_t(B_{a,t}(C))$.
\end{enumerate}

\begin{figure}[!htbp]
\centering
\includegraphics[width=\linewidth]{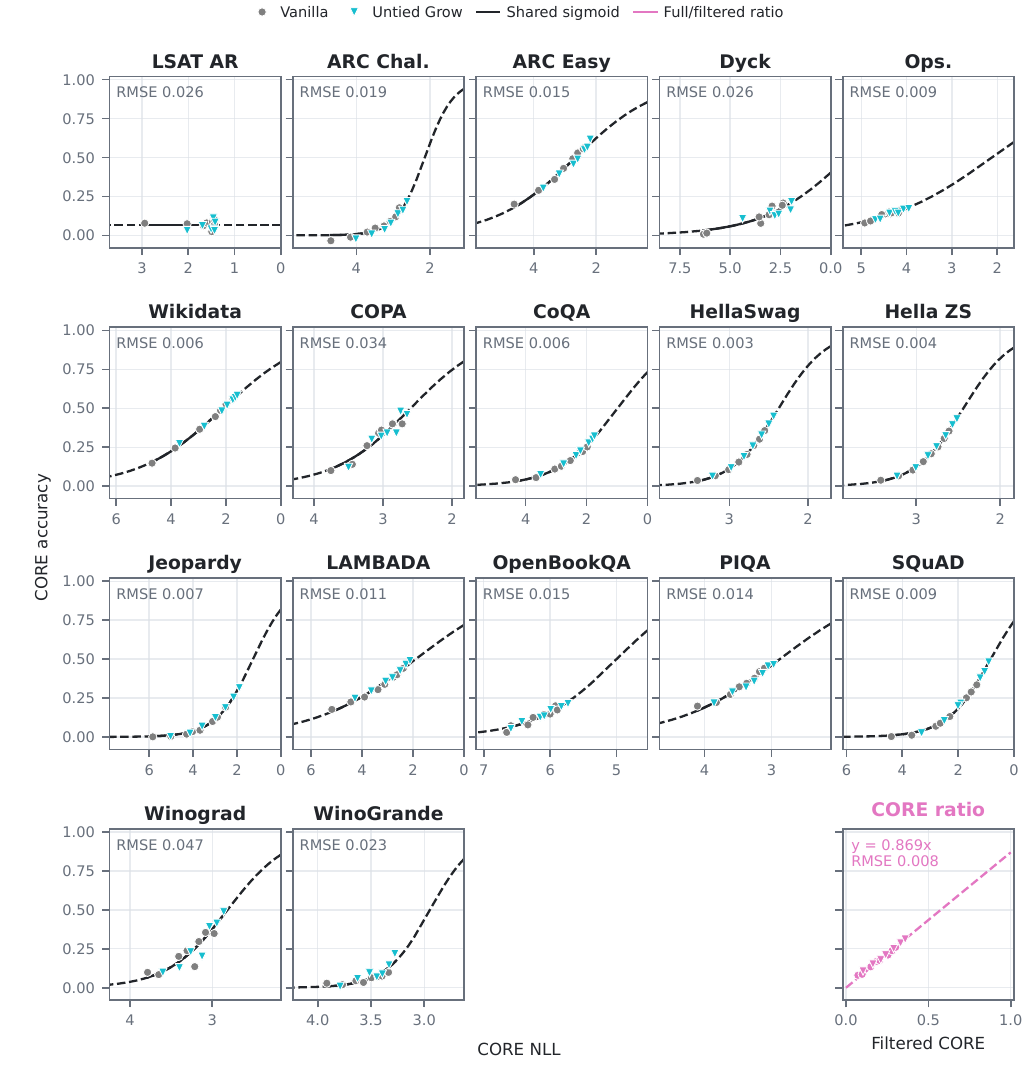}
\caption{FineWeb-Edu calibration for the architecture comparison. The first 17 panels show centered task accuracy against task CORE NLL for the Vanilla-selected tasks. Each solid segment is the shared Huber sigmoid over all FineWeb-Edu Vanilla and \DepGrow measurements; dashed segments extend the fitted curve beyond the measured NLL range. The final purple panel shows the no-intercept conversion from the 17-task filtered score to full CORE across the same 15 measurements.}
\label{fig:sigmoids-linear-fits}
\end{figure}

For the FineWeb-Edu architecture comparison, calibration pools eight Vanilla and seven \DepGrow checkpoints, giving $r=0.86912$. Figure~\ref{fig:sigmoids-linear-fits} shows the task sigmoids and the filtered-to-full conversion. The corpus comparison fits a separate shared calibration to seven FineWeb and seven FineWeb-Edu \DepGrow checkpoints, giving $r=0.87104$. Only task selection is inherited from the FineWeb Vanilla filter; both calibrations use the measurements in their respective comparisons.

The held-out $d26$ model is excluded from every fit. At its compute of $1.225\times10^{21}$ FLOPs, the architecture-comparison pipeline predicts CORE accuracy $0.3837$, versus the observed $0.3865\pm0.0015$. The error bar is one standard deviation across evaluation seeds, not uncertainty in the fitted extrapolation. Matching the GPT-3 reference scores \citep{miniseries} with these curves gives projected Vanilla-to-\DepGrow compute ratios of about $2.5$--$3.5\times$ (Figure~\ref{fig:dep-grow-extrap}); these projections assume that the small-scale relationships continue beyond the measured ladders.

\subsection{Scope of the Comparisons}

The ladders use a fixed batch size and common hardware, so their fitted recipes do not address joint optimization of batch size and model scale. The filtered score and task residuals supplement the full-suite metrics, while extrapolated CORE predictions additionally depend on the fitted task-NLL laws and calibration curves.

\clearpage
\section{Additional Results for Data-Constrained Scaling}
\label{app:data-const}

We extend Section~\ref{sec:data-constr} across repetition levels, weight decay, and architecture controls. At fixed token exposure, we compare spending additional compute on stored model size or extra core passes.

\subsection{Data Repetition Changes the Preferred Recurrence}

\begin{figure}[!htbp]
    \centering
    \includegraphics[width=\linewidth]{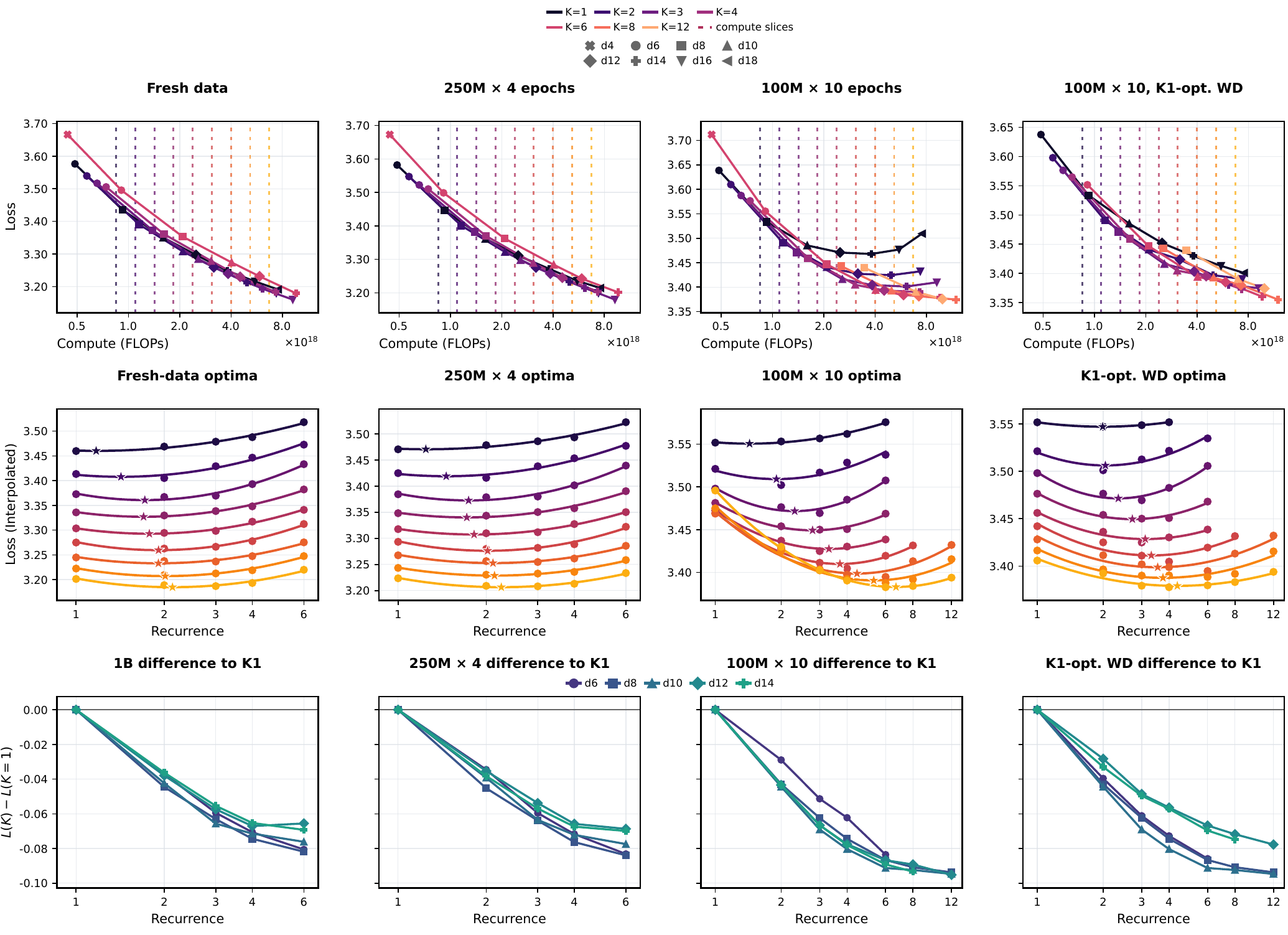}
    \caption{\textbf{Recurrence across data-repetition regimes.} Columns compare fresh data, $250$M tokens repeated four times, and $100$M tokens repeated ten times at fixed weight decay $0.8$, followed by the ten-epoch regime with each size's $K=1$-selected weight decay. Rows show loss versus compute, matched-compute recurrence fits, and loss differences from $K=1$ at fixed stored depth. Colors identify recurrence in the top row, compute slices in the middle row, and depth in the bottom row. Filled stars mark interior fitted optima; hollow stars mark the best measured recurrence when the fit is censored. The ten-epoch columns include $K=8,12$ only where their measured compute ranges bracket the comparison budget.}
    \label{fig:data-constr-regimes}
\end{figure}

Figure~\ref{fig:data-constr-regimes} compares approximately 1B token exposures from fresh data, a 250M-token pool repeated four times, and a 100M-token pool repeated ten times. The first three columns use weight decay $0.8$ and otherwise fixed base hyperparameters. Under fresh data and four epochs, the preferred recurrence remains near one to two passes across the measured budgets. With ten epochs, increasing model size at low recurrence eventually worsens loss, while higher recurrence postpones this upturn; the best recurrence therefore rises more strongly with compute.

The fourth column repeats the ten-epoch comparison with weight decay selected at $K=1$ for each model size and then held fixed across recurrence. This reduces overfitting and the shift toward larger $K$, but does not remove the shift. The bottom row measures $L(K)-L(1)$ at fixed stored depth: extra passes continue to reduce loss even when increasing stored size becomes less useful. These fixed-depth comparisons spend more compute as $K$ grows; the middle row makes the equal-compute comparison.

At each compute slice, we interpolate within measured ranges and fit a quadratic in log recurrence and log loss. Filled stars mark interior minima; hollow stars mark the best measured recurrence when no interior optimum is identified. The ten-epoch sweeps include $K=8,12$ where measurements bracket the budget; the other regimes end at $K=6$.

\subsection{Weight Decay and Recurrence Are Complementary}

\begin{figure}[!htbp]
    \centering
    \includegraphics[width=\linewidth]{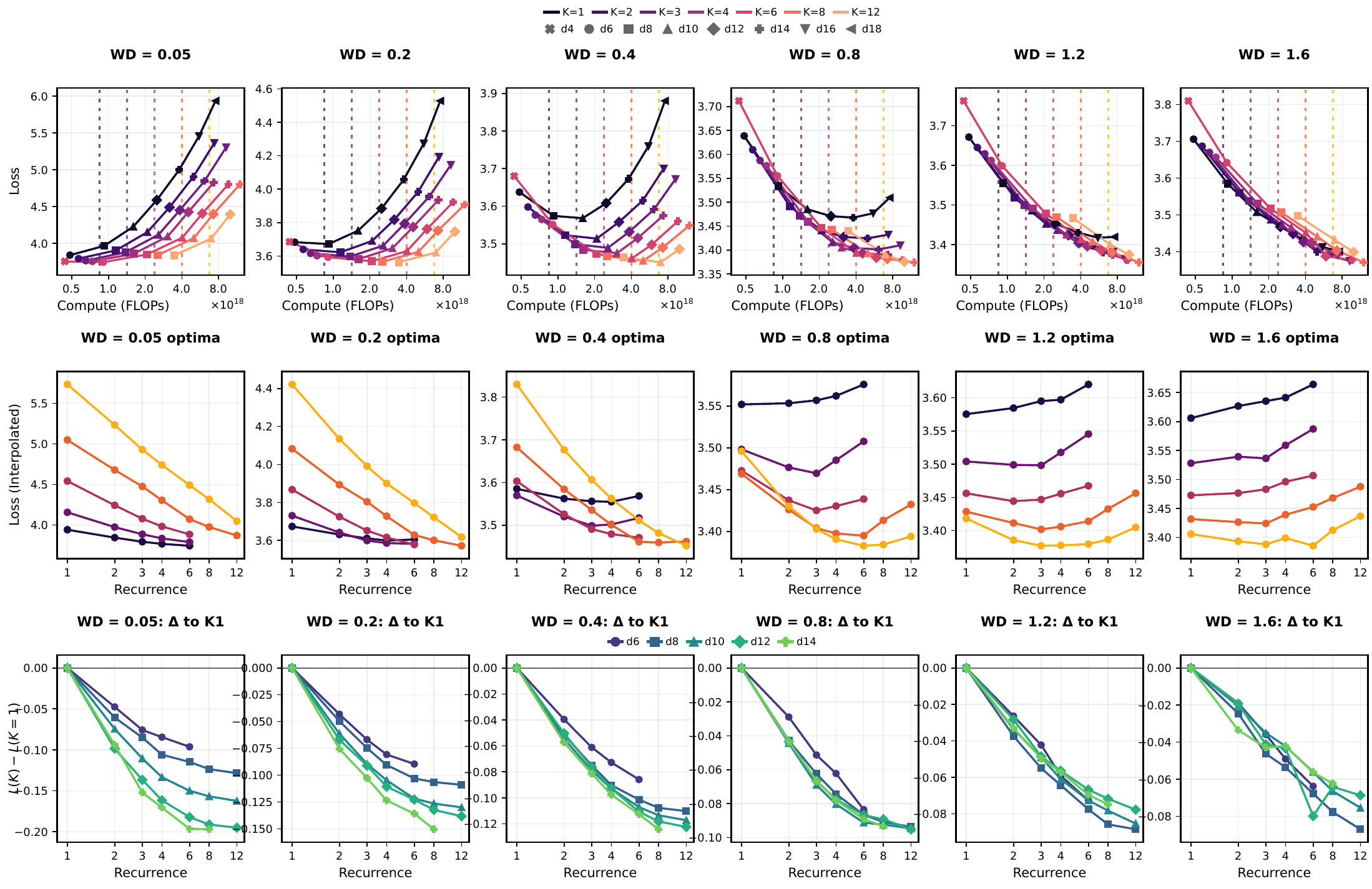}
    \caption{\textbf{Weight decay moderates the preference for recurrence.} Each column uses one of six weight-decay values under the $100$M-token, ten-epoch protocol. Rows show loss versus compute, matched-compute recurrence comparisons, and $L(K)-L(1)$ at fixed stored depth. Low weight decay produces stronger overfitting with model size and a greater preference for additional passes. Compute-slice colors match between the first two rows; the bottom row uses depth colors. Comparisons interpolate within measured ranges, including $K=8,12$ where available. Panels use independent vertical scales.}
    \label{fig:data-constr-wd}
\end{figure}

Figure~\ref{fig:data-constr-wd} crosses the ten-epoch protocol with weight decay in $\{0.05,0.2,0.4,0.8,1.2,1.6\}$. Weak regularization produces the strongest upturn in loss as stored size grows and favors high recurrence at large budgets. Stronger weight decay reduces that upturn and moderates the preferred recurrence. Extra passes still improve loss at fixed stored depth, although the gains are smaller. Thus, recurrence gains are largest when regularization is insufficient, but are not eliminated by tuning weight decay.

\subsection{Hyperparameter Transfer across Recurrence}

\begin{figure}[!htbp]
    \centering
    \includegraphics[width=\linewidth]{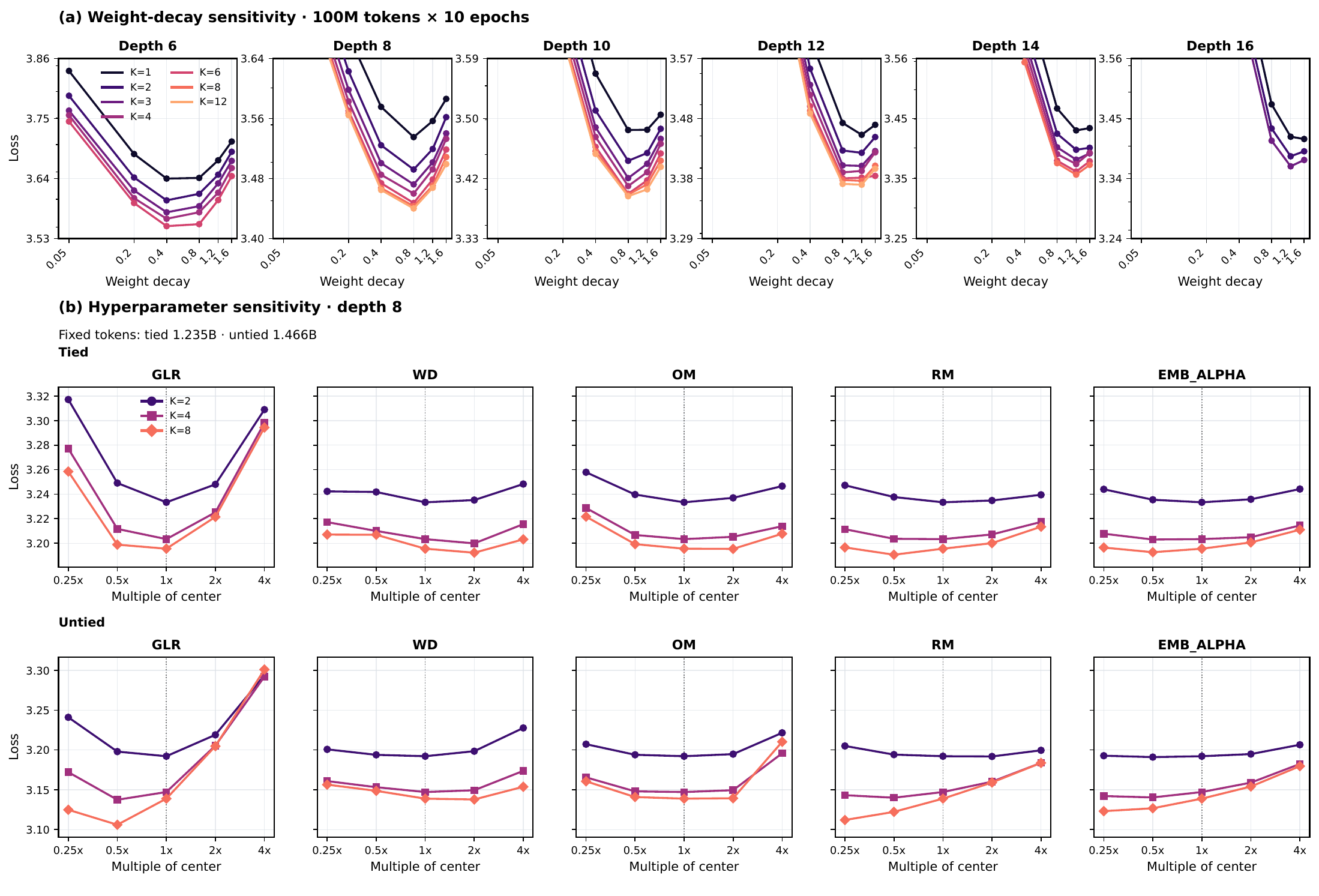}
    \caption{\textbf{Hyperparameter sensitivity to stored depth and recurrence.} \textbf{(a)} Weight-decay sweeps under the $100$M-token, ten-epoch protocol, with one panel per stored depth. The preferred weight decay shifts with depth but varies relatively little across tied recurrence. \textbf{(b)} Fresh-data sweeps at $d8$ with two, four, and eight passes, using fixed token counts within each family: $1.235$B for tied models and $1.466$B for untied models. Columns vary GLR, weight decay, output multiplier, residual multiplier, and injection scale by $\{1/4,1/2,1,2,4\}$ times their base values. Dotted lines mark the base recipe. Untied recurrence causes larger shifts in several optima.}
    \label{fig:hyper-sensitivity}
\end{figure}

The preferred weight decay changes more with stored depth than with tied recurrence (Figure~\ref{fig:hyper-sensitivity}a). On the shared grid, selecting weight decay at $K=1$ gives $0.4$ at $d6$, $0.8$ at $d8$--$d10$, $1.2$ at $d12$--$d14$, and $1.6$ at $d16$--$d18$. Reusing these values across $K$ yields the fourth column of Figure~\ref{fig:data-constr-regimes}.

A separate fresh-data sweep at $d8$ tests transfer of the global learning rate, weight decay, output multiplier, residual multiplier, and injection scale across $K=2,4,8$ (Figure~\ref{fig:hyper-sensitivity}b). Tied models retain broadly similar optima, while untied models shift more, particularly toward smaller learning-rate and residual multipliers at higher recurrence. Tokens are fixed within each family at approximately $1.235$B for tied models and $1.466$B for untied models. Compute increases with recurrence; for untied models, tokens per stored parameter also decrease.

\subsection{Architecture Controls: Sharing Weights and Adding Depth}

\begin{figure}[!htbp]
    \centering
    \includegraphics[width=\linewidth]{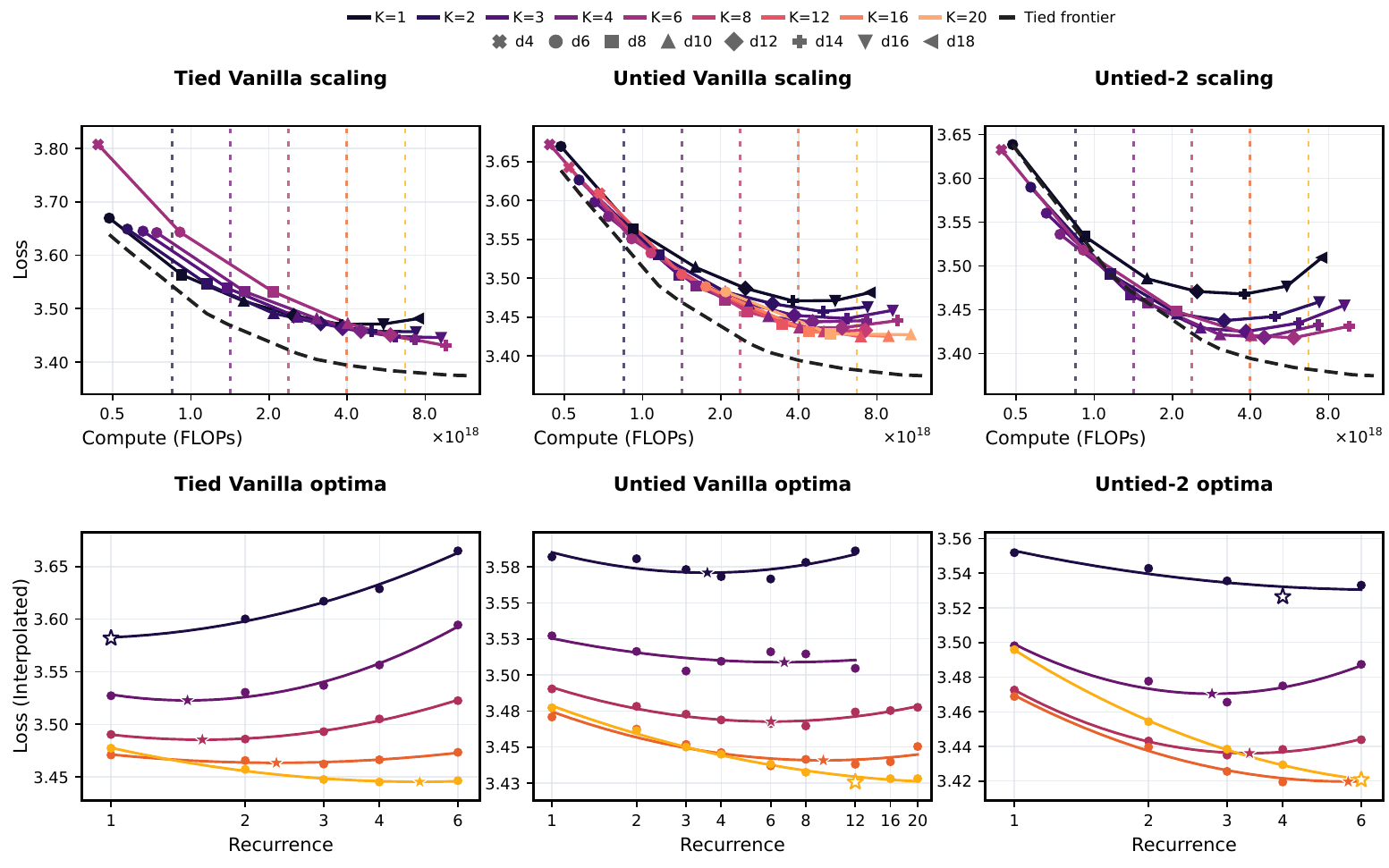}
    \caption{\textbf{Architecture controls under data repetition.} Columns show Tied Vanilla, Untied Vanilla, and the untied family with the boundary operator, all at weight decay $0.8$ on $100$M tokens for ten epochs. Top: loss versus compute, with the tied boundary-operator frontier repeated as a dashed reference. Bottom: matched-compute recurrence comparisons. Filled stars mark interior fitted optima; hollow stars mark censored fits at the best measured recurrence. Both plain-Transformer controls remain above the reference frontier. The untied operator family is slightly better at the two smallest reference budgets but worse at the three larger budgets.}
    \label{fig:data-constr-arch}
\end{figure}

Figure~\ref{fig:data-constr-arch} compares three controls under the 100M-token, ten-epoch protocol at weight decay $0.8$. Tied Vanilla repeats plain Transformer cores with shared weights. Untied Vanilla uses independent copies and is equivalent to a deeper plain Transformer. Both use the frozen Vanilla recipe without the boundary operator, share the $K=1$ ladder, and match FLOPs at each depth and recurrence. The third control varies the number of untied cores with the boundary operator.

The preferred recurrence generally increases with budget in all three families, so repetition can favor depth even without weight sharing. Both plain-Transformer controls remain above the tied boundary-operator frontier. The untied operator model is slightly better at the two smallest reference budgets but worse at the three larger ones: tied looping's advantage emerges as adding depth through new parameters becomes less effective.

\end{document}

%% file: math_commands.tex
\usepackage{amsmath,amsfonts,bm}

\def\eqref#1{equation~\ref{#1}}

\def\1{\bm{1}}

\DeclareMathAlphabet{\mathsfit}{\encodingdefault}{\sfdefault}{m}{sl}
\SetMathAlphabet{\mathsfit}{bold}{\encodingdefault}{\sfdefault}{bx}{n}

%% file: sections/looping_decisions.tex
\subsection{Compute-Optimal Decisions for Looping}
\label{app:family-design}

Looped models introduce four choices beyond the base Transformer recipe: how to allocate blocks across the prelude, core, and coda; how many times to apply the core; how to increase that count during training; and how to allocate tokens after growth. We organize the evidence around these decisions. The experimental controls differ across the sweeps: the fixed-token ladders compare model sizes and recurrence counts at matched compute, whereas the fixed-anchor sweeps trade training tokens for recurrence within each compute budget. Appendix~\ref{app:tpp-derivations} derives the TPP relationships used to interpret the latter sweeps.

\subsubsection{Allocating Blocks across the Prelude, Core, and Coda}
\label{app:block-allocation}

\begin{figure}[htbp]
\centering
\includegraphics[width=\linewidth]{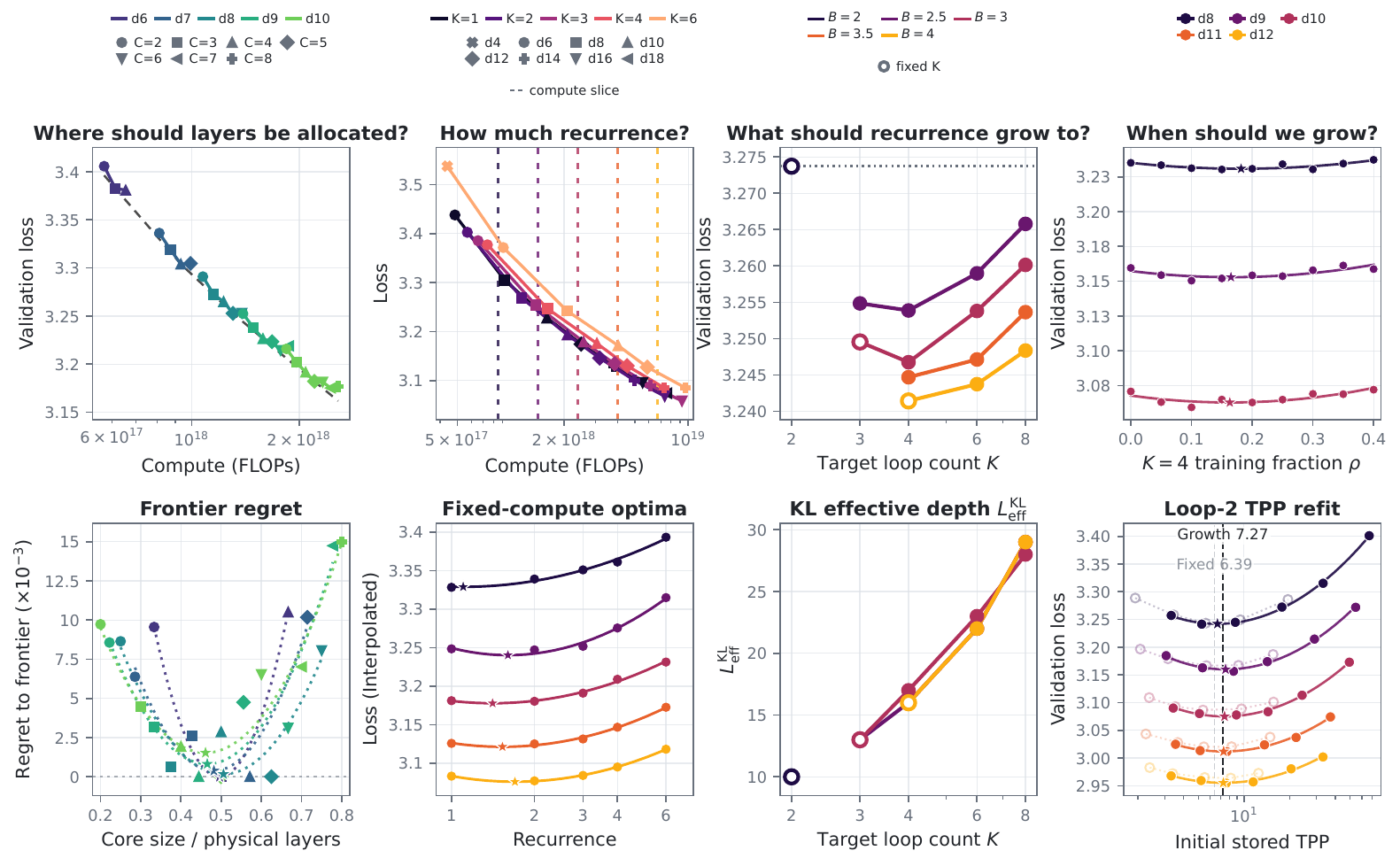}
\caption{\textbf{Allocation, recurrence, and growth sweeps.} \textbf{Left:} loss versus compute for the 1B-token core-allocation sweep, followed by regret to the compute frontier versus core fraction; curves fit each depth's regret and stars mark their minima. \textbf{Middle left:} fixed-token recurrence ladders and fixed-compute slices interpolated in log loss versus log compute. \textbf{Middle right:} loss and logit-KL effective depth versus the growth target, with filled points for growth from $K=2$ and hollow points for fixed-$K$ anchors. \textbf{Right:} Loop-2 growth-fraction sweeps at the depth-matched TPP-6 compute budgets, followed by the initial-stored-TPP refit; filled points use growth and hollow points are fixed controls. These panels summarize separate experiments with the controls described in the text.}
\label{fig:recurrence-design}
\end{figure}

We sweep the core size of tied, two-pass models at several stored depths, training each model for 1B tokens with the tuned $d8$ recipe. Because a larger core executes more blocks per token, we compare losses against a common compute frontier rather than only within a fixed stored depth (Figure~\ref{fig:recurrence-design}, left). The preferred core fraction varies little with depth, supporting a family whose three regions grow in proportion. We use a simple shared allocation: divide the blocks as evenly as possible across the prelude, core, and coda, assigning remainders first to the core and then to the coda.

\subsubsection{Choosing the Number of Core Passes}
\label{app:recurrence-choice}

\begin{figure}[htbp]
\centering
\includegraphics[width=\linewidth]{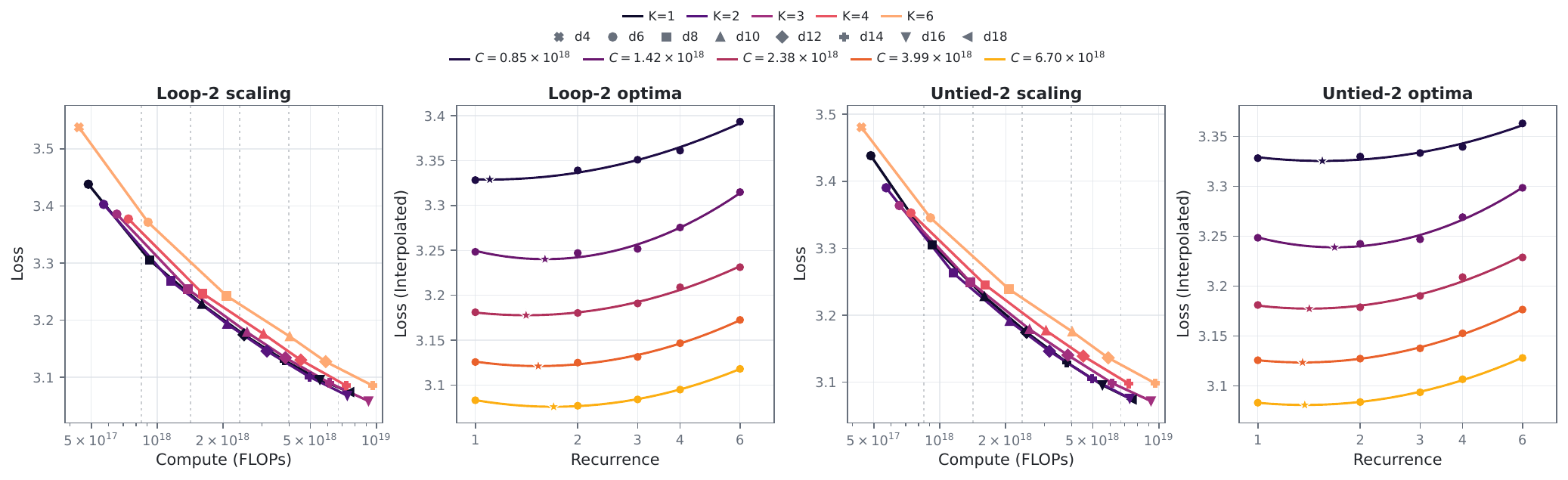}
\caption{\textbf{Recurrence choice in fixed-token ladders.} Every model trains on 1B tokens with the Operator-1 base recipe. For each family, the loss--compute ladders are interpolated at common budgets, and quadratics in log recurrence and log loss locate the fitted optima (stars). The two families share the $K=1$ ladder and are FLOPs-matched at each depth and recurrence.}
\label{fig:fresh-loop-optima}
\end{figure}

\paragraph{Fixed-token ladders favor one to two passes.}
For each core-pass count $K\in\{1,2,3,4,6\}$, we train a ladder of model sizes on 1B tokens with the Operator-1 base recipe. We interpolate each ladder's loss at common compute budgets and fit loss against $K$. For both tied and untied models, the fitted optimum stays between one and two passes across the measured budgets; $K=3,4,6$ give higher loss (Figure~\ref{fig:fresh-loop-optima}). We use $K=2$ for the fixed-recurrence variants on this basis.

\begin{figure}[htbp]
\centering
\includegraphics[width=\linewidth]{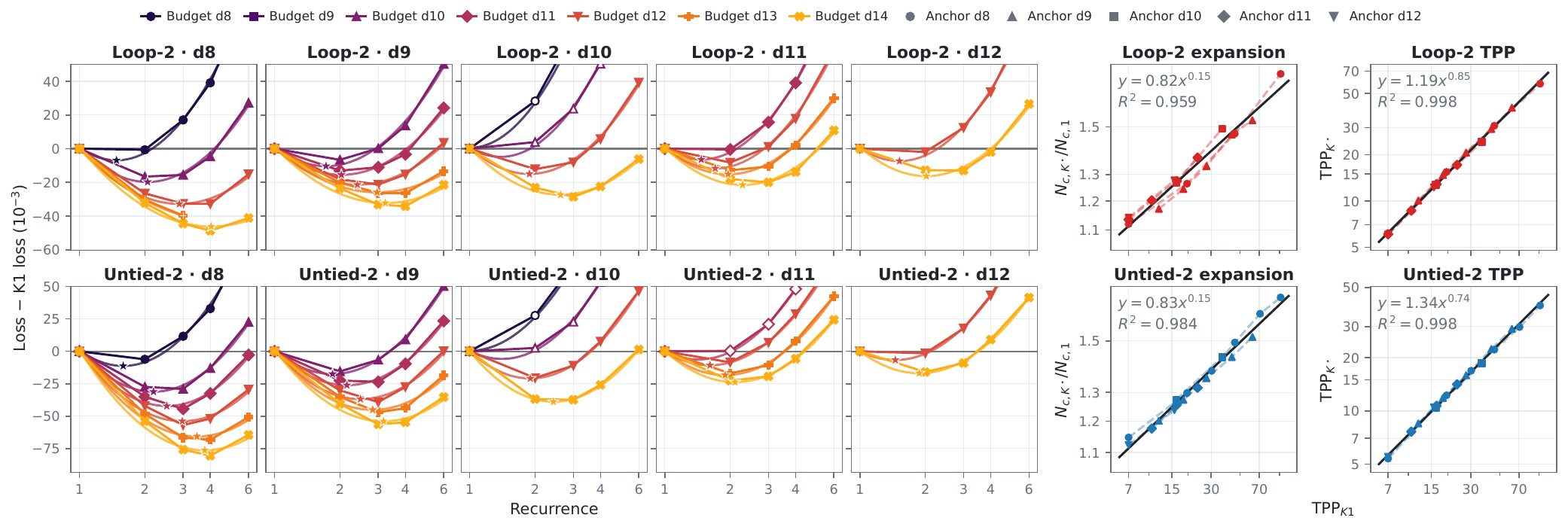}
\caption{\textbf{Recurrence choice at fixed anchor size and compute.} The first row uses tied cores and the second untied cores. \textbf{Left:} loss relative to $K=1$ versus recurrence, with anchor depth increasing across columns; each series holds compute fixed by adjusting tokens. Quadratic fits locate $K^\star$. \textbf{Right:} compute-active expansion and tokens per stored parameter at the fitted optimum versus the $K=1$ baseline TPP. Hollow points mark optima outside the measured recurrence grid and are excluded from the fitted laws.}
\label{fig:tpl}
\end{figure}

\paragraph{At fixed anchor size, more compute favors more passes.}
We next hold the anchor depth fixed, sweep the same $K$ values, and reduce training tokens as $K$ increases to match compute. Across anchors $d8$--$d12$, the fitted optimal count $K^\star$ increases with the budget (Figure~\ref{fig:tpl}, left), consistent with \citet{prairie2026parcae}. Thus, at fixed anchor size, larger budgets favor allocating some compute to extra passes. This comparison does not test whether extra passes outperform increasing model size.

Figure~\ref{fig:tpl} (right) summarizes the optima across anchors using baseline tokens per stored parameter, $\mathrm{TPP}_{K1}$: the tokens affordable at $K=1$ under the same budget, divided by that model's stored parameter count. Define $N_{c,K}$ as the \emph{compute-active parameter count}: the prelude, coda, and output-head matrix parameters counted once, plus the core matrix parameters counted $K$ times, even when shared. This count excludes the input embedding lookup; $N_{c,1}$ is the count for the same anchor with one core pass. The expansion is $E_K=N_{c,K}/N_{c,1}$. At $K^\star$, we fit
\begin{equation}
\begin{aligned}
E_{K^\star} &= 0.82\,\mathrm{TPP}_{K1}^{0.15} && \text{(tied)},\\
E_{K^\star} &= 0.83\,\mathrm{TPP}_{K1}^{0.15} && \text{(untied)}.
\end{aligned}
\label{eq:recurrence-expansion-fit}
\end{equation}
The rightmost panels show that tokens per stored parameter at the optimum also increase with baseline TPP, but sublinearly: some of the additional budget goes to recurrence. Appendix~\ref{app:tpp-derivations} gives these TPP fits and derives their relation to compute expansion.

\subsubsection{Choosing the Growth Target and Transition}
\label{sec:grow}
\label{app:grow}

\paragraph{Grow from two to four passes.}
Starting from the tied $d8$ model with a $2/3/3$ block split, we sweep growth targets $K\in\{3,4,6,8\}$ over 1B training tokens. At each compute budget, the transition is chosen to match the budget; thus a larger target is used for a smaller fraction of the run. Growing to four passes gives the lowest loss at every tested budget (Figure~\ref{fig:recurrence-design}, middle right). Larger targets increase logit-KL effective depth without consistently improving loss. We adopt the $2\to4$ schedule for the growth variants: \KtwoGrow reuses its tied core more times, while \DepGrow duplicates the trained untied cores. Deep Vanilla Grow applies the same duplication schedule to plain Transformer blocks with the same prelude--core--coda allocation.

\begin{figure}[htbp]
\centering
\begin{subfigure}[t]{0.49\linewidth}
\vspace{0pt}
\centering
\includegraphics[width=\linewidth]{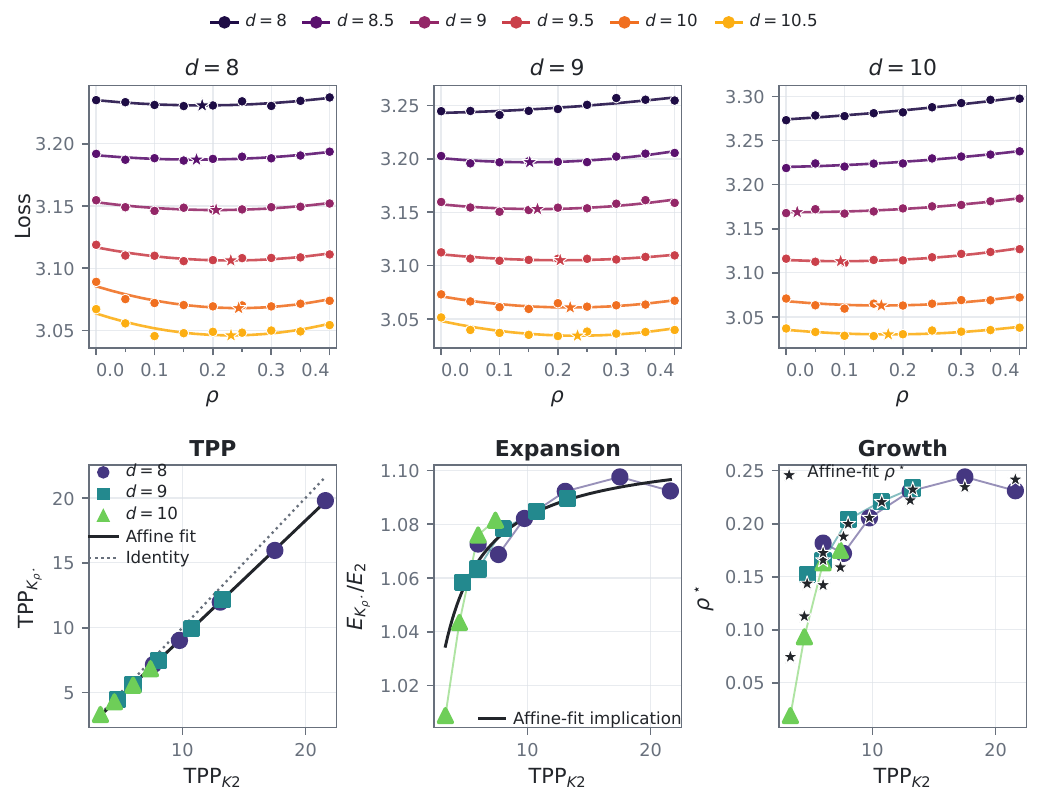}
\caption{Tied growth: \KtwoGrow}
\label{fig:opt-grow-tied}
\end{subfigure}\hfill
\begin{subfigure}[t]{0.49\linewidth}
\vspace{0pt}
\centering
\includegraphics[width=\linewidth]{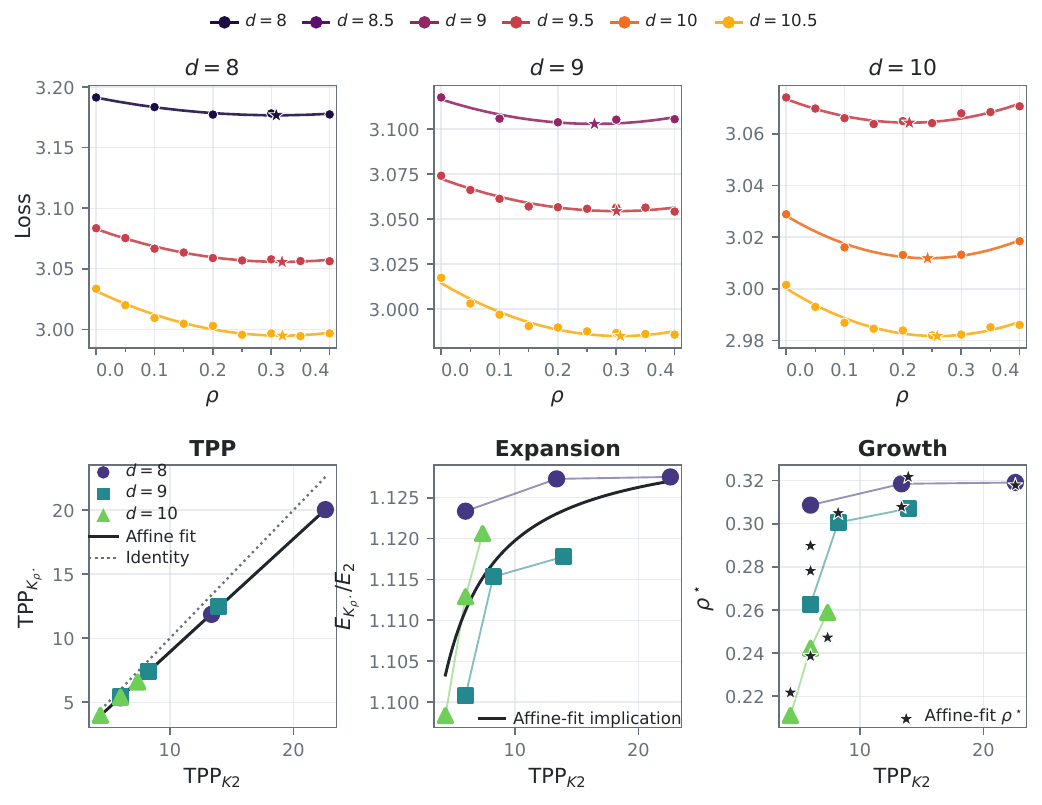}
\caption{Untied growth: \DepGrow}
\label{fig:opt-grow-untied}
\end{subfigure}

\medskip
\begin{subfigure}[t]{0.49\linewidth}
\vspace{0pt}
\centering
\includegraphics[width=\linewidth]{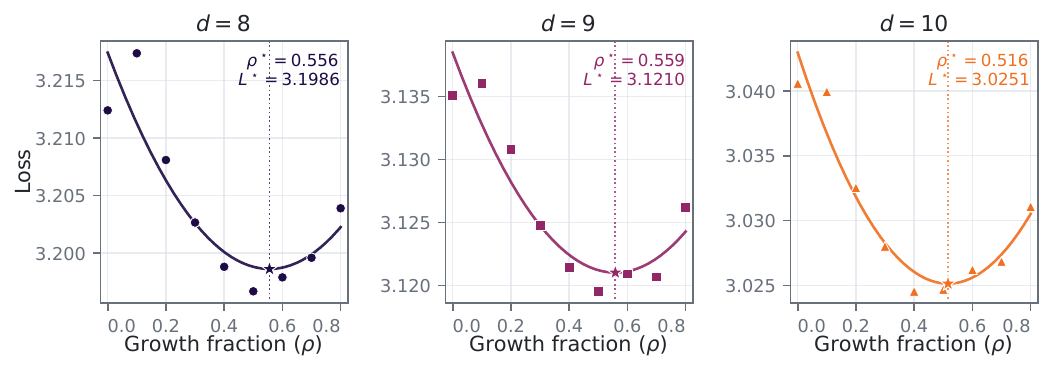}
\caption{Deep Vanilla Grow}
\label{fig:opt-grow-deepvan}
\end{subfigure}\hfill
\begin{subfigure}[t]{0.49\linewidth}
\vspace{0pt}
\centering
\includegraphics[width=\linewidth]{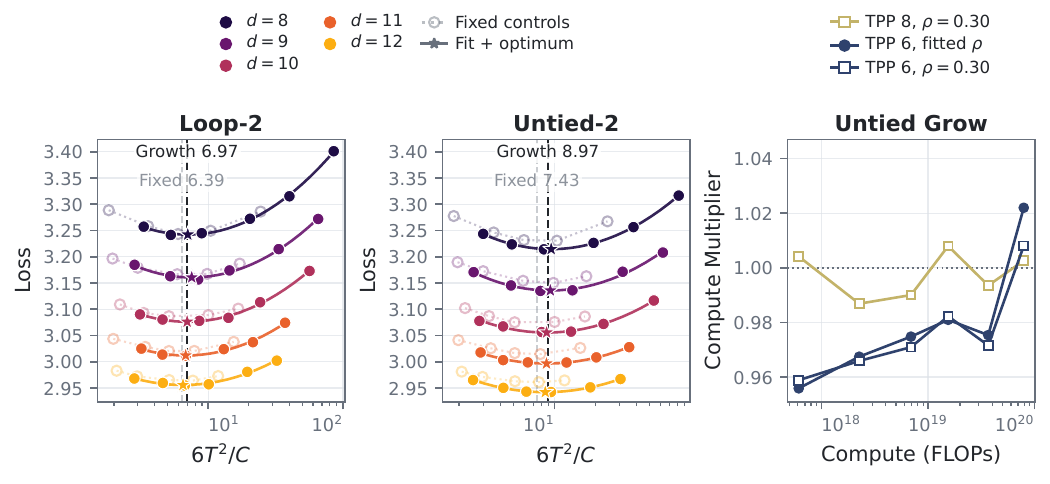}
\caption{Token allocation and growth prescription}
\label{fig:opt-grow-tokens}
\label{fig:opt-grow-regret}
\end{subfigure}
\caption{\textbf{Growth timing, token allocation, and recipe sensitivity.} \textbf{(a,b)} Top rows show validation loss versus the post-transition token fraction $\rho$ at three anchor depths; colors identify fixed-compute budgets, curves are quadratic fits, and stars mark bracketed minima. Bottom rows show initial-stored-parameter TPP at the fitted transition, compute-active expansion, and optimal growth fraction versus fixed-$K=2$ TPP. Colors and markers identify anchor depth; black curves and stars show the affine fit and its implications (Appendix~\ref{app:tpp-derivations}). \textbf{(c)} Deep Vanilla Grow timing sweeps at each anchor's fixed Deep Vanilla TPP-6 budget, with fitted minima marked by stars. \textbf{(d)} Left and middle: loss versus $6T^2/C$ for tied and untied growth, where $T$ is tokens and $C$ is compute. Filled points use growth, hollow points are fixed controls, stars mark fitted minima, and dashed lines mark mean optima. Right: the \DepGrow prescription ablation at baseline depths $d6$--$d16$, even. The multiplier is the TPP-8, fitted-$\rho$ baseline's compute divided by each alternative's compute at matched loss, using the two nearest measured points in log-compute/log-loss space.}
\label{fig:opt-grow}
\end{figure}

\paragraph{Choose the transition at fixed depth and compute.}
Let $\rho$ be the fraction of training tokens processed \emph{after} growth, so a larger $\rho$ means an earlier transition. With the target fixed at four passes, we sweep $\rho$ at three anchor depths and six compute budgets for \KtwoGrow, and at the same anchors with three budgets each for \DepGrow. Tokens are adjusted to keep each depth--budget pair at fixed compute. Quadratic fits to validation loss locate $\rho^\star$. Larger budgets generally favor earlier growth, but the minima are broad (Figures~\ref{fig:opt-grow-tied} and~\ref{fig:opt-grow-untied}, top rows).

Across depths and budgets, the token allocation at the fitted optimum follows an affine relationship,
\begin{equation}
\mathrm{TPP}_{K_{\rho^\star}}=a\,\mathrm{TPP}_{K2}+b,
\label{eq:growth-affine}
\end{equation}
where both TPP quantities use the initial $K=2$ stored parameter count: $\mathrm{TPP}_{K2}$ is the allocation without growth at the same compute, and $\mathrm{TPP}_{K_{\rho^\star}}$ is the allocation with the fitted transition. The coefficients are $(a,b)=(0.902,0.216)$ for \KtwoGrow and $(0.883,0.103)$ for \DepGrow, with $R^2>0.9999$ for both. As derived in Appendix~\ref{app:tpp-derivations}, this fit implies a compute expansion that rises and then plateaus with baseline TPP, motivating a nearly constant growth fraction at sufficiently high TPP. At the high-TPP end of the measured sweeps, the fitted fractions are approximately $0.23$--$0.24$ for \KtwoGrow and $0.26$--$0.32$ for \DepGrow. The exact conversion from expansion to $\rho$ depends slightly on the model's block allocation.

For Deep Vanilla Grow, we sweep one fixed-compute budget at each of $d8$, $d9$, and $d10$, using each anchor's fixed Deep Vanilla TPP-6 budget. The fitted fractions are $0.556$, $0.559$, and $0.516$, respectively, with similarly shallow minima near half of training (Figure~\ref{fig:opt-grow-deepvan}).

\subsubsection{Allocating Tokens and Testing Recipe Sensitivity}
\label{app:growth-sensitivity}

Growth increases average compute per token, so the optimal token allocation must be refitted. We repeat the iso-compute sweeps with growth in place (Figure~\ref{fig:opt-grow-tokens}, left and middle). The figure uses $6T^2/C$, where $T$ is training tokens and $C$ is training compute; under the leading-order relation $C=6TN_{c,\mathrm{eff}}$, this coordinate equals tokens per training-averaged compute-active parameter. The mean optimum in this coordinate rises from $6.39$ to $6.97$ for tied growth and from $7.43$ to $8.97$ for untied growth. This coordinate differs from tokens per initial \emph{stored} parameter, which we use to specify the ladder recipes in Table~\ref{tab:full-ladder-recipes}. Across the measured budgets, growth favors a smaller initial model trained on more tokens, although the loss curves are flat near their minima.

We test the cost of simplifying both the token allocation and the transition for \DepGrow (Figure~\ref{fig:opt-grow-regret}, right). The ablation crosses initial stored TPP $8$ versus $6$ with the fitted per-size $\rho$ versus a constant $\rho=0.30$. The matched-loss compute multipliers remain close to one: reusing TPP $6$ changes the estimated compute requirement by less than about $5\%$, and replacing fitted $\rho$ with $0.30$ changes validation loss by $-0.0009$ to $+0.0019$. Thus, within this ablation, precise per-size tuning has little benefit. The saturation of the fitted expansion and the broad loss minima support a simple recipe with a fixed growth fraction and a rounded TPP.

%% file: sections/tpp_derivations.tex
\clearpage
\subsection{TPP Relations for Recurrence and Growth}
\label{app:tpp-derivations}

The sweeps in Appendix~\ref{app:family-design} compare different recurrence counts at fixed training compute. This section derives how those comparisons change tokens per stored parameter. We distinguish stored parameters, which determine TPP, from compute-active parameters, which determine the leading-order compute cost. The identities below use $C=6DN_c$; the plotted compute budgets use the model FLOP estimator, including attention, as described in Appendix~\ref{app:architecture}.

\subsubsection{Fixed Recurrence}

For recurrence $K$, let $D_K$ be the number of training tokens and $N_{s,K}$ the stored parameter count. The compute-active count $N_{c,K}$, defined in Appendix~\ref{app:recurrence-choice}, counts weight-matrix parameters once per application, including all $K$ core passes, and excludes the input embedding lookup. The subscript $1$ denotes the same anchor model with one core pass. Define the compute-active and stored-parameter expansions relative to this baseline by
\begin{equation}
\mathrm{TPP}_K=\frac{D_K}{N_{s,K}},
\qquad
E_K=\frac{N_{c,K}}{N_{c,1}},
\qquad
S_K=\frac{N_{s,K}}{N_{s,1}}.
\label{eq:tpp-expansion-definitions}
\end{equation}
At a fixed compute budget, $D_KN_{c,K}=D_1N_{c,1}$, so $D_K=D_1/E_K$. Dividing by the stored parameter count gives
\begin{equation}
\mathrm{TPP}_K
=\frac{D_1/E_K}{S_KN_{s,1}}
=\frac{\mathrm{TPP}_{K1}}{E_KS_K}.
\label{eq:recurrence-tpp-identity}
\end{equation}
Increasing recurrence therefore reduces TPP through the increased compute per token and, for untied models, through the increased stored parameter count. Tied recurrence reuses one core, so $S_K=1$. Untied recurrence stores a separate core for each pass; when block parameters dominate, $S_K$ approaches $E_K$. With an approximately equal prelude--core--coda allocation, $E_K$ approaches $(K+2)/3$ for either family. At finite size, stored and compute-active counts differ, including because embedding lookup contributes stored parameters without the same matrix-multiplication cost.

Equation~\ref{eq:recurrence-tpp-identity} also holds at the fitted optimum $K^\star$. Together with the expansion fits in Equation~\ref{eq:recurrence-expansion-fit}, it motivates a power-law relationship between optimal TPP and baseline TPP. The measured fits in Figure~\ref{fig:tpl} are
\begin{equation}
\begin{aligned}
\mathrm{TPP}_{K^\star} &= 1.19\,\mathrm{TPP}_{K1}^{0.85} && \text{(tied)},\\
\mathrm{TPP}_{K^\star} &= 1.34\,\mathrm{TPP}_{K1}^{0.74} && \text{(untied)}.
\end{aligned}
\label{eq:recurrence-tpp-fit}
\end{equation}
These are empirical fits; the finite-size stored-parameter expansion enters the untied relation through $S_{K^\star}$.

\subsubsection{Growth from Two to Four Passes}

Let $\rho$ be the fraction of tokens processed after the transition from $K=2$ to $K=4$. The token-weighted average recurrence and compute-active parameter count are
\begin{equation}
\begin{aligned}
K_\rho &= 2(1-\rho)+4\rho,\\
N_{c,K_\rho} &= (1-\rho)N_{c,2}+\rho N_{c,4},\\
C &= 6D_{K_\rho}N_{c,K_\rho}.
\end{aligned}
\label{eq:growth-compute}
\end{equation}
Here $N_{c,K_\rho}$ denotes a training average, rather than a model executing a fractional number of passes. For growth, both TPP coordinates use the \emph{initial} stored parameter count $N_{s,2}$:
\begin{equation}
\mathrm{TPP}_{K_\rho}=\frac{D_{K_\rho}}{N_{s,2}},
\qquad
\mathrm{TPP}_{K2}=\frac{D_{K2}}{N_{s,2}}.
\end{equation}
The fixed-$K=2$ baseline spends the same compute as the growth run, so $D_{K_\rho}N_{c,K_\rho}=D_{K2}N_{c,2}$. Using $E_K=N_{c,K}/N_{c,1}$ gives
\begin{equation}
\mathrm{TPP}_{K_\rho}
=\frac{E_2}{E_{K_\rho}}\,\mathrm{TPP}_{K2},
\qquad
\frac{E_{K_\rho}}{E_2}
=1+\rho\left(\frac{E_4}{E_2}-1\right).
\label{eq:growth-tpp-expansion}
\end{equation}
There is no extra stored-parameter expansion factor here because both TPP quantities use the same initial denominator, including for untied growth.

At the fitted transition, the affine law in Equation~\ref{eq:growth-affine} implies
\begin{equation}
\begin{aligned}
\frac{E_{K_{\rho^\star}}}{E_2}
&=\frac{\mathrm{TPP}_{K2}}{a\,\mathrm{TPP}_{K2}+b},\\
\rho^\star
&=\frac{E_{K_{\rho^\star}}/E_2-1}{E_4/E_2-1}.
\end{aligned}
\label{eq:growth-rho-law}
\end{equation}
When $b$ is small relative to $a\,\mathrm{TPP}_{K2}$, the optimal compute expansion approaches $1/a$. This yields the rise and plateau in the bottom rows of Figures~\ref{fig:opt-grow-tied} and~\ref{fig:opt-grow-untied} and explains why the fitted growth fraction becomes weakly dependent on TPP. Equal baseline TPP predicts equal compute expansion, but need not predict exactly equal $\rho^\star$: the conversion also depends on $E_4/E_2$, which varies with the integer block allocation and non-core compute. When the three regions approach equal proportions and block compute dominates, $E_4/E_2\to3/2$, giving the limiting prescription $\rho^\star\to2(1/a-1)$.